\documentclass{egpubl}
\usepackage{egsgp2022}
\SpecialIssuePaper         

\usepackage[T1]{fontenc}
\usepackage{dfadobe}  

\usepackage{cite}  

\BibtexOrBiblatex
\electronicVersion
\PrintedOrElectronic
\ifpdf \usepackage[pdftex]{graphicx} \pdfcompresslevel=9
\else \usepackage[dvips]{graphicx} \fi

\usepackage{egweblnk} 

\usepackage{epsfig}
\usepackage{amsmath}
\usepackage{amssymb}
\usepackage{amsfonts} 
\usepackage{overpic}
\usepackage{wrapfig}
\usepackage{bm}
 \usepackage{comment}
\usepackage{pgfplots}
\usepackage{pgf}
\usepackage{pgfplots}
\usepackage{pgf,tikz}
\usepackage{array}
\usepackage{pict2e}
\usepackage{soul}
\usepackage{tabstackengine}
\usepackage{booktabs}

\usepackage{graphicx}
\usepackage{epsfig}
\usepackage{subcaption}

\newcommand{\subX}{\mathcal{R}}
\newcommand{\X}{\mathcal{X}}

\newcommand{\M}{\mathcal{M}}

\newcommand{\red}[1]{\textcolor{red}{#1}}

\newcommand{\blue}[1]{\textcolor{blue}{#1}}\definecolor{ourblue}{rgb}{0.00000,0.44700,0.74100}
\newcommand{\ourblue}[1]{\textcolor{ourblue}{#1}}

\definecolor{revisionColor}{rgb}{1.0, 0.25, 0}
\newcommand{\revision}[1]{\textcolor{black}{#1}}

\definecolor{first}{rgb}{1.0, 0.0, 0}
\definecolor{second}{rgb}{0.7, 0.0,0}
\definecolor{third}{rgb}{0.3, 0,0}
\newcommand{\first}[1]{\textbf{\textcolor{first}{#1}}}
\newcommand{\second}[1]{\textbf{\textcolor{second}{#1}}}
\newcommand{\third}[1]{\textbf{\textcolor{third}{#1}}}

\newcolumntype{C}{>{\centering\arraybackslash}p{0.95cm}}
\newcolumntype{k}{>{\centering\arraybackslash}p{1.7cm}}

\newcommand{\figref}[1]{Figure \ref{#1}}

\title[Localized Shape Modelling with Global Coherence]%
      {Localized Shape Modelling with Global Coherence:\\An Inverse Spectral Approach}

\author[M. Pegoraro \& al.]
{\parbox{\textwidth}{\centering 
          M. Pegoraro$^1$\orcid{0000-0001-5690-8403}   and S. Melzi$^{1,2}$\orcid{0000-0003-2790-9591} and U. Castellani$^3$\orcid{0000-0002-6099-5682} and R. Marin\thanks{Equal contribution}$^1$\orcid{0000-0003-2392-4612} and E. Rodolà\footnotemark[1]$^1$\orcid{0000-0003-0091-7241}
 }
        \\
{\parbox{\textwidth}{\centering 
    $^1$Sapienza University of Rome, Italy\\
         $^2$Bicocca University of Milan, Italy\\
         $^3$University of Verona, Italy
      }
}
}

\begin{document}


\maketitle
\begin{abstract}
Many natural shapes have most of their characterizing features concentrated over a few regions in space.
%
%
For example, humans and animals have distinctive head shapes, while inorganic objects like chairs and airplanes are made of well-localized functional parts with specific geometric features.
Often, these features are strongly correlated -- a modification of facial traits in a quadruped should induce changes to the body structure. 
However, in shape modelling applications, these types of edits are among the hardest ones; they require high precision, but also
%
%
a global awareness of the entire shape. 
Even in the deep learning era, obtaining manipulable representations that satisfy such requirements is an open problem posing significant constraints. 
In this work, we address this problem by defining a data-driven model upon a family of linear operators (variants of the mesh Laplacian), whose spectra capture global and local geometric properties of the shape at hand. Modifications to these spectra are translated to semantically valid deformations of the corresponding surface. By explicitly decoupling the global from the local surface features, our pipeline allows to perform local edits while simultaneously maintaining a global stylistic coherence.
%
%
%
We empirically demonstrate how our learning-based model generalizes to shape representations not seen at training time, and we systematically analyze different choices of local operators over diverse shape categories. 


\end{abstract}  
\section{Introduction}
\label{sec:introduction}
Generative modeling of $3$D shapes is a fascinating problem that unifies geometry and statistics at their finest. The interest is both theoretical and practical; we aim to better understand the rules lying ``under the surface'' and, as a direct consequence, to solve real-world problems like automatic content generation, $3$D shape reconstruction, and body tracking \cite{EG_star_generative,egger20203d}, among many others.
This problem is as compelling as it is challenging, and the research in the field has a long history. For several decades, many linear statistical approaches have been proposed \cite{robinette1999caesar,loper15, MANO2017}, until deep learning methods took the stage, unleashing powerful non-linear methods \cite{EG_star_generative}. However, even if the results are getting better every day, many underlying properties remain mysterious, hardly interpretable and controllable. 
\begin{figure}[t!]
\centering
    \begin{overpic}[trim=0cm 0cm 0cm 0cm,clip,width=0.92\linewidth]{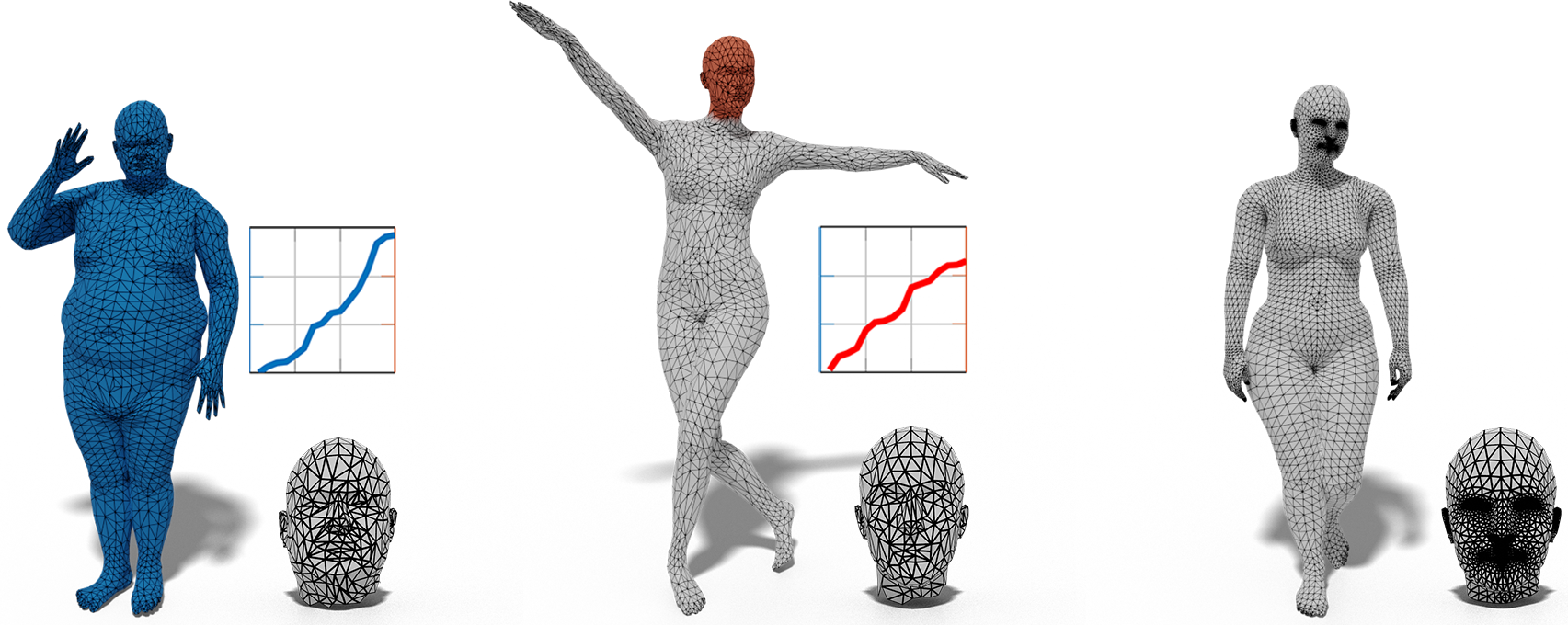}
    \put(31,18){\large $\bigoplus$}
    \put(69,18){\large$\mathbf{=}$}
    \end{overpic}
\caption{\label{fig:teaser} An example of semantic control provided by our method. By combining a global spectrum (in blue) with a local spectrum (in red) of two shapes with different discretizations, we generate a new shape (on the right) possessing the global characteristics of the blue shape and the local features of the red one.}
\end{figure}

Inspired by recent work in spectral geometry processing, with this paper we propose a new shape modelling paradigm that addresses the following question: How should the global geometry of a shape be changed, to make it coherent with user-defined local edits that modify its semantics?

Instead of attempting to rigorously define a geometrical notion of `coherence', which is ultimately subjective, we propose to learn this concept from examples, namely, by finding latent correlations between locally and globally supported geometrical features on a collection of representative shapes. Technically, we achieve this by learning a sum operation between the spectra of standard and localized Laplacians, followed by an inverse learnable solver that recovers a 3D embedding from the combined spectrum.
Operating with spectra endows our pipeline with invariance to near-isometries by construction, as well as robustness to mesh discretization, and makes it applicable to any shape representation that admits the definition of a Laplace operator (e.g., point clouds).

We position our work within a recent line of approaches that emphasize the practical potential of using spectra as a rich, albeit compact representation of the shape geometry~\cite{isosp,hamiltonian,instant2020,advatt,Instant2021}. Differently from these approaches, which focus on applications such as shape correspondence, region detection, style transfer and adversarial attacks, here we regard the spectra as {\em manipulable} representations. Further, instead of operating with the standard Laplacian eigenvalues, we demonstrate for the first time how the combination of multiple spectra from different operators can lead to a practical benefit in shape modelling applications.

%
%
While working with multiple spectra at once makes the shape recovery step more challenging (since the network must interpret them all in one shot), we show that mixing spectra also improves the reconstruction quality, as well as generalizing more easily to unseen shapes. 
%
%
In the example of Fig.~\ref{fig:teaser}, we combine the Laplacian eigenvalues of the blue shape with the eigenvalues of a Laplace operator localized on the red region (head) of the shape in the middle, generating a new shape that globally reflects the features of the former (height and robustness), but which is coherent with the semantics of the latter (physiological gender). As shown in the example, our method can deal with shapes that do not share the same connectivity or pose.
As a final point, we show that this statistical correlation between local spectra and geometry is so strong that it holds even on unorganized point clouds, i.e., with noisy spectra and without known correspondence across the training data.

To summarize, our contribution is threefold: 
\begin{enumerate}
    \item 
    We address the task of enforcing global semantic consistency of 3D geometry, when this undergoes local user edits. We do this in a shape-from-spectrum setup, by proposing a generative model from \emph{multiple} spectra. The combination of global and local information is a novelty of our work, showing that this representation is capable of providing not only better reconstruction, but also new application possibilities;
    \item We propose a decoder-only architecture that directly connects the spectrum to the $3$D geometry. We show that this simple approach outperforms previous more sophisticated methods, it is more interpretable, and provides new insights on inverse spectral geometric problems;
    \item We propose a new dataset designed for analyzing inverse spectral methods, together with new error measures, establishing a sound protocol for evaluating the performance on this task.
\end{enumerate}
Code and data are available online\footnote{\url{https://github.com/Marco-Peg/Localized-Shape-Modelling-with-Global-Coherence}}.
\section{Related work}
\label{sec:related}

%

\vspace{1ex}\noindent\textbf{Generative models.} From a technical perspective, our method can be classified as a {generative model} due to its ability to generate new shapes by sampling a learned parametric space. 
In the realm of 3D shapes, existing generative models differ depending on the final application, and on the chosen representation for the 3D geometry. 
Popular representations 
include voxels \cite{3dgan}, triangle meshes \cite{COMA, Tretschk2020DEMEA, ATLASNET}, implicit functions  \cite{bhatnagar2020ipnet,ShapeGF}, and point clouds \cite{qi2017pointnet, achlioptas2018learning}. While each representation requires a specific architecture, 
%
%
shapes that undergo non-rigid deformations, and in particular the class of human bodies, have received increasing attention in the recent literature \cite{xu2020ghum, bhatnagar2020ipnet, Jiang2020HumanBody,joo2018total}. 

While most of these works focus on reconstruction quality, only a few have investigated ways to inject semantics in the generation process, in a controllable way. \cite{geodisent} proposed an autoencoder with a disentangled latent space, enabling a separate control of intrinsic and extrinsic deformations; \cite{cosmo2020limp} showed that plausibility of the generated shapes can be improved by promoting metric preservation in the loss function; \cite{zhou20unsupervised} proposed an unsupervised technique to disentangle shape and pose in the latent space representation\revision{; \cite{Chen_2021_CVPR} encoded geometric details as a style property that conditions the refinement of a low-resolution coarse voxel shape through a generative adversarial network (GAN).}
Other related works addressed the generation of rigid composite-objects\cite{L2G2019, shapehandles, Wu_2020_CVPR,mo2019structurenet, yin2020coalesce} exploiting hierarchical neural network architectures or probabilistic mixture models to manipulate shape parts~\cite{achlioptas2018learning, hertz2020pointgmm, luo2021simpmodeling}.

\vspace{1ex}\noindent\textbf{Shape from spectrum.} Recently, the Laplacian eigenvalues have been used as a compact representation to recover and manipulate $3$D geometry. According to a physical interpretation, the eigenfunctions of the Laplace operator on a surface relate to the evolution of waves over it, and the associated eigenvalues are the frequencies of such waves. These are determined uniquely by the intrinsic geometry of the shape, and are fully invariant to isometric deformations. However, the inverse problem (i.e., determining the intrinsic geometry from a set of Laplacian eigenvalues) has been an open question for a long time \cite{kac1966can}, with the negative result of Gordon and colleagues \cite{gordon1992one} posing a theoretical tombstone to the problem. 

The vision community has recently rediscovered interest in this problem from a practical perspective, with \cite{isosp,hamiltonian} showing that this inverse problem can be solved through a complex optimization. 
 The recent work \cite{instant2020}, and its extension \cite{Instant2021}, replace the costly optimization with a data-driven framework, where a latent encoding is connected with the Laplacian spectrum via trainable maps. At test time, the network can instantaneously recover a shape from its spectrum. While we consider these works the closest to ours by their data-driven nature, the authors of \cite{instant2020,Instant2021} limit their analysis to the standard Laplacian, without investigating localized operators such as those studied in \cite{LMH,Choukroun}. Importantly, the methods 
 of \cite{instant2020,Instant2021} address the generation problem by designing a network that is hard to interpret, without providing the user with a way to exert control on the desired output. 

\vspace{1ex}\noindent\textbf{Our method.} With this paper, we propose a generative model for 3D shapes that makes full use of the spectrum as an informative, compact, and manipulable representation. Our method is straightforward as it relies upon a simple decoder-only network, and considers a combination of different spectra. The only loss we use is a standard reconstruction loss, without any ad-hoc regularizer. This way, our network is encouraged to discover by itself the hidden mechanism that links a (combined) spectrum to the corresponding $3$D shape. Hence, the statistical relations within an object class emerge, providing in turn a better control of the generative process.

\section{Background and notation}
\label{sec:background}

\vspace{1ex}\noindent\textbf{Smooth setting.}
In this setting, a 3D shape is modelled as a compact and connected Riemannian surface (2-dimensional manifold) $\X$ embedded in $\mathbb{R}^3$. 
%
Each surface $\X$ is equipped with a Laplace-Beltrami operator (LBO) $\Delta_\X$, which generalizes the classical Laplacian operator to non-Euclidean domains.
From now on, we will refer to this operator as Laplacian to streamline our notation.
The Laplacian $\Delta_\X$ is a positive semi-definite operator. From its eigendecomposition we obtain the eigenvalues $\{\lambda_{\ell} \}$
with $\lambda_{\ell} \in \mathbb{R}$,
$0 \le \lambda_1\le\lambda_2\le\ldots\le\lambda_{\infty}$, 
and associated orthogonal eigenfunctions $\{\phi_{\ell}\}$.  
%
%
%
%
When $\X$ is a manifold with boundary $\partial\X$, we consider homogeneous Dirichlet boundary conditions as done in \cite{hamiltonian}:
\begin{align}
    \phi_{\ell}(x) = 0, \ \forall x\in\partial\X \,.
    \label{eq:bc}
\end{align}
The ordered discrete sequence of Laplacian eigenvalues 
of $\X$ is usually referred to as the {\em spectrum} of $\X$.
%
When $\X$ is a 1D Euclidean domain, the eigenvalues coincide with the squares of frequencies of Fourier basis functions. 
Following this analogy, we only consider a truncated spectrum corresponding to the $k$ eigenvalues with smallest absolute values as a band-limited representation.
%
%
In this paper, we test values of $k$ in the range from $10$ to $80$.
%
%

%
\setlength{\columnsep}{7pt}
\setlength{\intextsep}{1pt}

\begin{wrapfigure}[6]{r}{0.3\linewidth}
\vspace{-0.6cm}

\begin{center}
\begin{overpic}
[trim=0cm 0cm 0cm 0cm,clip,width=\linewidth]{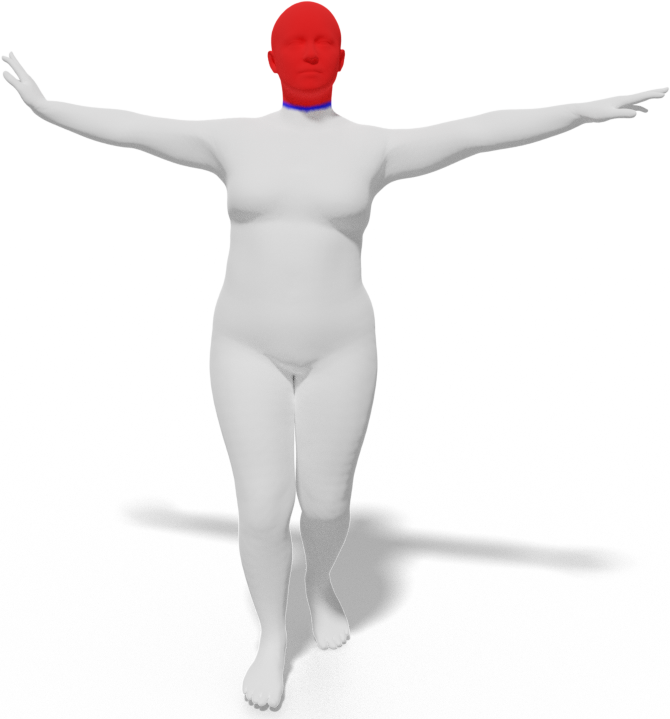}
\put(55,53.5){ $\X$}
\put(50,90.5){\footnotesize $\red{\subX}$}
\put(23,85){\footnotesize $\blue{\partial\subX}$}
\end{overpic}
\end{center}
\end{wrapfigure}
In this work, we also consider connected submanifolds $\subX \subset \X $ 
with boundary $\partial\subX$ 
(see inset figure for an illustration). The region $\subX$ inherits the metric from the complete manifold $\X$, and is similarly equipped with a Laplacian operator $\Delta_{\subX}$.

\vspace{1ex}\noindent\textbf{Discrete setting.}
We discretize smooth surfaces, submanifolds, and their Laplacians in two alternative ways: (i)
\emph{Triangle meshes}: $\M = (X,F)$ with $n$ vertices $X$ and $m$ triangular faces $F$. Submanifolds are constructed as subsets of vertices and faces, with their local connectivity preserved. We adopt the cotangent formula to discretize the Laplacian~\cite{pinkall1993computing}. (ii) \emph{Unorganized point clouds}: $\mathcal{P} = (X)$. Submanifolds are simply subsets of the vertices. We compute the Laplacian using the implementation of~\cite{Sharp2020}\footnote{\url{https://github.com/nmwsharp/robust-laplacians-py}}.
In both representations, the matrix $X\in \mathbb{R}^{n \times 3}$ encodes vertex coordinates.

\section{Method}
\label{sec:method}
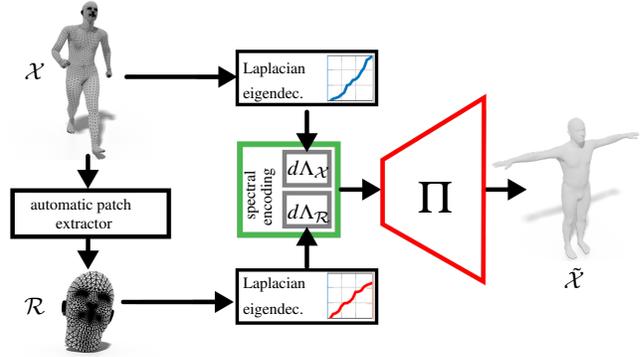
\begin{figure}[t!]
    \begin{overpic}[trim=0cm 0cm 0cm 0cm,clip,width=1\linewidth]{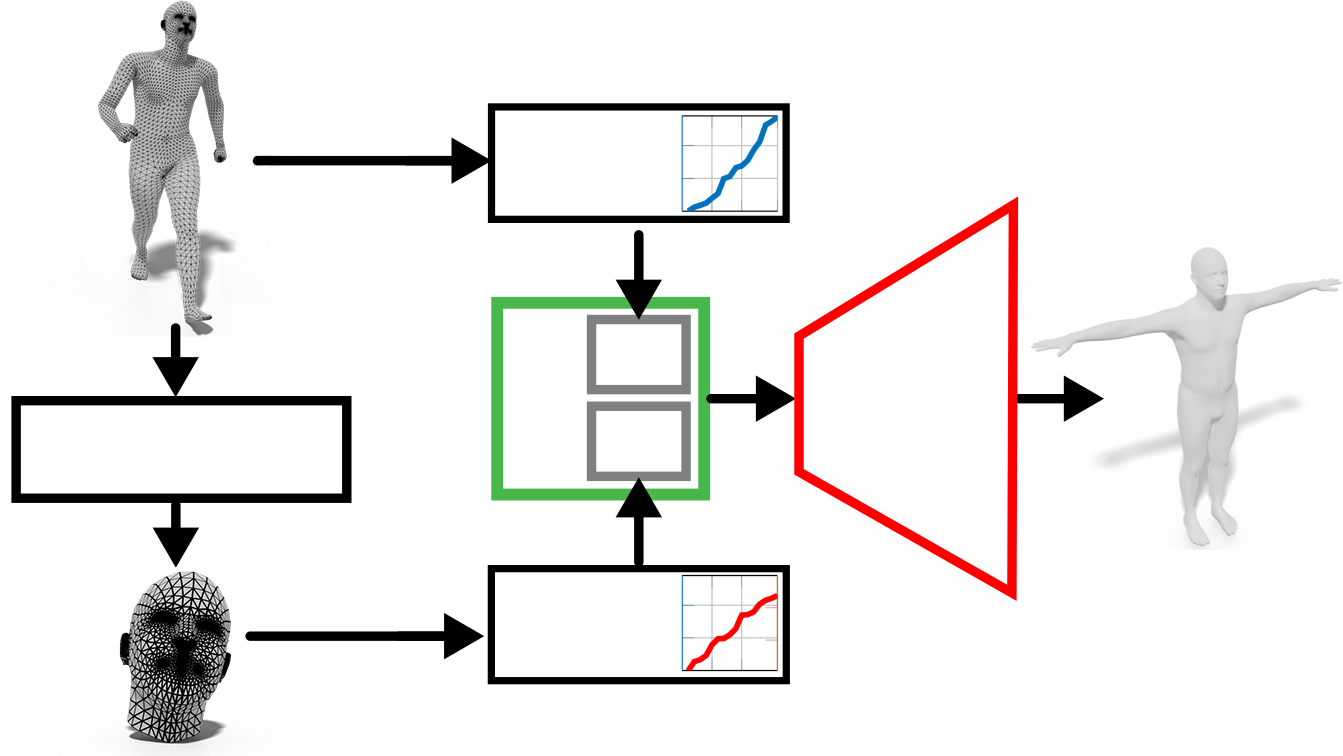}
    \put(3,44){$\X$}
    \put(3,7){$\subX$}
    \put(44.5,29){\footnotesize $d\Lambda_{\X}$}
    \put(44.5,22){\footnotesize $d\Lambda_{\subX}$}
    \put(4,23.5){\scriptsize automatic patch}
    \put(8,20.5){\scriptsize extractor}
    \put(65,23){\huge{$\Pi$}}
    \put(88,11){$\tilde{\X}$}
    \put(37.5,11){\scriptsize Laplacian}
    \put(37.5,7){\scriptsize eigendec.}
    \put(37.5,45){\scriptsize Laplacian}
    \put(37.5,41){\scriptsize eigendec.}
    \put(37.7,21.2){\rotatebox{90}{\scriptsize{ spectral}}}
    \put(40.3,20){\rotatebox{90}{ \scriptsize{encoding}}}
    
    \end{overpic}
    
\caption{\label{fig:arch} The proposed model. A neural network $\Pi$ is trained to reconstruct 3D shapes from spectral encodings, obtained as the simple concatenation of a global and a local spectral encoding. Note the lack of a learnable block between shapes and spectral encoding, which differentiates our model from a classical autoencoder architecture. Therefore, $\Pi$ directly maps spectral encodings to 3D coordinates, without ever seeing the input region.}
\end{figure}

This section outlines the proposed method, providing details about its implementation and theoretical insights. A visualization of our pipeline is given in Fig.~\ref{fig:arch}. 

\vspace{1ex}\noindent\textbf{Overall intuition.}
From a technical perspective, the main idea behind our method is to learn a map that directly translates eigenvalues to 3D coordinates. However, the eigenvalues are a mixture of a {\em global} spectrum (computed from the classical Laplacian) and a {\em local} spectrum (computed from a Laplace-like operator localized on a region); crucially, the learnable map does {\em not} know the region used for computing the local spectrum. This has two important consequences:
\begin{itemize}
    \item Since the map is required to reconstruct the entire shape as accurately as possible, it must learn on its own that the local spectrum encodes geometric details of some surface region, while the global spectrum encodes the overall geometry.
    \item By seeing global and local spectra jointly, the network learns how the two interact; namely, it learns to associate changes in the local spectrum with changes in the global one.
\end{itemize}
In other words, the network learns (i) correlations between eigenvalues and geometric features, and (ii) correlations between global and local spectra.

Both properties have a practical impact. Property (i) allows us to use the eigenvalues directly as a parametric encoding of 3D shapes, without the need to learn a new representation as done with the autoencoder paradigm \cite{instant2020}. For this reason, we can adopt a simple decoder-only architecture. Further, eigenvalues are interpretable due to their analogy with Fourier analysis, follow a natural ordering, are easy to compute, and are robust to discretization. Property (ii), on the other hand, allows to recover statistical correlations between local details and global shapes. This enables new paradigms for shape modeling, as shown in Figure~\ref{fig:teaser} and several other examples in Section~\ref{sec:results}.

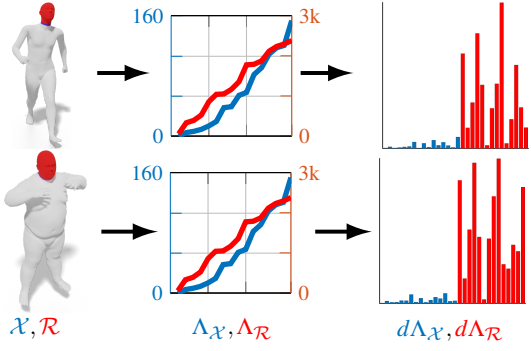
\begin{figure}[t!]
    \begin{tabular}{ccc}
    
    \begin{minipage}{0.25\linewidth}
    \linethickness{1.5pt}
        \begin{overpic}
        [trim=0cm 0cm 0cm 0cm,clip,width=0.95\linewidth]{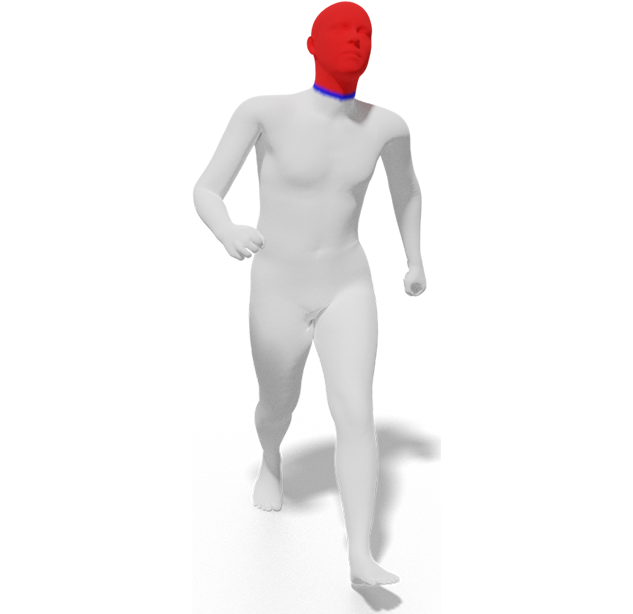}
        \put(87,50){\color{black}\vector(1,0){35}}
        \put(222,50){\color{black}\vector(1,0){35}}
        \end{overpic}
    \end{minipage}
    &

        \hspace{-0.5cm}
        
        \begin{minipage}{0.2\linewidth}
        
         \footnotesize
%
%
\definecolor{mycolor1}{rgb}{0.00000,0.44700,0.74100}%
\definecolor{mycolor2}{rgb}{0.85000,0.32500,0.09800}%
\begin{tikzpicture}
\pgfplotsset{set layers}
\begin{axis}[%
width=0.95\linewidth,
height=0.95\linewidth,
at={(0.758in,0.481in)},
scale only axis,
xmin=0,
xmax=16,
separate axis lines,
 axis y line*=left,
every outer y axis line/.append style={mycolor1},
every y tick label/.append style={font=\color{mycolor1}},
every y tick/.append style={mycolor1},
xtick={40, 5, 10, 16},
ytick={0, 53, 106, 160},
yticklabels={0,  , ,160 },
xticklabels={0,  , , },
xmajorgrids,
ymajorgrids,
ymin=0,
ymax=160,
axis background/.style={fill=white},
yticklabel pos=right
]
\addplot [color=mycolor1, line width=2.0pt, forget plot]
  table[row sep=crcr]{%
1	1.51470208554144e-14\\
2	4.67324998690563\\
3	6.28553022513816\\
4	8.92314895342906\\
5	13.5094556203005\\
6	19.3677470904578\\
7	37.7995273321811\\
8	39.0020732905495\\
9	54.0679812160388\\
10	57.9332366783737\\
11	81.8070347499671\\
12	90.803140655867\\
13	108.840159894013\\
14	117.337296664585\\
15	120.477082110534\\
16	153.444434081014\\
};
\end{axis}
\begin{axis}[%
width=0.95\linewidth,
height=0.95\linewidth,
at={(0.758in,0.481in)},
scale only axis,
 axis y line*=right,
xmin=0,
xmax=16,
every outer y axis line/.append style={mycolor2},
every y tick label/.append style={font=\color{mycolor2}},
every y tick/.append style={mycolor2},
xtick={40, 5, 10, 16},
ytick={0, 1000, 2000, 3000},
yticklabels={0,  , ,3k },
xticklabels={,  , , },
ymin=0,
ymax=3000,
axis background/.style={fill=white},
yticklabel pos=right
]
\addplot [color=red, line width=2.0pt, forget plot]
  table[row sep=crcr]{%
1	56.6761802907672\\
2	331.030608665176\\
3	390.771161053598\\
4	523.620146217781\\
5	856.26794511607\\
6	1039.98966678266\\
7	1049.08032097702\\
8	1164.88066739763\\
9	1352.02576706253\\
10	1773.76142771419\\
11	1788.41760542974\\
12	1872.82040888071\\
13	2114.14394761951\\
14	2232.68810530213\\
15	2293.16042215152\\
16	2378.19556629161\\
};
\end{axis}
\end{tikzpicture}%
        \end{minipage}

    &

    \hspace{1.2cm}
    \begin{minipage}{0.23\linewidth}
        \begin{overpic}
        [trim=0cm 0cm 0cm 0cm,clip,width=1\linewidth]{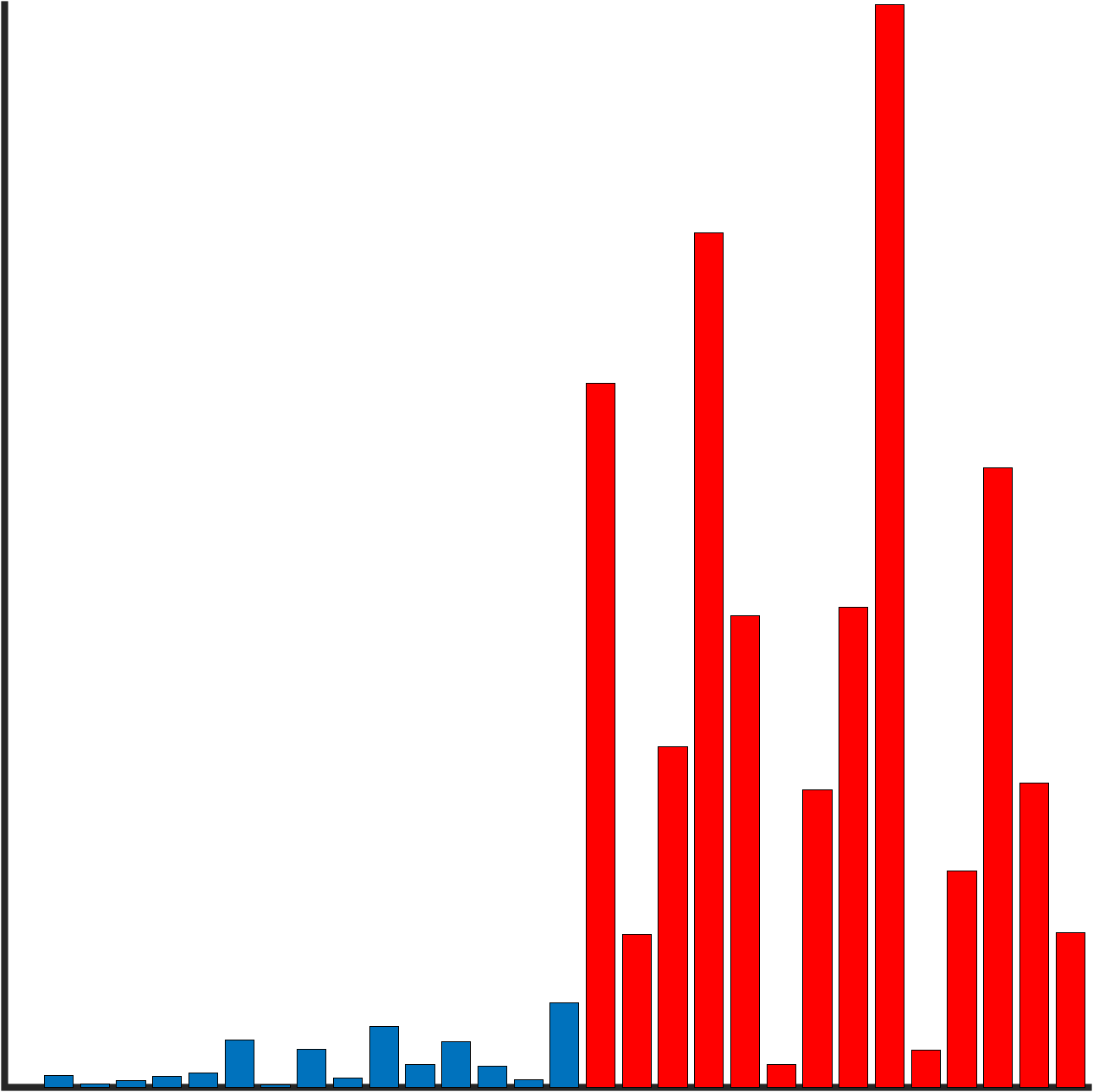}
        \put(25,25){}
        \end{overpic}
    \end{minipage}

    \\

      \begin{minipage}{0.13\linewidth}  
        \linethickness{1.5pt}
        \begin{overpic}
        [trim=0cm 0cm 0cm 0cm,clip,width=1\linewidth]{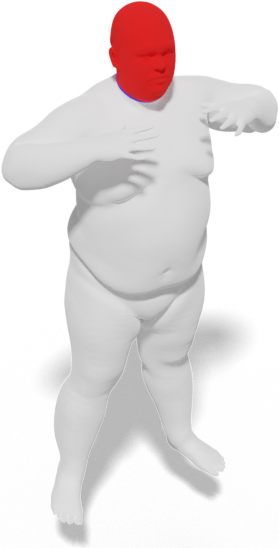}
        \put(3,-13.5){$\ourblue{\X}, \red{\subX}$}
        \put(116,-13.5){$\ourblue{\Lambda_{\X}}, \red{\Lambda_{\subX}}$}
        \put(60,50){\color{black}\vector(1,0){35}}
        \put(192,50){\color{black}\vector(1,0){35}}
        \end{overpic}
        \end{minipage}
&
        \vspace{1cm}
        \vspace{-1.6cm}
        \hspace{-0.6cm}
        
        \begin{minipage}{0.2\linewidth}
        \hspace{-1cm}
        \footnotesize
%
%
\definecolor{mycolor1}{rgb}{0.00000,0.44700,0.74100}%
\definecolor{mycolor2}{rgb}{0.85000,0.32500,0.09800}%
\begin{tikzpicture}
\pgfplotsset{set layers}
\begin{axis}[%
width=0.95\linewidth,
height=0.95\linewidth,
at={(0.758in,0.481in)},
scale only axis,
xmin=0,
xmax=16,
separate axis lines,
 axis y line*=left,
every outer y axis line/.append style={mycolor1},
every y tick label/.append style={font=\color{mycolor1}},
every y tick/.append style={mycolor1},
xtick={40, 5, 10, 16},
ytick={0, 53, 106, 160},
yticklabels={0,  , ,160 },
xticklabels={0,  , , },
xmajorgrids,
ymajorgrids,
ymin=0,
ymax=160,
axis background/.style={fill=white},
yticklabel pos=right
]
\addplot [color=mycolor1, line width=2.0pt, forget plot]
  table[row sep=crcr]{%
1	1.51470208554144e-14\\
2	4.67324998690563\\
3	6.28553022513816\\
4	8.92314895342906\\
5	13.5094556203005\\
6	19.3677470904578\\
7	37.7995273321811\\
8	39.0020732905495\\
9	54.0679812160388\\
10	57.9332366783737\\
11	81.8070347499671\\
12	90.803140655867\\
13	108.840159894013\\
14	117.337296664585\\
15	120.477082110534\\
16	153.444434081014\\
};
\end{axis}
\begin{axis}[%
width=0.95\linewidth,
height=0.95\linewidth,
at={(0.758in,0.481in)},
scale only axis,
 axis y line*=right,
xmin=0,
xmax=16,
every outer y axis line/.append style={mycolor2},
every y tick label/.append style={font=\color{mycolor2}},
every y tick/.append style={mycolor2},
xtick={40, 5, 10, 16},
ytick={0, 1000, 2000, 3000},
yticklabels={0,  , ,3k },
xticklabels={,  , , },
ymin=0,
ymax=3000,
axis background/.style={fill=white},
yticklabel pos=right
]
\addplot [color=red, line width=2.0pt, forget plot]
  table[row sep=crcr]{%
1	56.6761802907672\\
2	331.030608665176\\
3	390.771161053598\\
4	523.620146217781\\
5	856.26794511607\\
6	1039.98966678266\\
7	1049.08032097702\\
8	1164.88066739763\\
9	1352.02576706253\\
10	1773.76142771419\\
11	1788.41760542974\\
12	1872.82040888071\\
13	2114.14394761951\\
14	2232.68810530213\\
15	2293.16042215152\\
16	2378.19556629161\\
};
\end{axis}
\end{tikzpicture}%
        \end{minipage}
&
        \hspace{1.2cm}
        
        \begin{minipage}{0.23\linewidth}

        \begin{overpic}
        [trim=0cm 0cm 0cm 0cm,clip,width=1\linewidth]{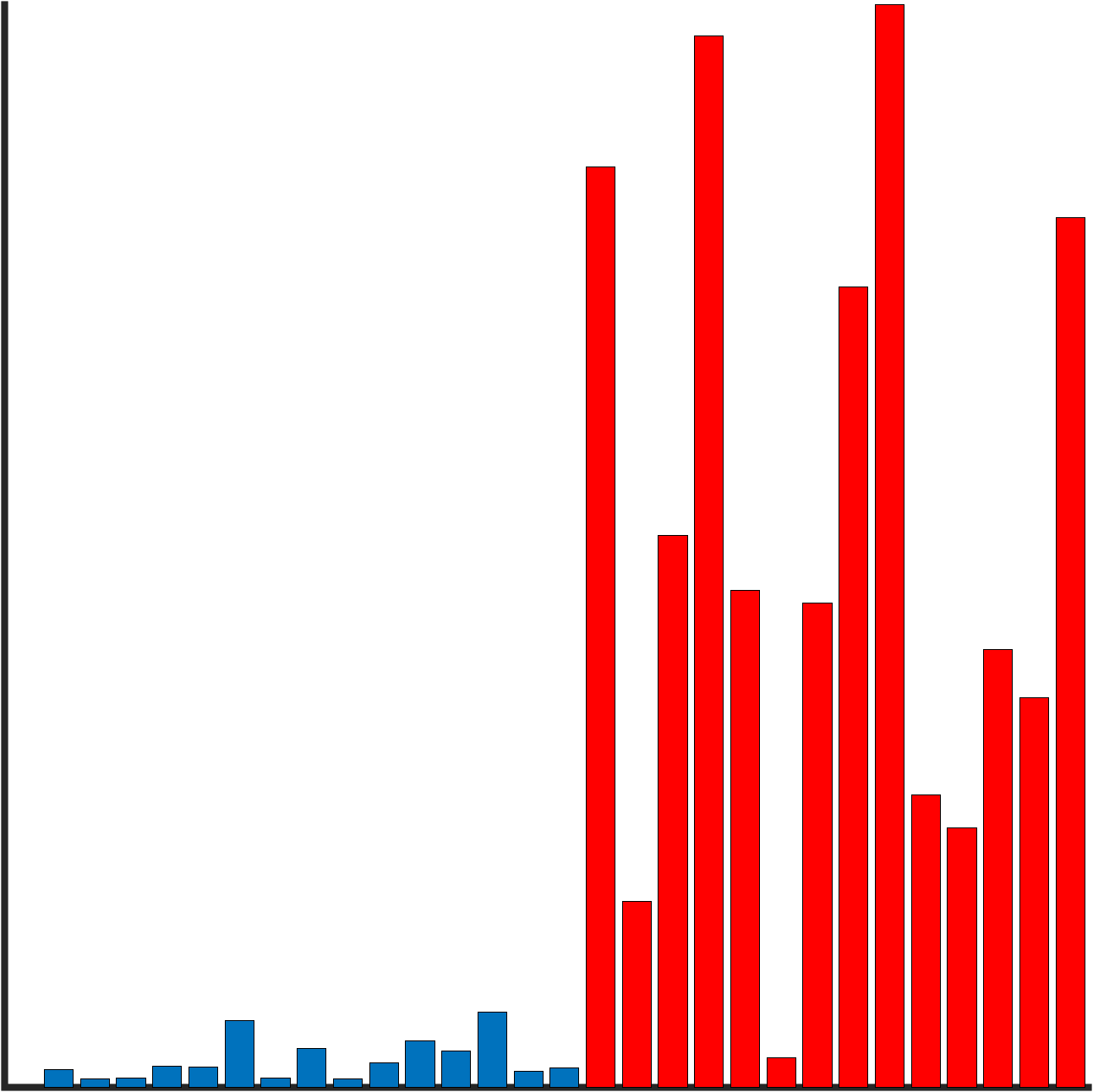}
        \put(12,-22){$\ourblue{d\Lambda_{\X}}, \red{d\Lambda_{\subX}}$}
        \end{overpic}
    \end{minipage}
    \vspace{1.0cm}

    \end{tabular}

\caption{\label{fig:spec} Two examples of the computation of the proposed {spectral encoding}. {\em Left}: Starting shapes and regions (in red); {\em Middle}: Global and local spectra, in blue and red respectively; {\em Right}: Bar plots representing the differences of Eq.~\eqref{eq:senc}, which compose our encoding.}
\end{figure}

\subsection{Proposed model}
\label{sec:model}
Given an input shape $\X$, our pipeline involves the computation of a local spectrum on a given region $\subX \subset \X$. Here we detail the selection criteria for the region, and the choice of a localized operator. 
%
%

\vspace{1ex}\noindent\textbf{Local region.}
The region is selected to be informative for the final task, and the choice should be coherent across all shapes in the training set. For example, if the application expects the user to modify facial features, then the human head should be included as a region of interest in the training data.
%
Well-established segmentation approaches may be used to select $\subX$ automatically, e.g. for man-made objects such as airplanes~\cite{Kalogerakis,sharp2020diffusion,bic17, Kleiman19} or for organic shapes such as humans (e.g., see the head extractor proposed in~\cite{FARM}).

\vspace{1ex}\noindent\textbf{Localized operator.}
Once a region $\subX \subset \X$ has been identified, we compute a localized operator over it and, in turn, its truncated spectrum $\Lambda_{\subX}$.
Perhaps the most natural choice is to disconnect the region from the rest of the shape, and compute the standard Laplacian on the resulting surface with boundary conditions; we refer to this choice as \textbf{PAT}. Other possibilities include the definition of a Hamiltonian operator with a sharp potential~\cite{Choukroun} (\textbf{HAM}), and the localized manifold harmonics~\cite{LMH}, which yields a Hamiltonian-like operator whose eigenfunctions are orthogonal to the Laplacian eigenbasis (\textbf{LMH}). In Section~\ref{sec:eval} we compare these choices experimentally. 


\vspace{1ex}\noindent\textbf{Spectral encoding.}
Given the (truncated) global spectrum $\Lambda_{\X} \in \mathbb{R}^{k}$, sorted non-decreasingly, we first compute the differences:
\begin{align}
    d\lambda_{\ell} = \lambda_{\ell} - \lambda_{\ell-1}, \ \forall \ell \in {2, \ldots , k}\,,
    \label{eq:difference}
\end{align}
where $d\lambda_{\ell} \geq 0, \ \forall \ell$, and store them in a vector:
\begin{equation}
    d\Lambda_{\X} = [d\lambda_{2}, \ldots , d\lambda_{k}] \in \mathbb{R}^{k-1}\,.
\end{equation}
We do the same for the local spectrum $\Lambda_{\subX} \in \mathbb{R}^{h}$, 
obtaining $d\Lambda_{\subX} \in \mathbb{R}^{h-1}$.
Finally, we simply concatenate the two vectors $d\Lambda_{\X}$ and $d\Lambda_{\subX}$ to generate our \emph{spectral encoding}:
\begin{equation}\label{eq:senc}
   d\Lambda = [d\Lambda_{\X}\,;\, d\Lambda_{\subX}] \in \mathbb{R}^{(k-1)+(h-1)}  \,.
\end{equation}
An illustration of this computation is depicted in Fig.~\ref{fig:spec}, and is represented in green in the middle of Fig.~\ref{fig:arch}. 
This encoding exploits the natural hierarchy carried by each set of eigenvalues. We observed that using differences between subsequent eigenvalues, in place of their absolute values, has a regularizing effect that helps training more effectively.
\revision{Taking differences does not disrupt the geometric information encoded in the spectra, and can be effectively used by the network to recover the original eigenvalues if needed.}

\vspace{1ex}\noindent\textbf{Map training.}
We aim to learn the map $\Pi$, which receives as input a spectral encoding $d\Lambda$, and outputs a $3$D shape that corresponds to that encoding. 
Given a collection of training shapes $\{\X_i \}$ from a given class, 
we compute for each of them the {spectral encoding} $\{d\Lambda_i \}$ 
following the process described above. 
We implement the map $\Pi$ as a fully-connected decoder, and train it to minimize a standard reconstruction loss:
\begin{equation}
\label{eqn:loss}
    Loss = 
    \sum_{i} 
    \|\Pi(d\Lambda_i) - X_i \|^{2}_{F} \,.
\end{equation}
When we deal with point clouds, we replace the Frobenius norm with the Chamfer distance defined in \cite{achlioptas2018learning}.
%
%
We remark that while local regions are involved in the computation of the spectral encoding, we are not using any specific loss to guide the reconstruction of the corresponding local geometry. 

\vspace{1ex}\noindent\textbf{Modeling at test time.}
Once the model is trained, one can feed the network a previously unseen {spectral encoding} $d\Lambda$. The resulting $3$D shape exhibits the geometric details encoded in $d\Lambda$, but with the discretization of the training set. One can also compose global and local encodings from different shapes, interpolate the encodings, or perform other operations as shown in the next Section.

\vspace{1ex}\noindent\textbf{Network architecture.}
The proposed network is composed of 4 fully connected layers. We refer to the supplementary materials for further details.

\section{Results}
\label{sec:results}

\subsection{Datasets}
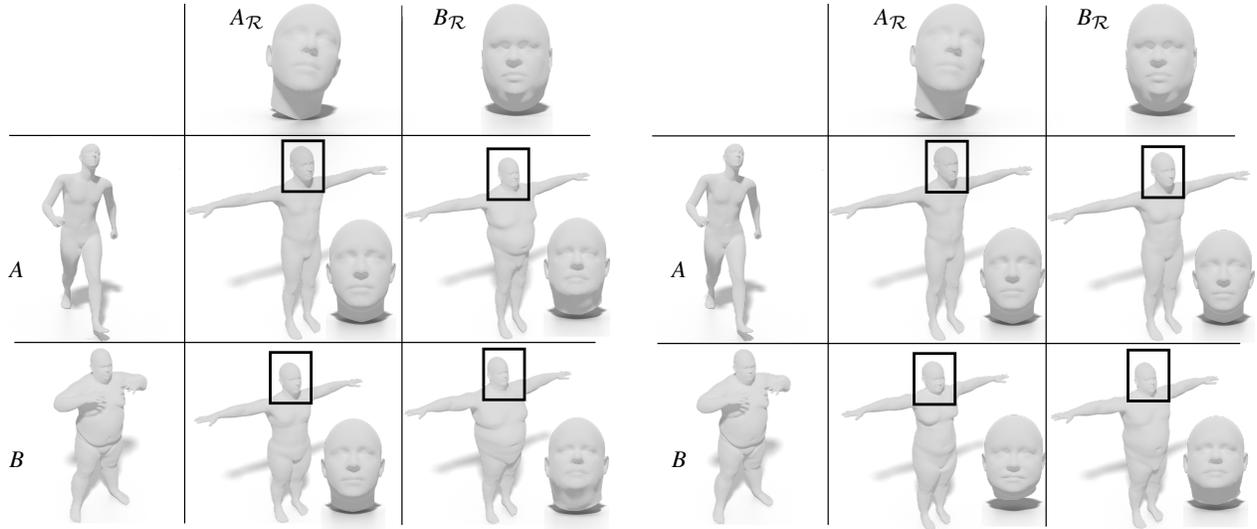
\begin{figure*}[!t]
\begin{tabular}{cc}
\begin{minipage}{0.5\linewidth}
\begin{center}
        \begin{overpic}
        [trim=0cm 0cm 0cm 0cm,clip,width=0.95\linewidth]{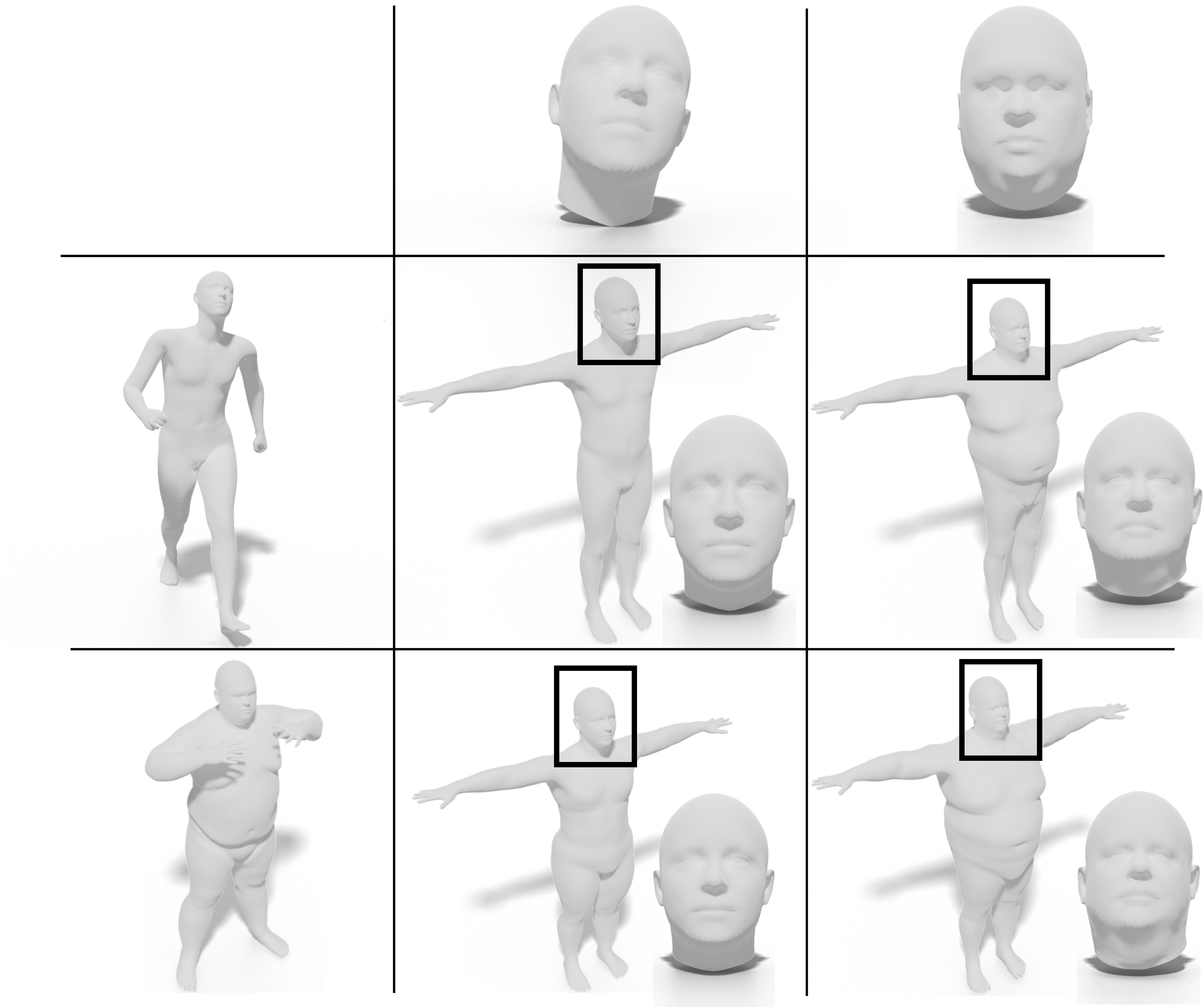}
        \put(5,40){$A$}
        \put(5,10){$B$}
        \put(40,80){$A_\mathcal{R}$}
        \put(72,80){$B_\mathcal{R}$}
        \end{overpic}

\end{center}
\end{minipage}
&
\hspace{-0.8cm}
\begin{minipage}{0.5\linewidth}
\begin{center}
        \begin{overpic}
        [trim=0cm 0cm 0cm 0cm,clip,width=0.95\linewidth]{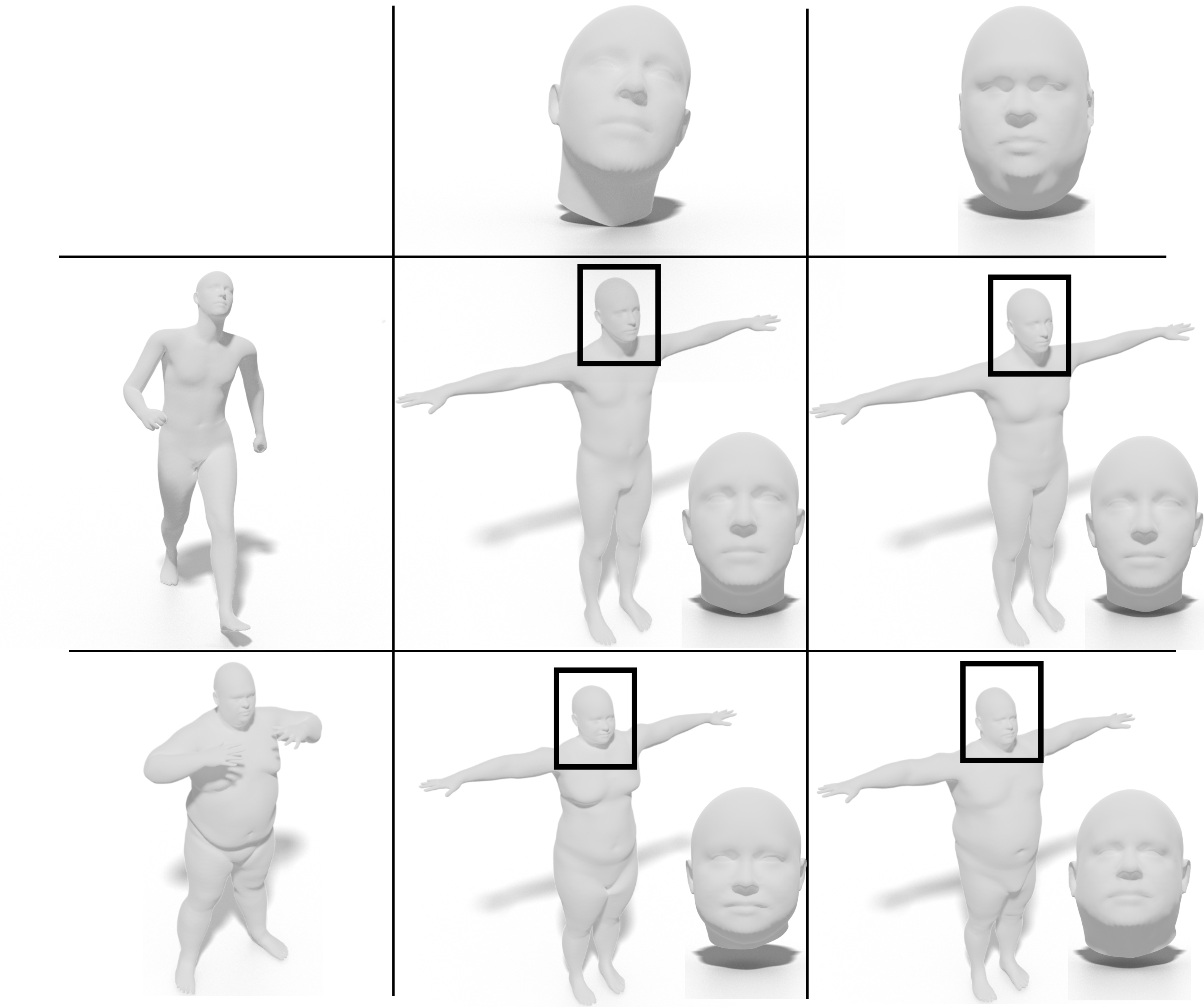}
        \put(8,40){$A$}
        \put(8,10){$B$}
        \put(40,80){$A_\mathcal{R}$}
        \put(72,80){$B_\mathcal{R}$}
        \end{overpic}

\end{center}
\end{minipage}
\end{tabular}
    \caption{On the left, semantic swap experiment with PAT15+15. On the right, the same using LBO30.}
    \label{fig:SurrealSWAP}
\end{figure*}
\setlength{\columnsep}{1pt}
\setlength{\intextsep}{1pt}

\vspace{1ex}\noindent\textbf{\em CUBE.} 
\begin{wrapfigure}[3]{r}{0.25\linewidth}
\vspace{-0.35cm}
\begin{center}
 \begin{overpic}
        [trim=0cm 0cm 0cm 0cm,clip,width=0.7\linewidth]{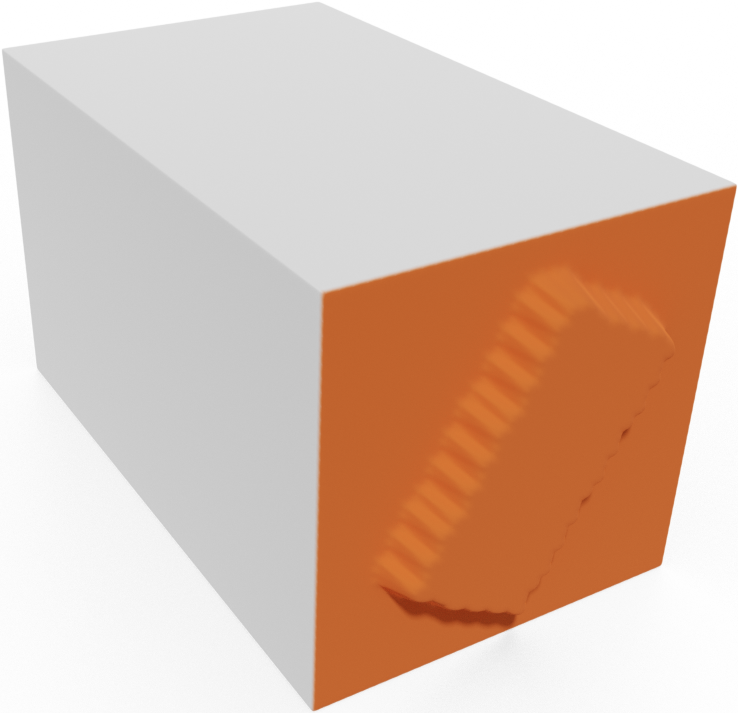}
        \put(100,25){\footnotesize $\textcolor[rgb]{0.941176470588235,0.501960784313726,0.501960784313726}{\mathcal{R}}$}
    \end{overpic}
\end{center}
\end{wrapfigure}
This is a synthetic dataset comprising $1000$ cube meshes with $7350$ vertices each. Each cube shows one of $125$ different patterns on its front face, and has a depth picked at random from $8$ possible values (see inset for an example). These two factors of variation are uncorrelated and provide a controlled setup for our tests. 
As region $\subX$ on each of these samples, we select the front face since this is where we apply the local variations. 

\vspace{1ex}\noindent\textbf{\em SURREAL.} To challenge our model on more realistic data, we collect $2337$ human shapes from SURREAL~\cite{varol17_surreal}.
As $\subX$ on these human bodies, we choose the head for three reasons: 1) it encodes the identity characteristics; 2) it is unique in the body and hard to confuse with other body parts; 3) in contrast with cubes, where pattern and depth are uncorrelated, head and body tend to correlate. We are interested in verifying how this impacts our learning process.

\vspace{1ex}\noindent\textbf{\em SMAL.} We test our model with another example of a realistic dataset: SMAL\cite{SMAL}, a dataset of 3D mesh animals generated by a morphable model learned by scanning toy figurines.
As a dataset, we choose 4872 animals belonging to five different classes: tiger, wolf, cow, zebra, hippo. 
Each mesh has 3889 vertices and represents the animal in the rest pose. 
As $\subX$ we consider the head for the same reasons as above. 
Like SURREAL, SMAL allows us to test the model in a realistic setting where each local variation may correlate with the rest of the body. Moreover, SMAL is composed of multiple kinds of animals with distinctive and different features. The higher diversity allows us to discriminate better the different contribution of the local and global spectra during reconstruction.

\vspace{1ex}\noindent\textbf{\em AIRPLANES.} The above datasets have a common discretization. To test if our model can discover the underlying relation between spectrum and geometry even in more general cases, we selected $448$ airplanes from ShapeNet~\cite{yi2017large} and sampled $500$ points from each of them, producing unordered point clouds without known correspondence. As $\subX$ we selected the tail segment.

We report complete details about the datasets we used in the supplementary materials.
\revision{If not differently stated, the shapes we adopt in all our experiments and figures have never been seen during training, and belong to the test set or a completely different dataset.}

\subsection{Shape modeling}

Here and throughout the manuscript, we adopt the notation PAT$_{\mathcal{R}}$15+15 to denote the spectral encoding composed of 15 eigenvalues of the standard (global) Laplacian and 15 eigenvalues of the local PAT operator defined by the region $\mathcal{R}$. Other choices of the operator, local region, and dimensions of the spectral encoding follow the same notation. When the region we are considering is clear and unique, we remove the subscript $\mathcal{R}$, and we only write PAT15+15. As the main baseline, we consider LBO30, which corresponds to the encoding provided by the first 30 eigenvalues of the global Laplacian.

 \begin{figure}
    \centering
    \begin{overpic}
        [trim=0cm 0cm 0cm 0cm,clip,width=1\linewidth]{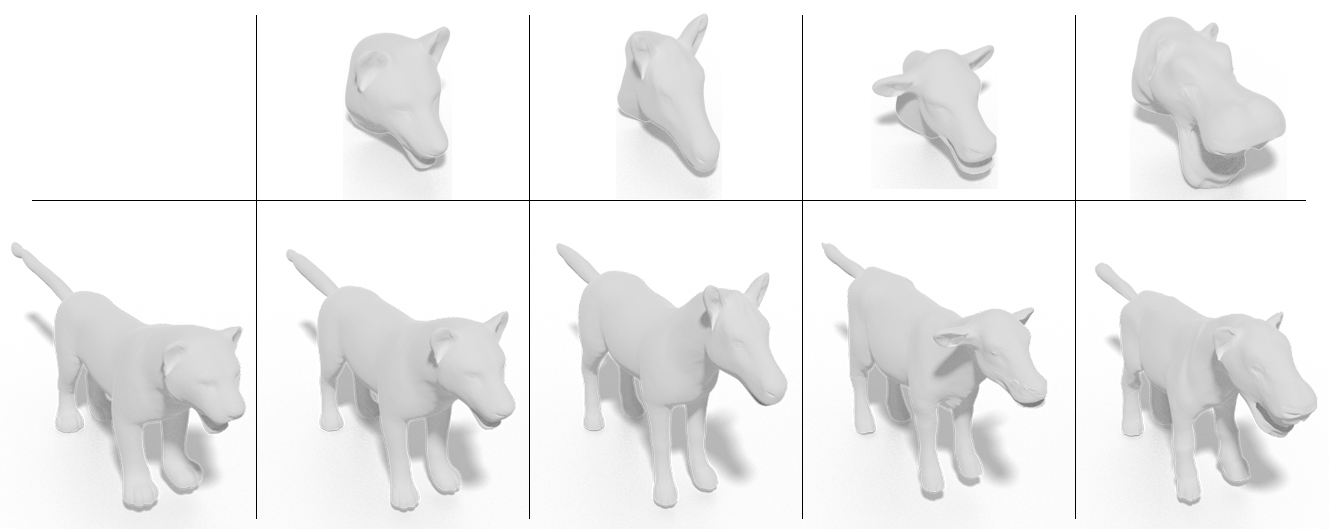}
        \end{overpic}

    
    \caption{Shapes obtained by combining the global spectrum of a tiger with the local spectra of the heads of different animals (one per column).  Observe the change in body shapes induced by the different heads; in the last column, the body does not change much, but the head has an intermediate shape between tiger and hippo. The operator used in these tests is PAT$15+15$. 
    }
    \label{fig:SMALswapTIGER}
\end{figure}

\begin{figure*}[t!]
            \begin{overpic}
            [trim=0cm 0cm 0cm 0cm,clip,width=0.99\linewidth]{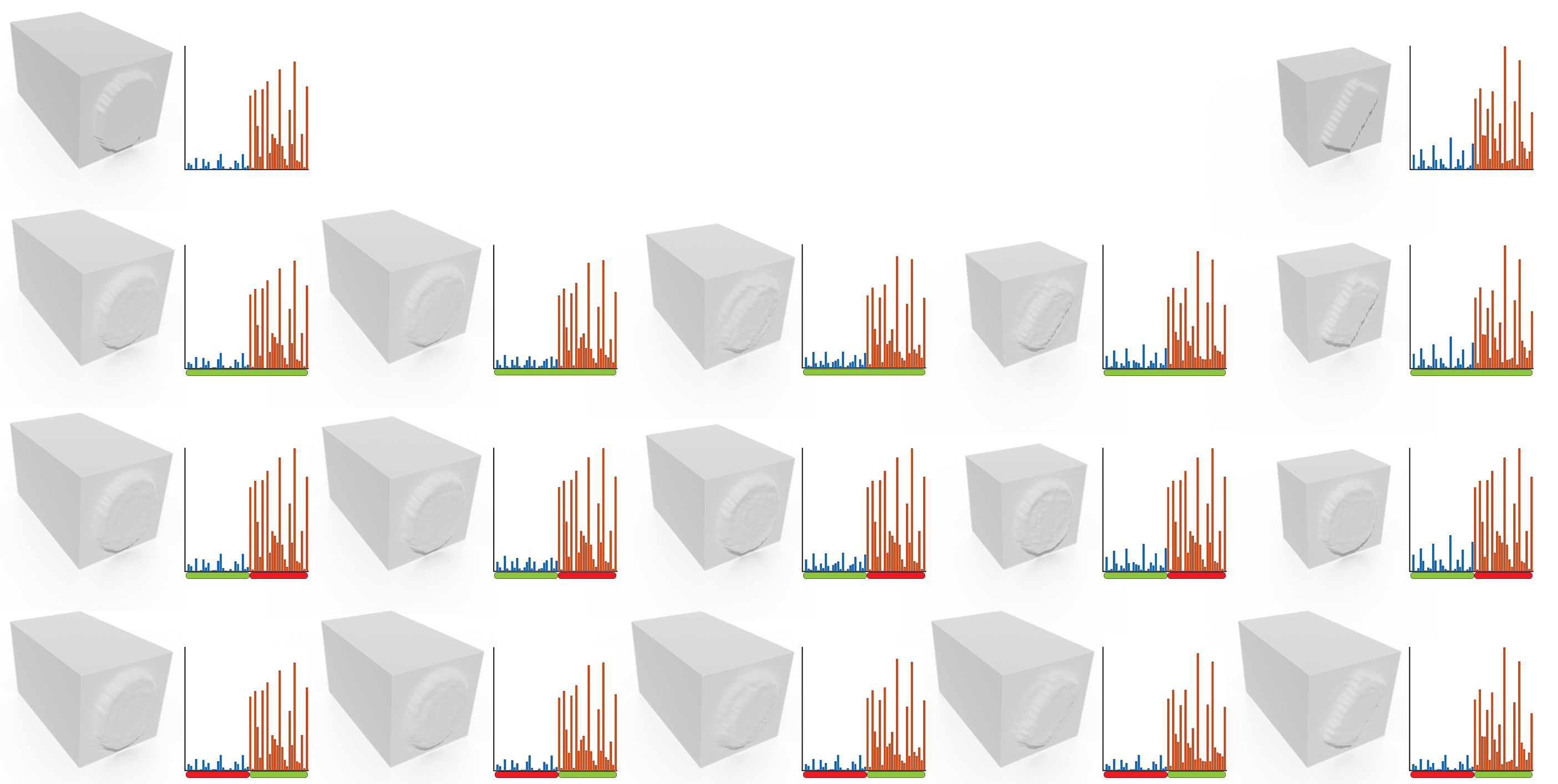}
            \put(1,37){$d\Lambda$}
            \put(1,24.5){$d\Lambda_{\X}$}
            \put(1,11.5){ $d\Lambda_{\subX}$}
            \put(3,50){Start}
            \put(83,50){End}
            \end{overpic}
        \caption{Interpolation results for a pair of cubes: \revision{ the left one is taken from the test set, while the rightmost from the train set}. We plot the spectral encoding associated with each shape as a bar plot; we highlight in green the part of the encoding that we are interpolating in each row, and in red the values that we keep fixed.
        {\em First row}: Two input cubes.
        {\em Second row}: Interpolation of the entire encoding. 
        {\em Third row}: Interpolation of the global part only (blue in the bar plots); observe how the pattern on the front face does not change, while the volume of the entire cube is correctly interpolated.
        {\em Last row}: Interpolation of the local part of the encoding (red in the bar plots), inducing a change in the front pattern only.}
        \label{fig:interpolationCUBEplot}
    \end{figure*}
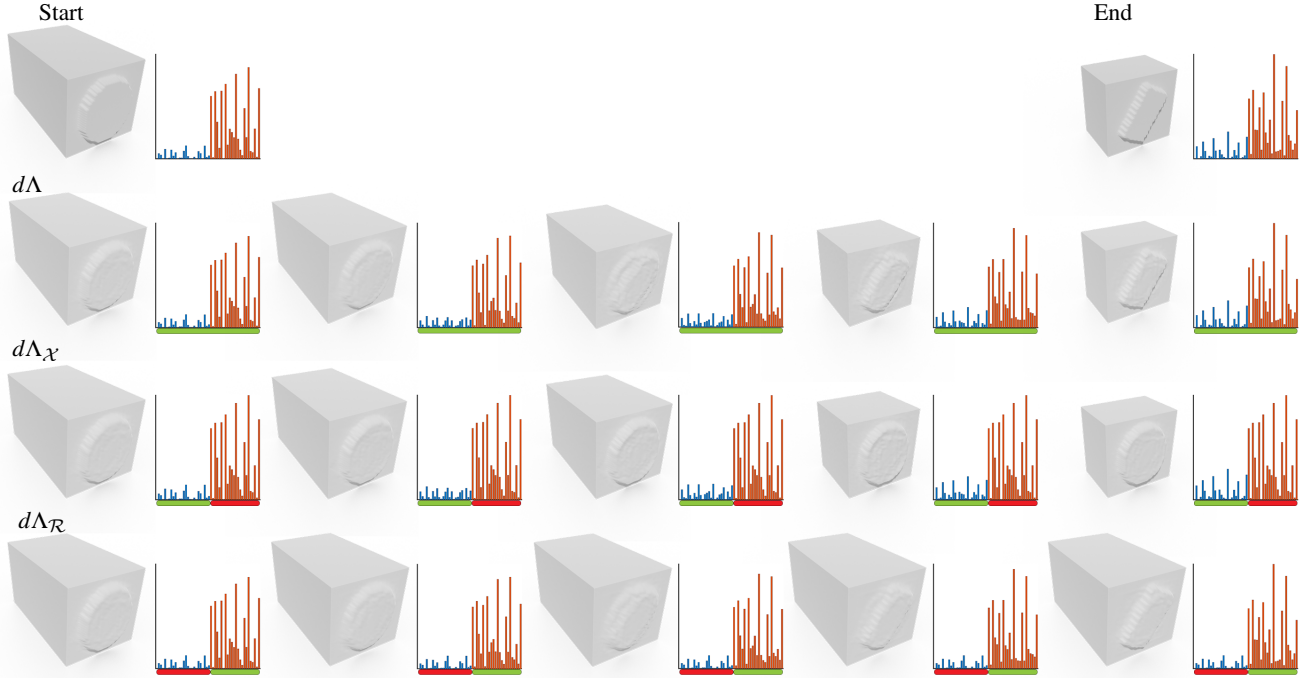
    
\begin{figure}
    \centering
    \begin{overpic}
        [trim=0cm 0cm 0cm 0cm,clip,width=1\linewidth, grid=false]{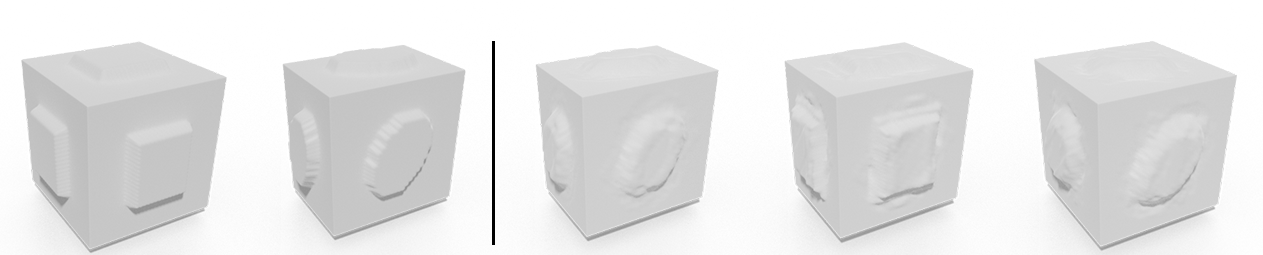}
        \put(47,18){$d\Lambda$}
        \put(67,18){$d\Lambda_{\X}$}
        \put(87,18){$d\Lambda_{\subX}$}
        \put(6,18){Start}
        \put(27,18){End}
        \end{overpic}
    
    \caption{\label{fig:interpolationCUBErLocal} \revision{Final results of the interpolation from Start to End. On the right side we show the interpolation of the entire encoding ($d\Lambda$), of the global part only ($d\Lambda_{\X}$) and of the local part only ($d\Lambda_{\subX}$).}
    }
    
\end{figure}

\begin{figure*}[!t]
\begin{center}
        \begin{overpic}
        [trim=0cm 0cm 0cm 0cm,clip,width=0.8\linewidth]{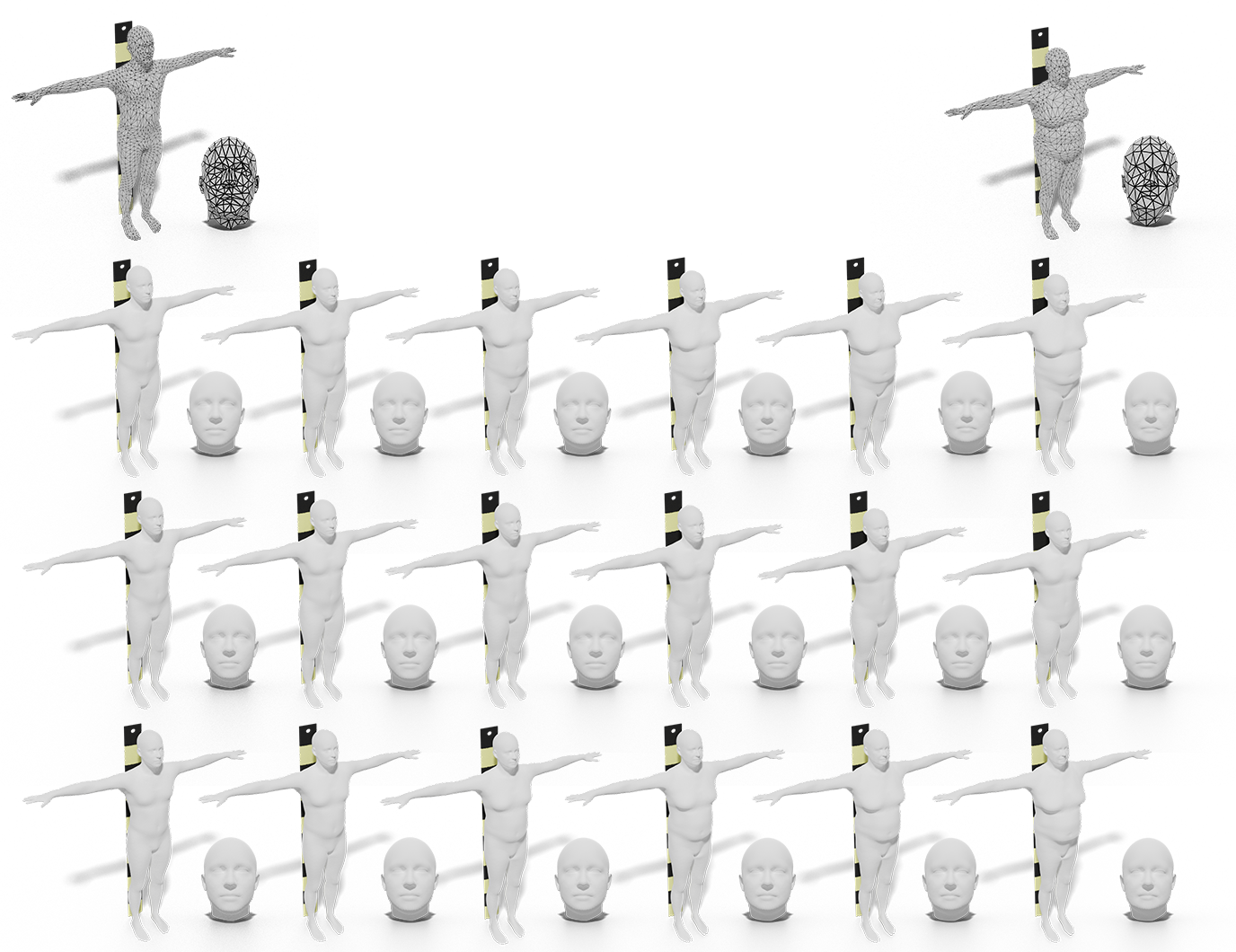}
        \put(3,54){$d\Lambda$}
        \put(3,35){$d\Lambda_{\X}$}
        \put(3,16){$d\Lambda_{\subX}$}
        \put(3,74){Start}
        \put(78,73.5){End}
        \end{overpic}

\end{center}
    \caption{Different interpolations between two inputs from human shapes (first row): Global+Local ($d\Lambda$-second row), only Global($d\Lambda_{\X}$-third row) and only Local($d\Lambda_{\subX}$-fourth row). We add a ruler behind each shape to emphasize the height variation.}
    \label{fig:interpolationSURREAL}
\end{figure*}

\noindent\textbf{Semantic swap.} 
To better present the impact of our method in shape modeling applications, we provide qualitative examples in different contexts.
We start by showing how our method allows to recombine the encoding of different objects to generate novel shapes with natural coherence. 

In Figure \ref{fig:SurrealSWAP} we show an example on humans, where we consider two shapes, namely $A$ and $B$, from different datasets(\cite{varol17_surreal} and \cite{Dyna2015}, respectively), and fix for both the local region on the head, respectively denoted as $A_{\mathcal{R}}$ and $B_{\mathcal{R}}$. On the left we use the map recovered for the PAT15+15 input, whereas on the right we use the map obtained from the standard LBO30 input. On the main diagonal of each grid, we report the reconstruction results from the original encoding of the two shapes. We notice that both approaches return reliable results, generalizing to datasets unseen at training time (such as $B$). 

The off-diagonal entries show the results produced by mixing the spectra of the two inputs. In the top right, we concatenated the first $15$ values of the encoding of $A$ with the last $15$ of the encoding of $B$. Both PAT15+15 and LBO30 start with the first $15$ eigenvalues of the LBO from A  while for PAT15+15 the second part is given by the first 15 eigenvalues of the localized operator on the region $B_\mathcal{R}$ region, while for LBO30 the eigenvalues from $16$ to $30$ of the LBO of B. In the bottom left, the same with the inverse role of B and A. 
We notice that PAT15+15 succeeded to produce meaningful modifications but keeping an overall coherence. Injecting into the network the spectra of $A$ and the one of $B_\mathcal{R}$ produces a man with similar height and proportions of $A$, but more robust (while not as robust as $B$). Using the head of $A_\mathcal{R}$ on the body of $B$ has the effect to obtain similar proportions to $B$ (producing a shorter person), but respecting the semantics suggested by $A_\mathcal{R}$, which does not suggest a robust human. On the contrary, the grid on the right shows that working with the LBO30 representation does not provide the same level of control.

A clearer example is depicted in \figref{fig:SMALswapTIGER}. We take the global spectrum of a tiger and combine it with the local spectrum (one per column) of all the others animals in the dataset.
As before, when we change the local spectrum, the identity of the shape changes. The most peculiar instance is the combination between the global spectrum of the tiger and the local spectrum of the hippo (last column). 
The resulting mesh has ears similar to the tiger, but has a snout different from both animals. This intermediate result may be due to the dominance of the global features of the tiger that prevent the snout to puff up like in the hippo. 
The rest of the body remains more similar to the tiger.


\noindent\textbf{Shape space exploration.}
Our encoding also allows to explore the space of shapes. We report what we believe is a significant example from the CUBE dataset. 
This dataset has, by construction, two factors of variation: the depth of the cube and the pattern on the front face. These two features have no statistical correlation and are thus fully disentangled by purpose.

In Figure \ref{fig:interpolationCUBEplot} we report the results obtained by interpolating the spectra of two different cubes (`Start' and `End' respectively). 
We interpolate the entire encoding (global+local, second row), only the global part (third row) or only the local part (last row). 
These tests suggest that the learned map has learned to disentangle between the factors of variations in this dataset, which would not be possible by using only a global spectrum. It would also be hard to obtain with a standard autoencoder architecture, unless a disentanglement technique is explicitly implemented.
\revision{
To see how our method can effectively learn correlations between local regions and the whole shapes, we train our network on a different setting where cubes have the same pattern applied to all the faces and the local region $\subX$ is a single face as before. On the right of Figure \ref{fig:interpolationCUBErLocal}, we show the final results of the interpolations from the `Start' to the `End' cube (depicted on the left).
The interpolation of the global encoding changes only the length of the cube without modifying the pattern on the faces, while the local interpolation changes all the patterns coherently but leaves the cube's depth unchanged. We report the full experiment in the supplementary materials.
These two experiments confirm the ability of our encoding to discover the correlation between local details and global variation if present (Figure \ref{fig:interpolationCUBErLocal}) and to preserve their independence if it is the case (Figure \ref{fig:interpolationCUBEplot}). 
A variation in the local encoding produces more or less localized changes according to the level of dependency between the selected region $\subX$ and the rest of the shape. In any case, the generated deformation, even if local, can induce a wider variation to maintain global consistency. 
}

A similar example is depicted in Figure \ref{fig:interpolationSURREAL}. In this example, the two shapes have different discretizations, emphasizing that our method is agnostic to them. We observe that the method discovered a relation between the local (head) region and the global region, as we expected for human data. Interpolating the global spectra change keeps the identity suggested by the head while changing the body proportion (the ruler behind each shape eases the comparison of the heights). Interpolating the head requires keeping the body dimension while changing the subject identity; it is worth noting that the final head is coherent with the target shape.

Finally, our method is efficient at inference time, enabling real-time shapes explorations. We attach a video of interactive navigation via an intuitive interface.

\noindent\textbf{\em Remark.} At training time, our network does {\em not} know the association between local spectrum and region, but it just sees 30 values without knowing where they originate from. Therefore, it effectively learns what values are responsible for local changes and what are those responsible for the global changes in geometry.
    
\noindent\textbf{Different discretizations and point clouds.}
In previous experiments, we require all the shapes in the training set to be in full point-to-point correspondence, which is of great help for the network to discover patterns in the data. 
To investigate if the relationship between the input eigenvalues and the 3D points is strong enough to arise even without providing point correspondences, we remove the mesh connectivity and consider a noisy scenario of unordered point clouds of airplanes. Differently from~\cite{instant2020}, which requires redesigning the encoder to handle unordered point clouds, our model does not require any modification.

\begin{figure}[!t]
\begin{center}
        \begin{overpic}
        [trim=0cm 0cm 0cm 0cm,clip,width=1\linewidth]{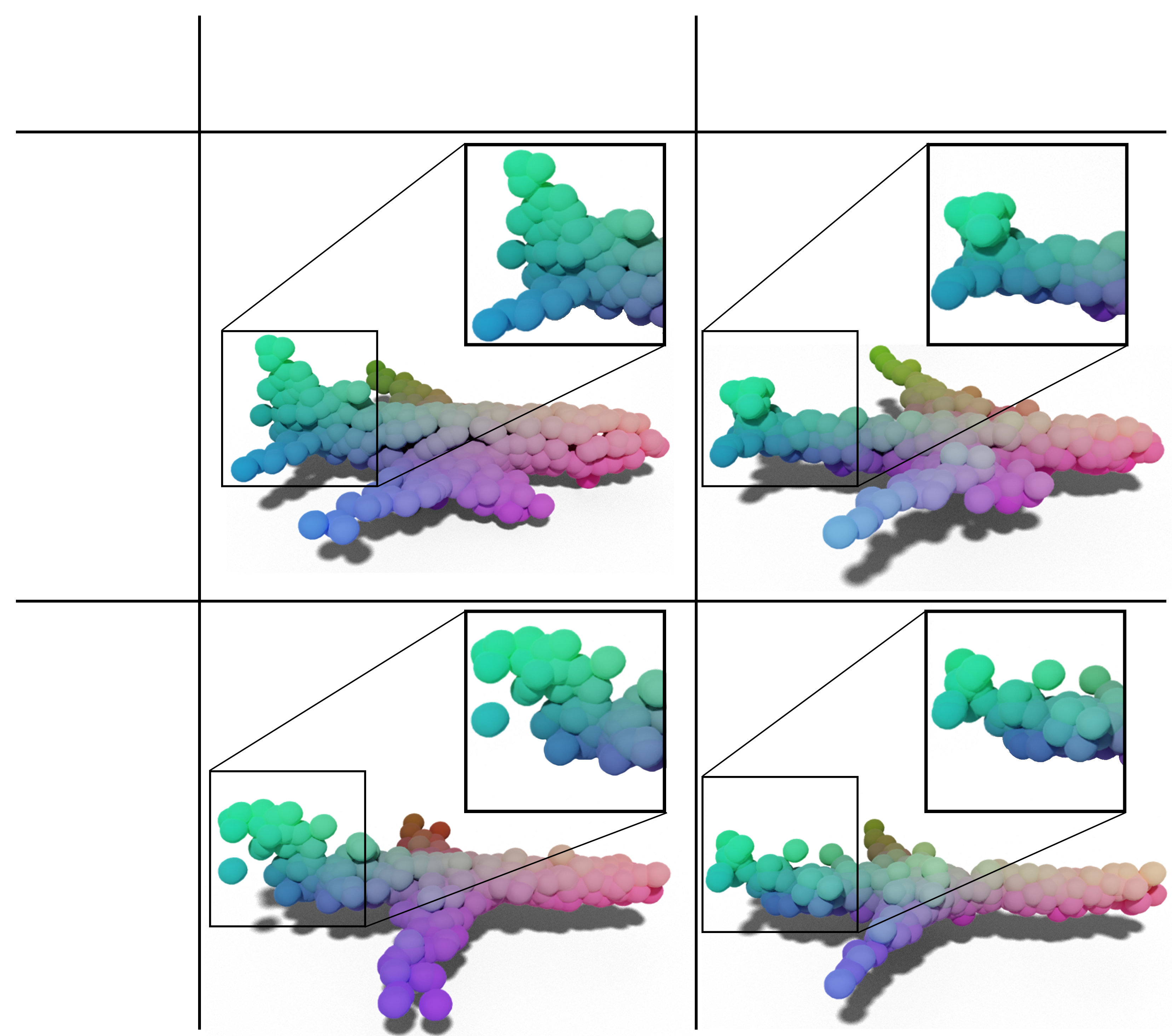}
        \put(10,58){\makebox[0pt]{\rotatebox[origin=c]{90}{$ \Lambda_{\X}^{min} $}}}
        \put(10,18){\makebox[0pt]{\rotatebox[origin=c]{90}{$ \Lambda_{\X}^{max} $}}}
        \put(40,80){\makebox[0pt]{$\Lambda_{\subX}^{min}$}}
        \put(80,80){\makebox[0pt]{$\Lambda_{\subX}^{max}$}}
        \end{overpic}

\end{center}
    \caption{Minimum and maximum spectra mixing on point clouds (more details in the text).}
    \label{fig:pointcloudSWAP}
\end{figure}
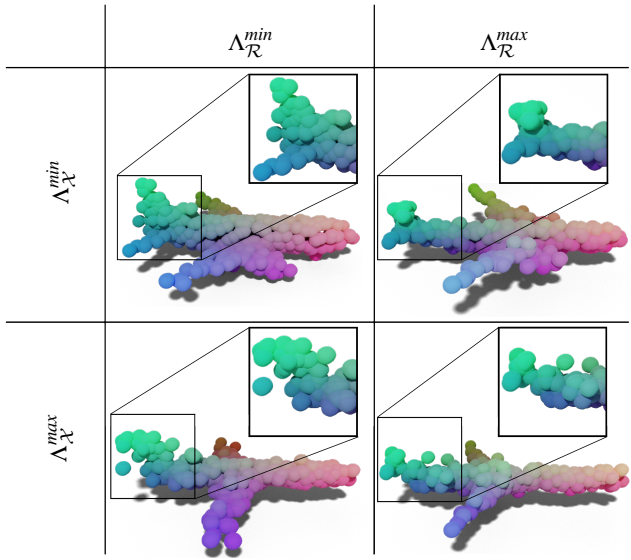

We considered the tail segment as the local region.
The tails of airplanes are interesting because their size and shape vary and identify specific categories (such as jet and passenger transport aircraft, among others).

In Fig.~\ref{fig:pointcloudSWAP}, we propose a spectra swap similar to Fig.~\ref{fig:SurrealSWAP}.
For all the point clouds in the AIRPLANES dataset, we consider the PAT15+15 encoding. For each of the 30 dimensions, we compute the minimum and the maximum among all the 448 shapes. Then we divide the sequence of 30 maxima and the sequence of the 30 minima into global (the first 15 values) and local information (the last 15). We refer to these four output vectors respectively as $\Lambda\mathbf{^{max}_{\X}}$, $\Lambda\mathbf{^{max}_{\subX}}$ from the maximum values, and $\Lambda\mathbf{^{min}_{\X}}$, $\Lambda\mathbf{^{min}_{\subX}}$ for the minimum ones. We denote the global values with $\X$ and the local ones with $\subX$.
We then compose 4 new spectra as the possible different combinations of the local and global parts. 
On the main diagonal, the generated airplanes are a large aircraft associated with the frequencies' minimum values (top left) and a thinner and longer jet for the maximum values (bottom right). As a first observation, this behavior is coherent with our expectations in terms of spectrum-geometry association. The two airplanes exhibit different empennage of the tails: the large aircraft has a conventional tail, while the jet has a T-tile. The kind of airplane determines the shape of the tail, and the coherence between the two would be critical for some applications.
What we observe in this case is that the global spectrum $\Lambda\mathbf{_{\X}}$ (rows) captures the type of airplane and determines the tail shape. Instead, by editing the local spectrum $\Lambda\mathbf{_{\subX}}$ (columns), we obtain a variation in tail dimensions, with slight modifications to airplane structure to adapt to this change, but without changing the class. 

We consider this a fascinating result because the network is trained {\em without} a point-to-point correspondence across the training shapes. The spectrum statistics are informative enough to relate spectrum and geometry through the unsupervised Chamfer distance. 
We report further examples of point clouds in the supplementary materials.

\section{Evaluation and analysis}\label{sec:eval}
Here we report our analysis of different elements of the method, providing justification of our choices and useful insights.

\subsection{Different local operators}
In Table~\ref{tab:dif_op}, we report the performance of our method under different choices of local operators. 
We consider three different error measures:
\begin{enumerate}
    \item MSE: mean squared error between the 3D coordinates of our reconstruction and the ground-truth, measured on the entire shape;
    \item MSE-$\subX$: mean squared error defined as above, measured only on the region $\subX$;
    \item Area: average difference between the area elements of each vertex of our reconstruction, and the corresponding area elements of the ground-truth shape; this measure quantifies the intrinsic metric distortion caused by the reconstruction module.  
\end{enumerate}
The values of MSE and MSE-$\subX$ in the table are all multiplied by 10e6, while the Area values are multiplied by 10e3.
We notice that PAT maintains the best performance in general, across different dataset and measures. We suppose that this is due to the following reason: PAT is the only operator that is fully localized on the region, because it treats the region as a disconnected component. The other two, HAM and LMH, have both global support and are computed by localizing the LBO by means of a scalar potential, which is known to cause leaking outside of the region~\cite{hamiltonian}. Moreover, LMH has an additional term of orthogonality that enforces its dependency on the LBO, thus potentially increasing redundancy and, in turn, reducing the amount of information that can be used for a faithful reconstruction.
%
\begin{table*}[t!]
\footnotesize
\centering

\begin{tabular}{@{\extracolsep{4pt}} k C C C C C C C C C}
\hline
    & \multicolumn{3}{c}{CUBE} & \multicolumn{3}{c}{SURREAL} & \multicolumn{3}{c}{SMAL} \\\cline{2-4}\cline{5-7}\cline{8-10}
  {  Method} &
  {  MSE} &
  {  \begin{tabular}[c]{@{} c@{}} MSE-$\subX$\end{tabular} } &
  { \begin{tabular}[c]{@{} c@{}} Area \end{tabular}}&
  {  MSE} &
  {  \begin{tabular}[c]{@{} c@{}} MSE-$\subX$\end{tabular} } &
  { \begin{tabular}[c]{@{} c@{}} Area \end{tabular}}&
  {  MSE} &
  {  \begin{tabular}[c]{@{} c@{}} MSE-$\subX$\end{tabular} } &
  { \begin{tabular}[c]{@{} c@{}} Area \end{tabular}} \\
  \hline
  LBO30     & 11   & 62.5 & 1.96 & 1.7  & 2.12 & 0.82 & 1.39 & 1.48 & 1.90 \\
  PAT15+15  & \textbf{5.66} & \textbf{27.1} & 2.31 & \textbf{0.71} & \textbf{0.5}  & \textbf{0.46} & \textbf{1.08} & \textbf{1.07} & 1.63 \\
  HAM15+15  &  6.61    & 32.87 & \textbf{1.86}  & 0.86 & 0.57 & 0.51 & 1.11 & 1.13 & \textbf{1.54} \\
  LMH15+15 &  13.64 &   61.25    & 3.54 &  1.5  & 0.98 & 0.71 & 1.78 & 1.9 & 1.93
\end{tabular}
\caption{In all settings we considered $30$ dimensional encodings; for LBO we used the first $30$ eigenvalues, for PAT, HAM, and LMH we considered the first $15$ eigenvalues from LBO and the first $15$ from the local operators.}
\label{tab:dif_op}
\end{table*}
\begin{figure}[!t]
         \footnotesize
  \begin{overpic}
         [trim=0cm 0cm 0cm 0cm,clip,width=\linewidth]{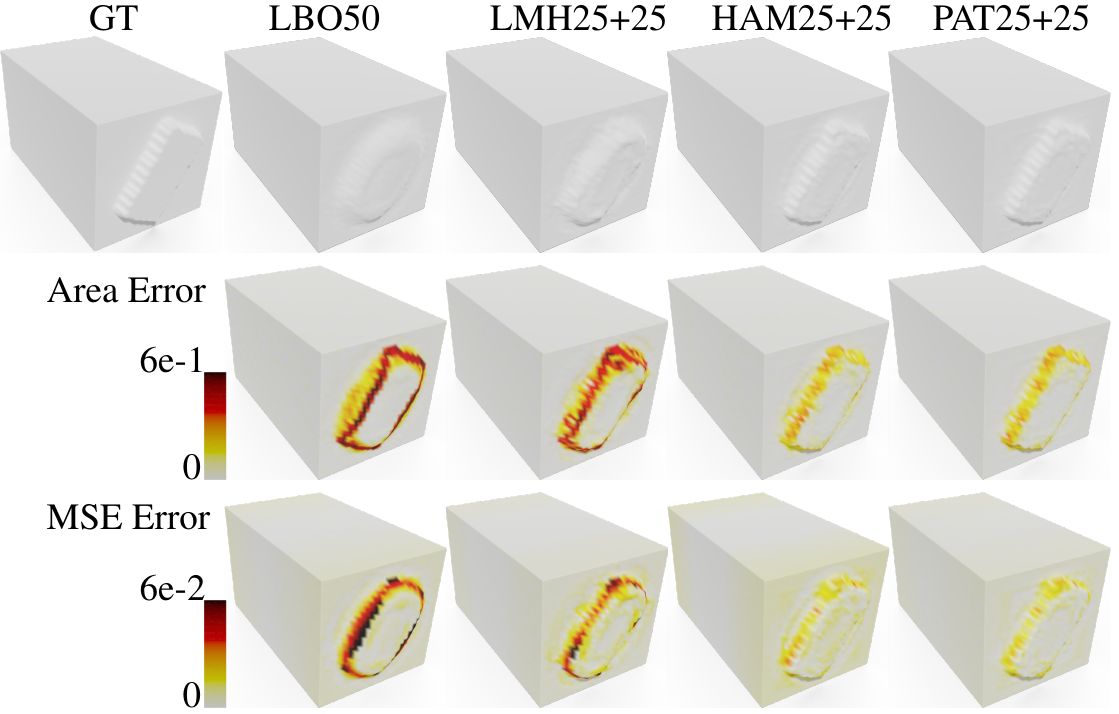}      
         \end{overpic}
     \caption{Qualitative results on the CUBE with extrinsic and intrinsic measures. With MSE, a wrong scale causes an error that accumulates on the cube extremities, while the Area error highlights the wrong pattern extrusion. Errors are color-coded, growing from white to dark red.}

 \end{figure}
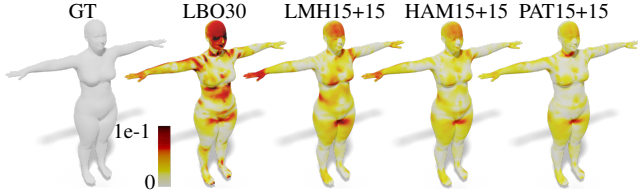
\begin{figure}[!t]
\footnotesize
        \begin{overpic}
        [trim=0cm 0cm 0cm 0cm,clip,width=1\linewidth]{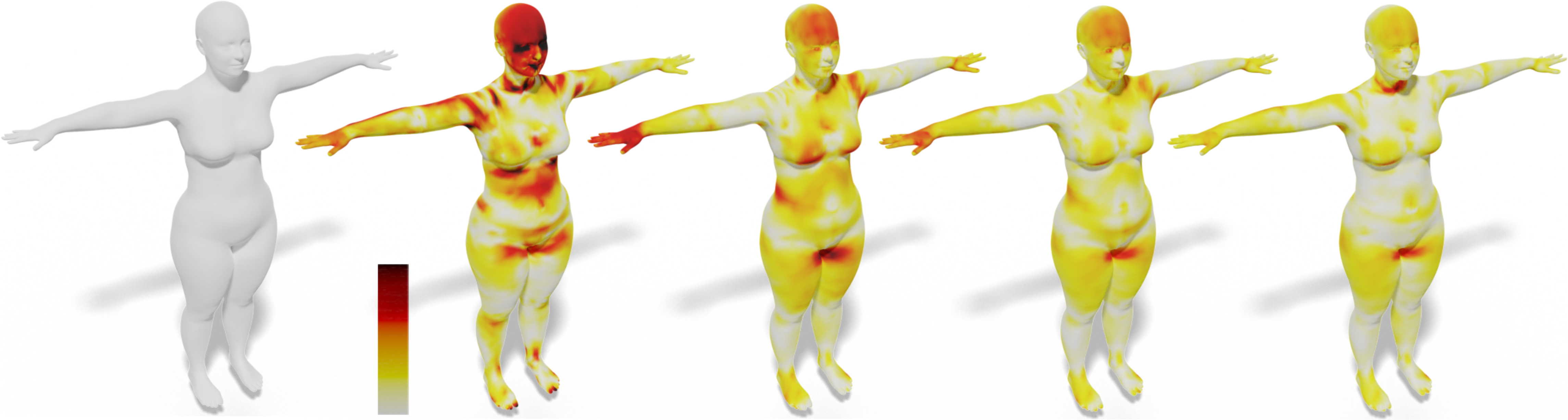}
        \put(10,27){GT}
        \put(28,27){LBO30}
        \put(44,27){LMH15+15}
        \put(63,27){HAM15+15}
        \put(81,27){PAT15+15}
        \put(17,8){1e-1}
        \put(22,0){0}
        \end{overpic}
    \caption{Qualitative result on the SURREAL. The Area error is shown on the reconstructed surfaces, encoded by color growing from white to dark red.}
    \label{fig:surrealError}
\end{figure}

\subsection{Dimension of the embedding}
\begin{table}
\centering
\begin{tabular}{@{\extracolsep{4pt}} k C C c }
\toprule
& \multicolumn{3}{c}{CUBE}\\\cline{2-4}
  {  Method} &
  {  MSE} &
  {  \begin{tabular}[c]{@{} c@{}} MSE-$\subX$\end{tabular} } &
  {Area}
  \\
  \midrule[0.09em]
  { LBO30} & { 11} &  { 62.5} &
  {1.96}  \\

    { LBO50} &
  {10.7} &  { 62} &
  {1.80}   \\

  { LBO80} & { 11.1} &  { 63.5} &
  {1.73}   \\ \midrule[0.07em] \midrule[0.07em]
  { PAT15+15} & { 5.66} & { 27.1} &
  {2.31}  \\
  { PAT25+25} &
  { \textbf{3.59} } &  \textbf{ 19.1} &  \textbf{1.33} \\
  { HAM25+25 } & {{4.07}} &{ 20.1} &
  {\textbf{1.33}}  \\ 
    { LMH25+25 } & { 14.7} & { 69.2} &
  {1.96} \\
\bottomrule
\end{tabular}
\caption{Reconstruction error on CUBE at varying size of input} 
\label{tab:dimensions}
\end{table}

Since previous works do not highlight the relation between the input dimension and the obtained reconstruction, we analyzed the method scalability for the different operators. In Table \ref{tab:dimensions} we report the results with LBO at varying dimensions. We see that the instability of higher frequencies dramatically impacts the scalability, preventing further improvement. Instead, considering local operators, both PAT and HAM show improvement when the encoding grows in size. In the supplementary, we report further results on different ratios between global and local information, showing that, in general, an even splitting between the two provides the best results.
\hspace{1.5cm}

\subsection{Number and kind of local regions}
\setlength{\columnsep}{3pt}
\setlength{\intextsep}{1pt}
\begin{table}
\begin{tabular}{@{\extracolsep{4pt}} k C C C C }
\toprule
& \multicolumn{2}{c}{SURREAL} & \multicolumn{2}{c}{SMAL}\\ \cline{2-3}\cline{4-5}
  {  Method} &
  {  \begin{tabular}[c]{c} MSE \end{tabular}} & 
  { \begin{tabular}[c]{@{} c@{}} Area\end{tabular}} &
  {  \begin{tabular}[c]{c} MSE \end{tabular}} & 
  { \begin{tabular}[c]{@{} c@{}} Area \end{tabular}} 
  \\
  \midrule[0.09em]

  {  \begin{tabular}[c]{@{} c@{}}LBO30\end{tabular}} &
  {  1.7} &
  {0.82} & {  1.39} &  {1.90}
  \\ 
  \midrule[0.07em] \midrule[0.07em] 

  {  PAT$_\mathcal{H}$15+15 } &
  \textbf{ 0.71} &
  \textbf{0.46} & {\textbf{1.08}} & {\textbf{1.63}} 
  \\ \midrule[0.07em] \midrule[0.07em]
  
  { PAT$_\mathcal{F}$15+15} &
  { 4.33 } & 
  {1.3} & & 
  \\ \hline
  
  { PAT$_\mathcal{T}$15+15} &
  { 1.71} &  
  {0.94} & {3.93} & {3.05}
  \\ 
  \midrule[0.07em] \midrule[0.07em]

  { PAT$_{\mathcal{H}+\mathcal{T}}$ 10+10+10} &
  { 0.89 } & {0.56} & {1.96}& {2.37}
  \\ 
  \hline
  
    { PAT$_{\mathcal{H}+\mathcal{T}}$ 15+15+15} &
    \textit{ 0.68 } & {0.48} & {1.82} & {2.24} 
  \\ 

\bottomrule

\end{tabular}
\caption{Reconstruction error on SURREAL and SMAL test sets considering various regions} 
\label{tab:regions}
\end{table}

\begin{wrapfigure}[4]{r}{0.25\linewidth}
\vspace{-0.3cm}
\begin{center}
 \begin{overpic}
        [trim=0cm 0cm 0cm 0cm,clip,width=0.7\linewidth]{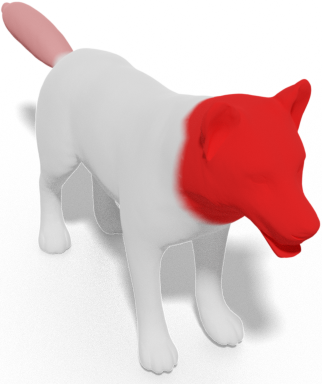}
        \put(83,55){\footnotesize $\textcolor[rgb]{1,0,0}{\mathcal{H}}$}
        \put(-11,77){\footnotesize $\textcolor[rgb]{0.94, 0.,0.5}{\mathcal{T}}$}
        
    \end{overpic}
\end{center}
\end{wrapfigure}
In this section, we investigate the importance of the choice of the selected region.
Previously, we consider a unique region $\mathcal{R}$ for each class, claiming that it characterizes the objects. In particular, we stated that this selection lets the network correlate between local geometric patterns and global features of the shape, and that this relation respects some semantics. 
\setlength{\columnsep}{7pt}
\setlength{\intextsep}{1pt}
\begin{wrapfigure}[6]{r}{0.35\linewidth}
\begin{center}
\begin{overpic}
[trim=0cm 0cm 0cm 0cm,clip,width=0.9\linewidth]{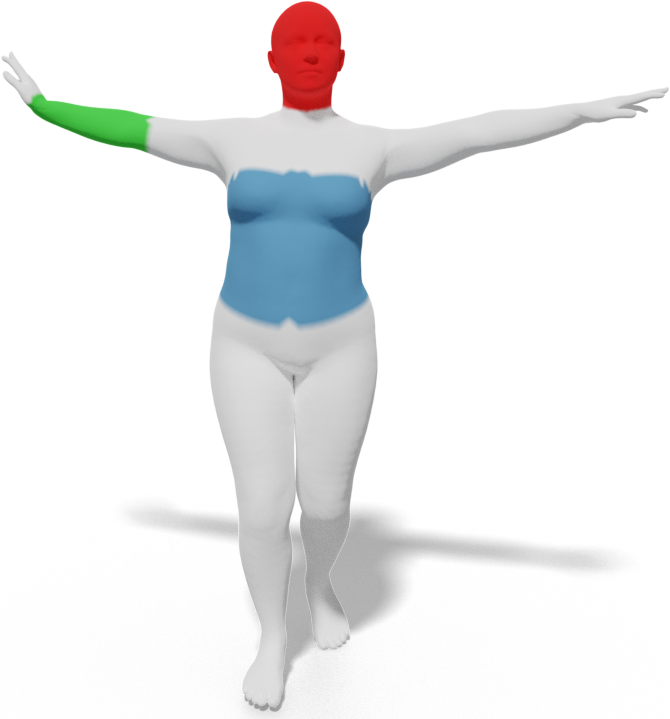}
\put(6,66){\footnotesize $\textcolor[rgb]{0.19608,0.80392,0.19608}{\mathcal{F}}$}
\put(50,91){\footnotesize $\textcolor[rgb]{1,0,0}{\mathcal{H}}$}
\put(54,61){\footnotesize $\textcolor[rgb]{0.27451,0.5098,0.70588}{\mathcal{T}}$}
\end{overpic}
\end{center}
\end{wrapfigure}
In the SMAL dataset, we tested two different regions: head $\mathcal{H}$ and tail $\mathcal{T}$, respectively depicted in red and pink in the inset figure, and as PAT$_\mathcal{H}$15+15 and  PAT$_\mathcal{T}$15+15 in Table \ref{tab:regions}. With respect to LBO30, with PAT$_\mathcal{H}$15+15 the error decreases, while PAT$_\mathcal{T}$15+15 worsens significantly. We justify this behavior with the absence of enough high-frequency details on the tail to produce an informative spectrum for the entire shape. 

In the same spirit of the previous experiment, we validate this hypothesis on the SURREAL dataset, selecting another region with few features, i.e., in which the details are less characterizing.
We choose the right forearm as local region $\mathcal{F}$ on the human body (highlighted in green in the inset figure), and we trained the model PAT$15+15$. We refer to this method as PAT$_{\mathcal{F}}15+15$, while for the method with the head as region $\mathcal{H}$ we use the notation PAT$_\mathcal{H}15+15$.
In the third row of Table \ref{tab:regions}, we report the errors for this choice.
Similar to the tail region in SMAL, the errors are higher than PAT$_\mathcal{H}15+15$ and LBO30. 
We also test PAT$15+15$ with the torso as region $\mathcal{T}$, (highlighted in blue in the inset figure) and refer to it as PAT$_{\mathcal{T}}15+15$.
The torso region comprises both the chest (that can significantly change between men and women) and the abdomen (that is larger or thinner in different subjects). $\mathcal{T}$ can be comparable to the tail region but with more continuous and significant variations.
In all cases, on SURREAL, $PAT15+15$ performs significantly better. 
These experiments emphasize the importance of the input representation and that some local regions contain more information on the whole shape than others.
In particular the torso region, besides having important features, has a central position in the shape. Thus its improvement affects the MSE of all the other body parts.
On the contrary, the tail region on the SMAL dataset has indeed similar important features but it is a peripheral region and its improvement does not flow on other regions.
The elbow instead is a region with poor features that do not add enough information to the global spectrum. 

\noindent\textbf{Multi-region.} Until now, we experimented only using a single local spectrum as input. To underline how our method can be used with multiple regions, we perform additional experiments on SURREAL and SMAL without modifying the parameters of the decoder.
In SMAL, we have already tested the single efficacy of the head and tail. In the next experiment, we test their efficacy when combined together.
We train the PAT version using as input the local spectra computed both on the head and tail surface. We indicate the results with {PAT$_{\mathcal{H}+\mathcal{T}}k+h'+h''$} where $k$ is the number of LBO eigenvalues, $h'$ the number of local eigenvalues for the first region (head) and $h''$ the number of local eigenvalues for the second region (tail for SMAL and torso for SURREAL). In particular, we consider the cases $10+10+10$ and $15+15+15$. Notice that in this second case the input encoding is larger then other methods.
In the last two rows of Table \ref{tab:regions} we show the instrinsic and extrinsic errors.
The addition of multiple regions does not improve the performance of PAT$_\mathcal{H}$15+15, but neither worsens them as in PAT$_\mathcal{T}$15+15. 

Our conclusions are that including other regions can be done, giving more freedom during the generation. At the same time, there is a trade-off between the control and the reconstruction quality, since increasing the number of regions limits the amount of encoding assigned to each part \revision{(PAT$_\mathcal{T}$10+10+10)}. Discovering the patterns across multiple regions is more challenging; it requires a design in the encoding partition (i.e., assigning to each region eigenvalues proportional to its semantic significance, as we report in supplementary materials) or a deeper network to properly exploit the encoding information.
\revision{
Even when we increased the number of total eigenvalues in input (PAT$_{\mathcal{H}+\mathcal{T}}$ 15+15+15) the performance did not always improve. 
We attribute this slight drop to the fact that, for a fair comparison, we trained all the models with the same number of epochs, while PAT$_{\mathcal{H}+\mathcal{T}}$ 15+15+15 may have needed more time to optimize the larger input information effectively.
}

\subsection{Decoder-only vs Autoencoder}
In Table \ref{tab:deconly}, we compare our architecture (32M parameters) against the one proposed in ~\cite{instant2020} (9M parameters) also by empowering its decoder and, coherently, its encoder (namely ~\cite{instant2020}$_{big}$ in the table, 90M parameters). We trained~\cite{instant2020} using the absolute value of eigenvalues, as proposed in the original paper. Results show not only that our decoder approach is better than the full architecture even in the LBO setting, but that~\cite{instant2020} is not equally capable of combining local information with its latent space.
\begin{table}[t!]
\footnotesize
\centering

\begin{tabular}{@{\extracolsep{4pt}} k C C C}
\hline
    & \multicolumn{3}{c}{SURREAL} \\\cline{2-4}
  {  Method} &
  {  MSE} &
  {  \begin{tabular}[c]{@{} c@{}} MSE-$\subX$\end{tabular} } &
  { \begin{tabular}[c]{@{} c@{}} Area \end{tabular}} \\
  \hline
  LBO30      & 1.7  & 2.12 & 8.19   \\
  PAT15+15   & \textbf{0.71} & \textbf{0.5}  & \textbf{4.58}  \\
  \hline
  \cite{instant2020} PAT15+15 & 3.1 & 3.89 & 19.81\\
  \cite{instant2020}$_{big}$ PAT15+15 & 2.51 & 1.8 & 16.9 \\
  \cite{instant2020}$_{big}$ LBO30 & 2.51 & 2.26 & 15.74 
\end{tabular}
\caption{\label{tab:deconly}In all settings we considered $30$ dimensional encodings; for LBO we used the first $30$ eigenvalues, for PAT, HAM, and LMH we considered the first $15$ eigenvalues from LBO and the first $15$ from the local operators computed on the head region.}
\end{table}

\section{Conclusions}
\label{sec:conclusions}
This paper presented a novel approach for generating and modeling 3D shapes from a canonical and ubiquitous spectral representation. We consider this theoretically exciting task helpful for shape manipulation, especially in combining semantic characteristics of local and global parts. We highlighted several properties of local spectral operators and their relation with the standard Laplacian in this encoding. For the first time, we performed a shape from spectrum pipeline from a mix of spectra of different operators. Furthermore, we have also shown the close relationship between spectra and the geometry from which they come, even in a noisy and unsupervised scenario.  
Dealing with multi-region localization would be a stimulating future direction, considering parts with different semantics and proportions. While a complete analysis is beyond this paper's scope, our preliminary evidence suggests this is a promising field for further exploration.

The main theoretical limitation of our study is not considering the spectra of extrinsic operators, like the Dirac operator~\cite{Dirac}. Extrinsic operators could be successfully injected into our pipeline, providing a mixture of intrinsic and extrinsic information. Moreover, we do not consider inter-class experiments since the spectrum may be ambiguous among different classes. We believe that our work may elicit discussion in the community on these topics.
The main applicative limitations arise from the limitation of the intrinsic spectral representations in the presence of shapes with different topology, significant noise, or outliers. The research on these aspects is quite active in the community, and our method lends itself well to methodological progress.   

\paragraph*{Acknowledgements}
Parts of this work were supported by the ERC Starting Grant No. 802554 (SPECGEO), the SAPIENZA BE-FOR-ERC 2020 Grant (NONLINFMAPS) and the MIUR under grant “Dipartimenti di eccellenza
2018-2022” of the Department of Computer Science of the Sapienza
University of Rome and the University of Verona.

\bibliographystyle{eg-alpha-doi}  
\bibliography{egbibsample}  

\newcommand{\etalchar}[1]{$^{#1}$}
\begin{thebibliography}{\uppercase{PMRMB15}}

\bibitem[AATJD19]{geodisent}
\textsc{Aumentado-Armstrong T., Tsogkas S., Jepson A., Dickinson S.}:
\newblock Geometric disentanglement for generative latent shape models.
\newblock In \emph{International Conference on Computer Vision (ICCV)} (2019).

\bibitem[ADMG18]{achlioptas2018learning}
\textsc{Achlioptas P., Diamanti O., Mitliagkas I., Guibas L.}:
\newblock Learning representations and generative models for 3d point clouds.
\newblock In \emph{International Conference on Machine Learning} (2018),
  pp.~40--49.

\bibitem[BSTPM20]{bhatnagar2020ipnet}
\textsc{Bhatnagar B.~L., Sminchisescu C., Theobalt C., Pons-Moll G.}:
\newblock Combining implicit function learning and parametric models for 3d
  human reconstruction.
\newblock In \emph{European Conference on Computer Vision ({ECCV})} (2020).

\bibitem[CKF{\etalchar{*}}21]{Chen_2021_CVPR}
\textsc{Chen Z., Kim V.~G., Fisher M., Aigerman N., Zhang H., Chaudhuri S.}:
\newblock Decor-gan: 3d shape detailization by conditional refinement.
\newblock In \emph{Proceedings of the IEEE/CVF Conference on Computer Vision
  and Pattern Recognition (CVPR)} (June 2021), pp.~15740--15749.

\bibitem[CNH{\etalchar{*}}20]{cosmo2020limp}
\textsc{Cosmo L., Norelli A., Halimi O., Kimmel R., Rodol{\`a} E.}:
\newblock {LIMP: Learning Latent Shape Representations with Metric Preservation
  Priors}.
\newblock In \emph{European Conference on Computer Vision ({ECCV})} (2020).

\bibitem[CPR{\etalchar{*}}19]{isosp}
\textsc{Cosmo L., Panine M., Rampini A., Ovsjanikov M., Bronstein M.~M.,
  Rodol{\`a} E.}:
\newblock Isospectralization, or how to hear shape, style, and correspondence.
\newblock In \emph{Proceedings of the IEEE Conference on Computer Vision and
  Pattern Recognition (CVPR)} (2019), pp.~7529--7538.

\bibitem[CRXZ20]{EG_star_generative}
\textsc{Chaudhuri S., Ritchie D., Xu K., Zhang H.}:
\newblock Learning generative models of 3d structures.
\newblock \emph{Computer Graphics Forum {(CGF)} 39}, 2 (2020), 643--666.

\bibitem[CSBK20]{Choukroun}
\textsc{{Choukroun} Y., {Shtern} A., {Bronstein} A., {Kimmel} R.}:
\newblock Hamiltonian operator for spectral shape analysis.
\newblock \emph{IEEE Transactions on Visualization and Computer Graphics 26}, 2
  (2020), 1320--1331.

\bibitem[CYAE{\etalchar{*}}20]{ShapeGF}
\textsc{Cai R., Yang G., Averbuch-Elor H., Hao Z., Belongie S., Snavely N.,
  Hariharan B.}:
\newblock Learning gradient fields for shape generation.
\newblock In \emph{European Conference on Computer Vision({ECCV})} (2020).

\bibitem[DMB{\etalchar{*}}17]{bic17}
\textsc{{Denitto} M., {Melzi} S., {Bicego} M., {Castellani} U., {Farinelli} A.,
  {Figueiredo} M. A.~T., {Kleiman} Y., {Ovsjanikov} M.}:
\newblock Region-based correspondence between 3d shapes via spatially smooth
  biclustering.
\newblock In \emph{2017 IEEE International Conference on Computer Vision
  (ICCV)} (2017), pp.~4270--4279.

\bibitem[EST{\etalchar{*}}20]{egger20203d}
\textsc{Egger B., Smith W.~A., Tewari A., Wuhrer S., Zollhoefer M., Beeler T.,
  Bernard F., Bolkart T., Kortylewski A., Romdhani S., et~al.}:
\newblock 3d morphable face models—past, present, and future.
\newblock \emph{ACM Transactions on Graphics (TOG) 39}, 5 (2020), 1--38.

\bibitem[GFK{\etalchar{*}}18]{ATLASNET}
\textsc{Groueix T., Fisher M., Kim V.~G., Russell B., Aubry M.}:
\newblock {AtlasNet: A Papier-M\^ach\'e Approach to Learning 3D Surface
  Generation}.
\newblock In \emph{Proceedings IEEE Conf. on Computer Vision and Pattern
  Recognition (CVPR)} (2018).

\bibitem[GGC{\etalchar{*}}20]{shapehandles}
\textsc{Gadelha M., Gori G., Ceylan D., Mech R., Carr N., Boubekeur T., Wang
  R., Maji S.}:
\newblock Learning generative models of shape handles.
\newblock In \emph{IEEE Conference on Computer Vision and Pattern Recognition
  (CVPR)} (2020).

\bibitem[GWW92]{gordon1992one}
\textsc{Gordon C., Webb D.~L., Wolpert S.}:
\newblock One cannot hear the shape of a drum.
\newblock \emph{Bulletin of the American Mathematical Society 27}, 1 (1992),
  134--138.

\bibitem[HHGCO20]{hertz2020pointgmm}
\textsc{Hertz A., Hanocka R., Giryes R., Cohen-Or D.}:
\newblock Pointgmm: A neural gmm network for point clouds.
\newblock In \emph{Proceedings of the IEEE/CVF Conference on Computer Vision
  and Pattern Recognition} (2020), pp.~12054--12063.

\bibitem[JSS18]{joo2018total}
\textsc{Joo H., Simon T., Sheikh Y.}:
\newblock Total capture: A {3D} deformation model for tracking faces, hands,
  and bodies.
\newblock In \emph{Proceedings of the IEEE conference on computer vision and
  pattern recognition} (2018), pp.~8320--8329.

\bibitem[JZCZ20]{Jiang2020HumanBody}
\textsc{Jiang B., Zhang J., Cai J., Zheng J.}:
\newblock Disentangled human body embedding based on deep hierarchical neural
  network.

\bibitem[Kac66]{kac1966can}
\textsc{Kac M.}:
\newblock Can one hear the shape of a drum?
\newblock \emph{The american mathematical monthly 73}, 4P2 (1966), 1--23.

\bibitem[KAMC]{Kalogerakis}
\textsc{{Kalogerakis} E., {Averkiou} M., {Maji} S., {Chaudhuri} S.}:
\newblock 3d shape segmentation with projective convolutional networks.
\newblock In \emph{2017 IEEE Conference on Computer Vision and Pattern
  Recognition (CVPR)}, pp.~6630--6639.

\bibitem[KO19]{Kleiman19}
\textsc{Kleiman Y., Ovsjanikov M.}:
\newblock Robust structure-based shape correspondence.
\newblock \emph{Comput. Graph. Forum 38}, 1 (2019), 7--20.

\bibitem[LHW{\etalchar{*}}19]{L2G2019}
\textsc{Liu X., Han Z., Wen X., Liu Y.-S., Zwicker M.}:
\newblock L2g auto-encoder: Understanding point clouds by local-to-global
  reconstruction with hierarchical self-attention.
\newblock In \emph{Proceedings of the 27th ACM International Conference on
  Multimedia} (2019).

\bibitem[LJC17]{Dirac}
\textsc{Liu H.-T.~D., Jacobson A., Crane K.}:
\newblock A dirac operator for extrinsic shape analysis.
\newblock \emph{Computer Graphics Forum 36}, 5 (2017), 139--149.

\bibitem[LMR{\etalchar{*}}15]{loper15}
\textsc{Loper M., Mahmood N., Romero J., Pons-Moll G., Black M.~J.}:
\newblock {SMPL}: A skinned multi-person linear model.
\newblock \emph{ACM Trans. Graph. 34}, 6 (2015), 248:1--248:16.

\bibitem[LZZ{\etalchar{*}}21]{luo2021simpmodeling}
\textsc{Luo Z., Zhou J., Zhu H., Du D., Han X., Fu H.}:
\newblock Simpmodeling: Sketching implicit field to guide mesh modeling for 3d
  animalmorphic head design.
\newblock In \emph{The 34th Annual ACM Symposium on User Interface Software and
  Technology} (2021), pp.~854--863.

\bibitem[MGY{\etalchar{*}}19]{mo2019structurenet}
\textsc{Mo K., Guerrero P., Yi L., Su H., Wonka P., Mitra N., Guibas L.}:
\newblock Structurenet: Hierarchical graph networks for 3d shape generation.
\newblock \emph{ACM Transactions on Graphics (TOG), Siggraph Asia 2019 38}, 6
  (2019), Article 242.

\bibitem[MMRC20]{FARM}
\textsc{Marin R., Melzi S., Rodol\`a E., Castellani U.}:
\newblock Farm: Functional automatic registration method for 3d human bodies.
\newblock \emph{Computer Graphics Forum 39}, 1 (2020), 160--173.

\bibitem[MRC{\etalchar{*}}20]{instant2020}
\textsc{Marin R., Rampini A., Castellani U., Rodol\`a E., Ovsjanikov M., Melzi
  S.}:
\newblock Instant recovery of shape from spectrum via latent space connections.
\newblock In \emph{International Conference on 3D Vision (3DV)} (2020).

\bibitem[MRC{\etalchar{*}}21]{Instant2021}
\textsc{Marin R., Rampini A., Castellani U., Rodol{\`a} E., Ovsjanikov M.,
  Melzi S.}:
\newblock Spectral shape recovery and analysis via data-driven connections.
\newblock \emph{International Journal of Computer Vision} (2021), 1573--1405.

\bibitem[MRCB18]{LMH}
\textsc{Melzi S., Rodol\`{a} E., Castellani U., Bronstein M.~M.}:
\newblock Localized manifold harmonics for spectral shape analysis.
\newblock \emph{Computer Graphics Forum 37}, 6 (2018), 20--34.

\bibitem[PMRMB15]{Dyna2015}
\textsc{Pons-Moll G., Romero J., Mahmood N., Black M.~J.}:
\newblock Dyna: A model of dynamic human shape in motion.
\newblock \emph{ACM Transactions on Graphics, (Proc. SIGGRAPH) 34}, 4 (Aug.
  2015), 120:1--120:14.

\bibitem[PP93]{pinkall1993computing}
\textsc{Pinkall U., Polthier K.}:
\newblock Computing discrete minimal surfaces and their conjugates.
\newblock \emph{Experimental mathematics 2}, 1 (1993), 15--36.

\bibitem[QSMG17]{qi2017pointnet}
\textsc{Qi C.~R., Su H., Mo K., Guibas L.~J.}:
\newblock Pointnet: Deep learning on point sets for 3d classification and
  segmentation.
\newblock In \emph{Proceedings of the IEEE Conference on Computer Vision and
  Pattern Recognition} (2017), pp.~652--660.

\bibitem[RBSB18]{COMA}
\textsc{Ranjan A., Bolkart T., Sanyal S., Black M.~J.}:
\newblock Generating {3D} faces using convolutional mesh autoencoders.
\newblock In \emph{European Conference on Computer Vision (ECCV)} (2018).

\bibitem[RDP99]{robinette1999caesar}
\textsc{Robinette K.~M., Daanen H., Paquet E.}:
\newblock The caesar project: a 3-d surface anthropometry survey.
\newblock In \emph{Proc. Second International Conference on 3-D Digital Imaging
  and Modeling} (Washington, DC, USA, oct 1999), IEEE, pp.~380--386.

\bibitem[RPC{\etalchar{*}}21]{advatt}
\textsc{Rampini A., Pestarini F., Cosmo L., Melzi S., Rodol{\`{a}} E.}:
\newblock Universal spectral adversarial attacks for deformable shapes.
\newblock In \emph{{IEEE} Conference on Computer Vision and Pattern
  Recognition, {CVPR}} (2021), pp.~3216--3226.

\bibitem[RTB17]{MANO2017}
\textsc{Romero J., Tzionas D., Black M.~J.}:
\newblock Embodied hands: Modeling and capturing hands and bodies together.
\newblock \emph{ACM Transactions on Graphics, (Proc. SIGGRAPH Asia) 36}, 6
  (Nov. 2017).

\bibitem[RTO{\etalchar{*}}19]{hamiltonian}
\textsc{Rampini A., Tallini I., Ovsjanikov M., Bronstein A.~M., Rodol{\`a} E.}:
\newblock Correspondence-free region localization for partial shape similarity
  via hamiltonian spectrum alignment.
\newblock In \emph{International Conference on 3D Vision (3DV)} (2019).

\bibitem[SACO20]{sharp2020diffusion}
\textsc{Sharp N., Attaiki S., Crane K., Ovsjanikov M.}:
\newblock Diffusion is all you need for learning on surfaces, 2020.
\newblock \href {http://arxiv.org/abs/2012.00888} {\path{arXiv:2012.00888}}.

\bibitem[SC20]{Sharp2020}
\textsc{Sharp N., Crane K.}:
\newblock {A Laplacian for Nonmanifold Triangle Meshes}.
\newblock \emph{Computer Graphics Forum (SGP) 39}, 5 (2020).

\bibitem[TTZ{\etalchar{*}}20]{Tretschk2020DEMEA}
\textsc{Tretschk E., Tewari A., Zollh\"{o}fer M., Golyanik V., Theobalt C.}:
\newblock {{DEMEA}: Deep Mesh Autoencoders for Non-Rigidly Deforming Objects}.
\newblock \emph{European Conference on Computer Vision ({ECCV})} (2020).

\bibitem[VRM{\etalchar{*}}17]{varol17_surreal}
\textsc{Varol G., Romero J., Martin X., Mahmood N., Black M.~J., Laptev I.,
  Schmid C.}:
\newblock Learning from synthetic humans.
\newblock In \emph{CVPR} (2017).

\bibitem[WZX{\etalchar{*}}16]{3dgan}
\textsc{Wu J., Zhang C., Xue T., Freeman W.~T., Tenenbaum J.~B.}:
\newblock Learning a probabilistic latent space of object shapes via 3d
  generative-adversarial modeling.
\newblock In \emph{Advances in Neural Information Processing Systems} (2016),
  pp.~82--90.

\bibitem[WZX{\etalchar{*}}20]{Wu_2020_CVPR}
\textsc{Wu R., Zhuang Y., Xu K., Zhang H., Chen B.}:
\newblock Pq-net: A generative part seq2seq network for 3d shapes.
\newblock In \emph{IEEE/CVF Conference on Computer Vision and Pattern
  Recognition (CVPR)} (June 2020).

\bibitem[XBZ{\etalchar{*}}20]{xu2020ghum}
\textsc{Xu H., Bazavan E.~G., Zanfir A., Freeman W.~T., Sukthankar R.,
  Sminchisescu C.}:
\newblock {GHUM} \& {GHUML}: Generative {3D} human shape and articulated pose
  models.
\newblock In \emph{Proceedings of the IEEE/CVF Conference on Computer Vision
  and Pattern Recognition} (2020), pp.~6184--6193.

\bibitem[YCC{\etalchar{*}}20]{yin2020coalesce}
\textsc{Yin K., Chen Z., Chaudhuri S., Fisher M., Kim V.~G., Zhang H.}:
\newblock Coalesce: Component assembly by learning to synthesize connections.
\newblock In \emph{2020 International Conference on 3D Vision (3DV)} (2020),
  IEEE, pp.~61--70.

\bibitem[YSS{\etalchar{*}}17]{yi2017large}
\textsc{Yi L., Shao L., Savva M., Huang H., Zhou Y., Wang Q., Graham B.,
  Engelcke M., Klokov R., Lempitsky V., et~al.}:
\newblock Large-scale 3d shape reconstruction and segmentation from shapenet
  core55.
\newblock \emph{arXiv preprint arXiv:1710.06104} (2017).

\bibitem[ZBPM20]{zhou20unsupervised}
\textsc{Zhou K., Bhatnagar B.~L., Pons-Moll G.}:
\newblock Unsupervised shape and pose disentanglement for 3d meshes.
\newblock In \emph{European Conference on Computer Vision (ECCV)} (August
  2020).

\bibitem[ZKJB17]{SMAL}
\textsc{Zuffi S., Kanazawa A., Jacobs D., Black M.~J.}:
\newblock {3D} menagerie: Modeling the {3D} shape and pose of animals.
\newblock In \emph{IEEE Conf. on Computer Vision and Pattern Recognition
  (CVPR)} (July 2017).

\end{thebibliography}

\clearpage
\appendix
\renewcommand*{\thesection}{\arabic{section}}

{%
   \centerline{\LARGE \textbf{Supplementary Material}}%
   \vspace*{12pt}%
}
\section{Experimental Setup}
\label{sec:setups}
\subsection{Dataset}

Here we report additional details about the datasets involved in our experiments.

\vspace{1ex}\noindent\textbf{\em CUBE.} 
In the CUBE dataset, the local variations are extrusions of simple geometric patterns (circle, ellipsis, square, and rectangle) applied on a selected face same for all cuboids. We vary dimensions and rotations of these patterns avoiding isometric shapes (that are indistinguishable by their eigenvalues).
For the global variations, we scaled the cube along the dimension orthogonal to the face with local variation by a factor in the range $[0.6, 2]$, obtaining cuboids with different depths. 

\vspace{1ex}\noindent\textbf{\em SURREAL.} 
The shapes in SURREAL are generated by SMPL~\cite{loper15}, a standard generative template with $6890$ vertices and two sets of parameters: one for the subject identity and one for its pose. Since pose changes generate near-isometric shapes, we set all the individuals in the same T-pose. The shape parameters are sampled from the ones available from SURREAL dataset~\cite{varol17_surreal}.

\vspace{1ex}\noindent\textbf{\em AIRPLANES.}
In the AIRPLANES dataset we chose the segment of the tail as the local region because we think that the tail is a semantically significant region of the airplane: it is related to the airplane type (e.g., Boeing, Jet, Fighter) and its size.

\subsection{Architecture and training details}
\label{sec:arc}
Our architecture is a simple decoder composed of 4 fully connected layers. All the hidden layers use batch normalization followed by a selu activation, while the last layer has a linear activation. We report the number of nodes for each layer in Tab.~\ref{tab:params}. For the SURREAL dataset we add a dropout layer with a $0.1$ drop rate to all hidden layers.
We trained our network on $90\%$ of the dataset and used the remaining $10\%$ for testing.
During training we used Adam optimizer with a learning rate of $2*10^{-3}$ for the first 1000 epochs and then we reduce it to $1.8*10^{-3}$ for the rest of the training. We fixed the maximum number of epoch in each dataset making sure each method reached convergence.
\revision{The output of $\Pi$ is a matrix $X\in \mathbb{R}^{n \times 3}$ encoding the vertex coordinates.}
In the second part of Table \ref{tab:params} we show the training parameters.

\begin{table}[]
\centering
\tiny
\begin{tabular}{c|c|c|c|c}
{} & {CUBE} & {SURREAL} & {SMAL} & {AIRPLANES}\\ \midrule[0.09em]
\multicolumn{5}{c}{Number of Nodes} \\ \hline
{Layer 1} & {258} & {258} & {258} & {258} \\\hline 
{Layer 2} & {1024} & {512} & {512} & {1024} \\ \hline 
{Layer 3} & {2048} & {1536} & {1536} & {2048}\\ \hline
{Layer 4} & {22050} & {20670} & {20670} & {1500} \\ \hline \hline
{Output size} & {7350 x 3} & {6890 x 3} & {3889 x 3} & {500 x 3} \\ \hline
{Number of epochs} & {2000} & {1000} & {1000} &  {4500} \\ \hline
{Batch size} & {64} & {32} & {32} & {8}\\ 
\bottomrule
\end{tabular}
\caption{Networks parameters for the different datasets involved in our experiments.}
\label{tab:params}
\end{table}

\revision{
\subsection{Computation time}
We trained all the models on a NVIDIA GeForce GTX 1050 Ti. On the CUBE, SURREAL and SMAL dataset, the training time is about 2.2 hours; while on the AIRPLANES dataset is about 12 hours.
}

\subsection{Comparison to the autoencoder of \cite{instant2020}}
\begin{figure}[t]
\centering
\begin{overpic}[trim=0.1cm 0cm 0cm 0cm,clip,width=0.99\linewidth, grid=false]{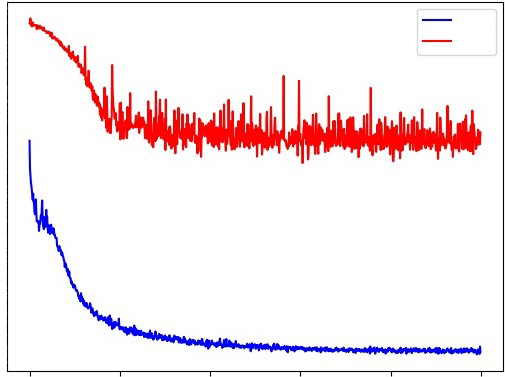}
        \put(90,70.5){\scriptsize Our}
        \put(89.5,66.5){\fontsize{4}{6}\selectfont \cite{instant2020}}
       \end{overpic}
\caption{Training loss comparison: our method in blue; in red the method proposed by \cite{instant2020}.}
\label{fig:train_losses}
\end{figure}

{ 
One of the main advantage of our method is the simplicity of the model: a single decoder composed of fully connected layers. 
This allows us to perform a more direct analysis of the linkage between spectral geometry processing and semantic modeling.
On the contrary the model proposed by \cite{instant2020} is composed of an autoencoder enhanced with an invertible module blurring a similar analysis.
In fact, the correspondence between the spectrum and the object geometry established by ~\cite{instant2020} passes through a latent space impacted by other components.
Moreover, our specialized architecture performs better than the architecture proposed in~\cite{instant2020} in synthesizing shapes from the spectrum.
An other advantage of our model choice is the training. 
Fig.~\ref{fig:train_losses} shows the training loss curves of our model (in blue) and of ~\cite{instant2020} (in red). It emerges that our model not only reaches lower errors but has also a more stable training. 
}

\clearpage 
\newpage
\section{Shape from spectrum}
\label{sec:results}
This section provides further results on our analysis of the reconstruction of a 3D shape from its spectrum.
\revision{If not differently stated, the shapes we adopt in all our experiments and figures have never been seen during the training and belong to the test set or a completely different dataset.}

\subsection{Different evaluation metrics and Nearest-Neighbor comparison}
\noindent\textbf{Evaluation metrics.} In the main manuscript, we considered two extrinsic measures (global and localized MSE) and an intrinsic measure (Area error). Here we report a more complete analysis, including additional metrics.

As extrinsic measures, we report the same error optimized by the loss, referred to as \textbf{MSE}. 
With \textbf{MSE-$\subX$} and \textbf{MSE-$\subX^{C}$} we denote the same measure computed inside or outside $\subX$.
As intrinsic measures, we consider:

\textbf{Area}: as in the main manuscript, it is the average difference of the area elements of each vertex, which relates to surface stretch.

\textbf{Metric}: the vertex-wise metric distortion, computed as the difference in the geodesic distances from a fixed set of 100 uniformly sampled points to all the points in the mesh.

\textbf{Align}: the MSE reconstruction error of the local region after the best rigid alignment, obtained by solving the Procrustes problem between the local patches.

The \textbf{Area-$\subX$} and \textbf{Metric-$\subX$} are the same as above, computed for the local region.

\noindent\textbf{Nearest-Neighbor comparison.} We directly compare each method against the nearest-neighbor baseline; as done in~\cite{instant2020}, given the spectrum of a new test shape, we look in the training set for the spectrum which is the nearest in the L$2$ sense. Then, we consider as baseline output the training shape associated with this spectrum. To compare with this baseline we include \textbf{ENN} which measures the MSE of the baseline, and \textbf{EM $<$ ENN}, which indicates the percentage in which the method outperforms the baseline. Every method uses a different dataset, so each one has its baseline; this is why we considered it a column rather than multiple rows.

\subsection{Further Results}
All the results are summarized in Tables~\ref{tab:errorsTOY},~\ref{tab:errorsSURREAL} and ~\ref{tab:errorsSMAL}. The columns represent the evaluation metrics presented in the previous Section, while the rows are the different combinations of local and global spectrum. 

\begin{table*}[t!]
\footnotesize
\centering
\resizebox{\textwidth}{!}{
\begin{tabular}{ k C C C C C C C C C C C }
\toprule
  {  Method} &
  {  \begin{tabular}[c]{c} MSE \\ ($*10^{-6}$)\end{tabular}} &
  {  \begin{tabular}[c]{@{} c@{}} MSE-$\subX$ \\ ($*10^{-6}$)\end{tabular} } &
  {  \begin{tabular}[c]{@{} c@{}} MSE-$\subX^C$ \\ ($*10^{-6}$)\end{tabular}} &
  {  \begin{tabular}[c]{@{} c@{}} ENN  \\ ($*10^{-6}$)\end{tabular}} &
  {  \begin{tabular}[c]{c} \hspace{-0.4cm} EM \textless ENN \end{tabular}} &
  { \begin{tabular}[c]{@{} c@{}} Area  \\($*10^{-2}$)\end{tabular}} &
  { \begin{tabular}[c]{@{} c@{}} Area-$\subX$ \\($*10^{-2}$)\end{tabular}} &
  { \begin{tabular}[c]{@{} c@{}} Metric \\($*10^{-3}$)\end{tabular}} &
  { \begin{tabular}[c]{@{} c@{}} Metric-$\subX$ \\($*10^{-3}$)\end{tabular}} \\ 
  \midrule[0.09em]
  
  { LBO 30} & { 11} &  { 62.5} & { 0.65} &
  {15} & { 60\%} &
  {1.96} & {6.80} &
  {6.63} & {33.8} \\ \hline

    { LBO 50} &
  {10.7} &  { 62} & \second{ 0.45}  &
  {15} & { 65\%}  &
  {1.80} &  {6.76} &
  {6.41} & {3.33}  \\ \hline

  { LBO 80} & { 11.1} &  { 63.5} & { 0.63} &
  { 14}  & { 61\%}  &
  {1.73} & {6.59} &
  {6.54} & {33.7}  \\ \midrule[0.07em] \midrule[0.07em]

  { PAT 20+10} & { 6.65} & { 37.8} & { \first{0.42}} &  
  {309} & { 88\%} &
  {1.78} & {6.25} &
  {4.81} & {24.2}  \\ \hline 
  
  { PAT 15+15} & { 5.66} & { 27.1} & { 1.36}  &
  {1090} & { 95\%} &
  {2.31} & {6.01} &
  {3.96} &  {17.5}  \\ \hline

  { PAT 10+20} & { 4.91} & { 26.3} & { 0.63} &
  {1720}  & {100\%} &
  \third{1.43} & {5.37} & 
  {3.59} & {17.9}  \\ \midrule[0.13em]
  
  { PAT 40+10} & { 6.76} & { 38.5} & \first{ 0.42}  &
  {79} & { 86\%} &
  {1.59} & {6.13} &
  {4.89} &  {24.9}  \\ \hline

  { PAT 25+25} &
  { \first{3.59} } &  \second{ 19.1} &  \third{ 0.5}  &
  {2060} &  {100\%} & 
  {\first{1.33}} & \second{4.52} &
  {\first{2.76}} &  \third{13.9}  \\ \hline 

  { PAT 10+40} &
  { 6.70} & { 23.4} & { 3.36} & 
  {1860} & {  100\%} &
  {1.51} & {4.62} &
  {3.63} &  \second{13.2}  \\ \midrule[0.13em]
    
  { PAT 40+40} &
  { \second{3.77}} &  { \first{18.3}} & { 0.85} & 
  {2310} & { 100\%} &
  \second{1.35} & {\first{4.44}} & 
  \second{2.84} & {\first{12.8}} \\

  \midrule[0.07em] \midrule[0.07em]
    
  { HAM 25+25 } & { \third{4.07}} & \third{ 20.1} & { 0.86}  &
  {2050} & { 99\%} &
  {\first{1.33}} & \third{4.56} &
  \third{3.05} &  {14.1}  \\ \midrule[0.07em] \midrule[0.07em]

  { LMH 25+25 } & { 14.7} & { 69.2} & { 3.81} &
  {137}  & { 65\%} &
  {1.96} & {6.85} &
  {6.08} & {27.7} \\

\bottomrule
\end{tabular} }
\caption{Reconstruction error on the CUBE test set. For each column, we highlighted the top three results in red with decreasing intensity.}
\label{tab:errorsTOY}
\end{table*}
\def\lenC{0.18\textwidth}
\def\lenColorBar{0.02\textwidth}
\def\lenGTspace{0.03\textwidth}
\def\lenVertical{0.5cm}

\begin{figure}[t]
\begin{overpic}
    [trim=0cm 0cm 0cm 0cm,clip,width=1\linewidth]{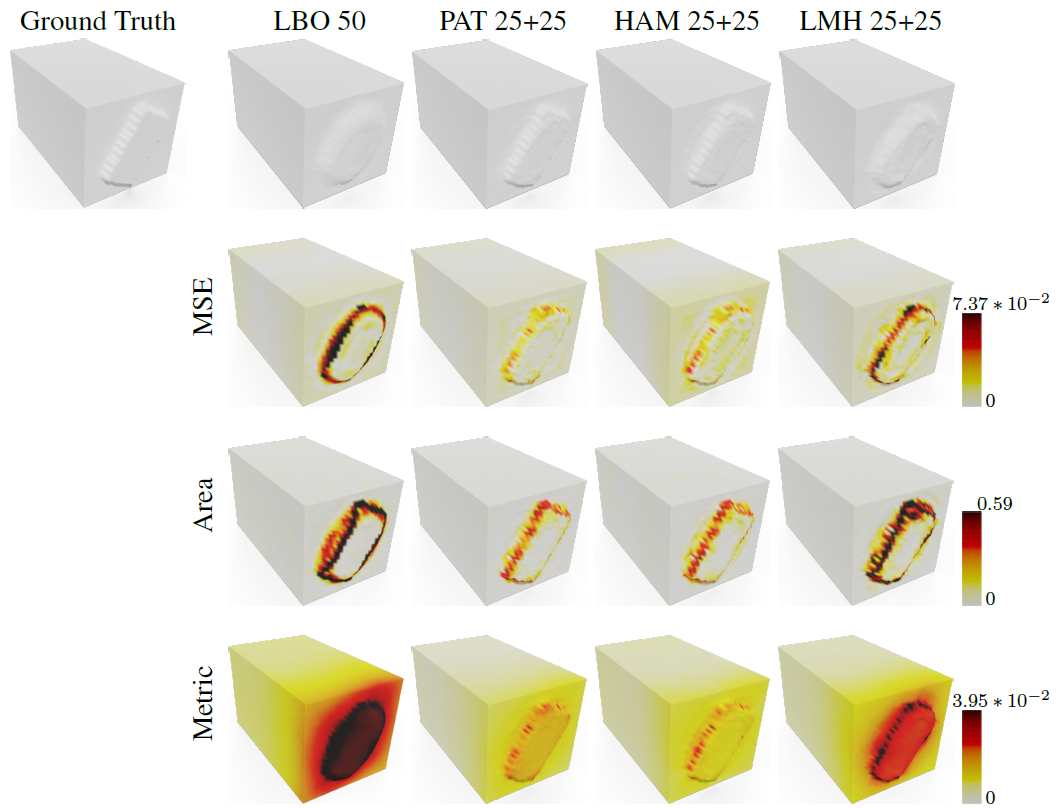}
\end{overpic}
\caption{Qualitative results on the CUBE dataset. In the top left we display the ground truth shape.
    In the first row there are the reconstructions of different methods. In the following rows we plot on the reconstruction the extrinsic (MSE) and intrinsic (Area,Metric) measures.
     Errors are color-coded, growing from white to dark red.}
    \label{fig:cubeError}    
\end{figure}

\subsection{CUBE}
With the CUBE dataset, we want to test how our model associates the orthogonal set of global and local variations with respect to the spectra. 

The first three rows of Table \ref{tab:errorsTOY} shows the results of the global spectrum with a different number of eigenvalues: LBO $k$. All the methods have a similar {MSE} with LBO 50 slightly better. 
The slightly higher error of LBO 80 may be due to the increasing uncertainty that is generated when we compute a higher frequency. 
In fact, in the computation of the LBO, the higher is the eigenvalue and the higher is the error that can be added to the computation.  
This factor encourages us to keep a lower number of global eigenvalues, while adding eigenvalues from a local spectrum.  

In \figref{fig:cubeError}, we can see a qualitative example.
In each row, we plot a different error on 4 different methods reported in Table \ref{tab:errorsTOY}: LBO 50, PAT 25+25, HAM 25+25, LMH 25+25.
Overall the error accumulates on the border of the extrusion. This is due to the difficulty of the decoder to generate steeper details. LBO50 produces the smoothest result: we believe that this is linked to the absence of high frequency on the input spectrum. 
The MSE metric mildly highlights the back face of the cuboids, while the Area metric only concentrates on the pattern face. Even if, in Table \ref{tab:errorsTOY}, the MSE and Area mean error correlates,  these qualitative differences allow us to distinguish the cause of the errors. In fact, we believe that the error on the back face is an accumulation error due to the difficulty of our model to stretch the triangles of the mesh.

This example supports our ideas on the limits of MSE. It represents a global measure that mixes up the structure and position of the shape. Therefore it is a good measure to validate the global accuracy of the reconstruction, but it hides where locally the reconstruction is better.  

\begin{table*}[t!]
\centering
\footnotesize
\resizebox{\textwidth}{!}{
\begin{tabular}{ k C C C C C C C C C C}
\toprule
  {  Method} &
  {  \begin{tabular}[c]{c} MSE \\ ($*10^{-6}$)\end{tabular}} & 
  {  \begin{tabular}[c]{c} MSE-$\subX$ \\ ($*10^{-6}$) \end{tabular}} & 
  {  \begin{tabular}[c]{c} MSE-$\subX^C$ \\ ($*10^{-6}$) \end{tabular}} &
  {  \begin{tabular}[c]{c} ENN \\ ($*10^{-6}$) \end{tabular}} & 
  { \begin{tabular}[c]{c} \hspace{-0.4cm} EM \textless ENN \end{tabular}} &
  { \begin{tabular}[c]{@{} c@{}} Area \\ ($*10^{-3}$)\end{tabular}} &
  { \begin{tabular}[c]{@{} c@{}} Area-$\subX$ \\ ($*10^{-3}$)\end{tabular}} &
  { \begin{tabular}[c]{@{} c@{}} Metric  \\ ($*10^{-3}$)\end{tabular}} &
  { \begin{tabular}[c]{@{} c@{}} Metric-$\subX$ \\ ($*10^{-3}$)\end{tabular}}\\
  \midrule[0.09em]

  {  \begin{tabular}[c]{@{} c@{}}LBO 30\end{tabular}} &
  {  1.7} & {  2.12} & {  1.61} &
  {  15.4} & {  99.57\%} &
  {8.19} & {10.1} & {2.62} & {5.23}\\ \midrule[0.07em] \midrule[0.07em] 

  {  \begin{tabular}[c]{@{} c@{}}PAT 25+5\end{tabular}} &
  {  1.1} &  {  1.42} & {  1.03} &
  {  19.6} & {  {100\%}} &
  {15} & {23.9} & {2.21} & {3.32}\\ \hline

  {  \begin{tabular}[c]{@{} c@{}}PAT 20+10\end{tabular}} & 
  \third{  0.77} & {0.75} & \third{ 0.77} & 
  {  28.8} &  {  { 100\%}} &
  {5.24} & \second{5.09} & {1.87} & {\first{2.25}}\\ \hline

  {  \begin{tabular}[c]{@{} c@{}}PAT 15+15\end{tabular}} &
  {  \first{0.71}} &  { \first{0.5}} &  \second{ 0.76} &
  {  39.1} &  {  {100\%}} &
  {\first{4.58}} & {\first{4.94}} & {\first{1.81}} & \second{2.27}\\ \midrule[0.07em] \midrule[0.07em]

  {  \begin{tabular}[c]{@{} c@{}}HAM 25+5\end{tabular}} &
  {  0.98} & {  1.13} & { 0.95} &  { 19} &  {  { 100\%}} &
  {5.33} & {5.65} & \third{1.93} & {2.47}\\ \hline

  {  \begin{tabular}[c]{@{} c@{}}HAM 20+10\end{tabular}} &
  \second{  0.73}& \third{  0.67} & {  \textbf{0.75}} &
  {  27.4} &  {  { 100\%}} &
  \third{5.11} & \second{5.09} & \second{1.86} & \third{2.29}\\ \hline

  {  \begin{tabular}[c]{@{} c@{}}HAM 15+15\end{tabular}} &
  {  0.86} & \second{ 0.57} & { 0.92} &
  {  38.9} & { { 100\%}} &
  \second{5.10} & \third{5.21} & {2.11} & {2.37}\\ \midrule[0.07em] \midrule[0.07em]

  {  \begin{tabular}[c]{@{} c@{}}LMH 25+5\end{tabular}} &
  { 2.7} & {  2.49} & {  2.75} &
  {  29.9} &  {  { 100\%}} &
  {10.6} & {11.7} & {3.37} & {5.71}\\ \hline
  {  \begin{tabular}[c]{@{} c@{}}LMH 20+10\end{tabular}} &
  { 2.4} & { 2.06} &  { 2.47} &
  { 32.5} & { { 100\%}} &
  {9.64} & {8.62} & {3.26} & {3.97}\\ \hline
  {  \begin{tabular}[c]{@{} c@{}}LMH 15+15\end{tabular}} &
  {  1.5} &  { 0.98} & {  1.61} &
  {  31.2} & { { 100\%}} &
  {7.15} & {6.16} & {2.62} & {2.67}\\ \midrule[0.07em] \midrule[0.07em]

{ \begin{tabular}[c]{@{} c@{}}\cite{InstantRecovery}$_{big}$ \\ LBO 30\end{tabular}} & {2.51} & {2.26} & {2.56} &  15.09 & 99.96\%   & 15.74 & 22.87 & 4.35 & 7.24 \\ \hline
{\begin{tabular}[c]{@{} c@{}}\cite{InstantRecovery}$_{big}$ \\PAT 15+15\end{tabular}} & {2.51} & {1.8} & {2.67} & 51.3 & { 100\%} & 16.9 & 22.25 & 4.52 &  9.28 \\ \hline
{ \begin{tabular}[c]{@{} c@{}}\cite{InstantRecovery} \\ PAT 15+15\end{tabular}} & {3.1} & {3.89} & {2.92} & 51.3 & { 100\%} & 19.81 & 32.93 & 4.46 & 8.44 \\

\bottomrule
\end{tabular}}
\caption{Reconstruction error on SURREAL test set.
 For each column, we highlighted the top three results in red with decreasing intensity.}
\label{tab:errorsSURREAL}
\end{table*}
\def\lenC{0.18\textwidth}
\def\lenColorBar{0.02\textwidth}
\def\lenGTspace{0.02\textwidth}

\begin{figure}[t!]
\begin{overpic}
    [trim=0cm 0cm 0cm 0cm,clip,width=1\linewidth]{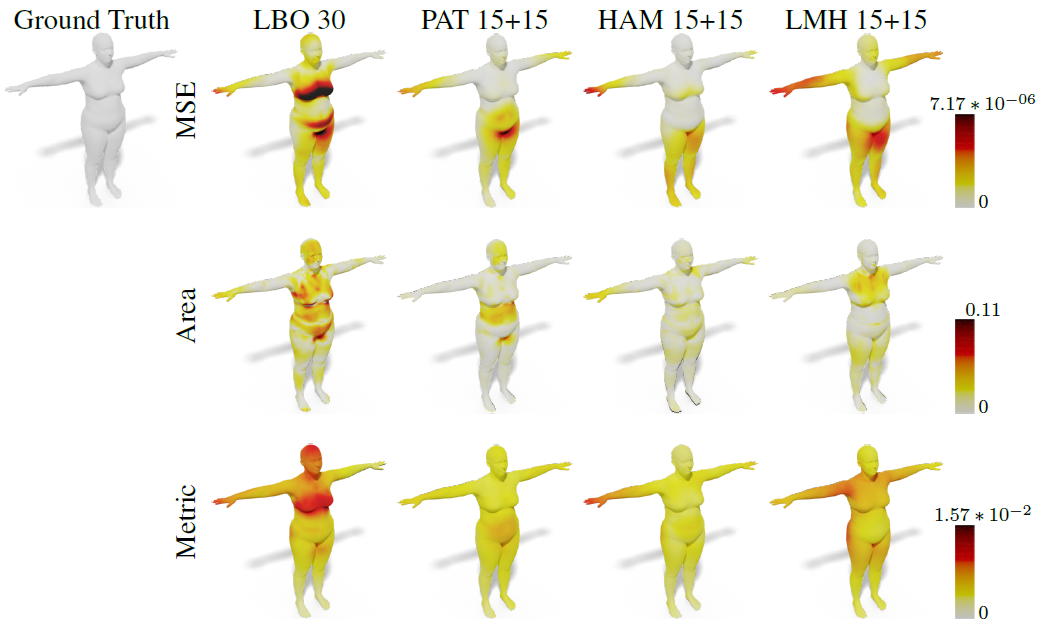}
\end{overpic}
    
    \caption{Qualitative result on the SURREAL dataset. The errors are shown on the reconstructed surfaces of a female with encoded color growing from white to dark red.}
    \label{fig:surrealError}
\end{figure}

\begin{figure}[t!]
   \begin{overpic}
    [trim=0cm 0cm 0cm 0cm,clip,width=1\linewidth]{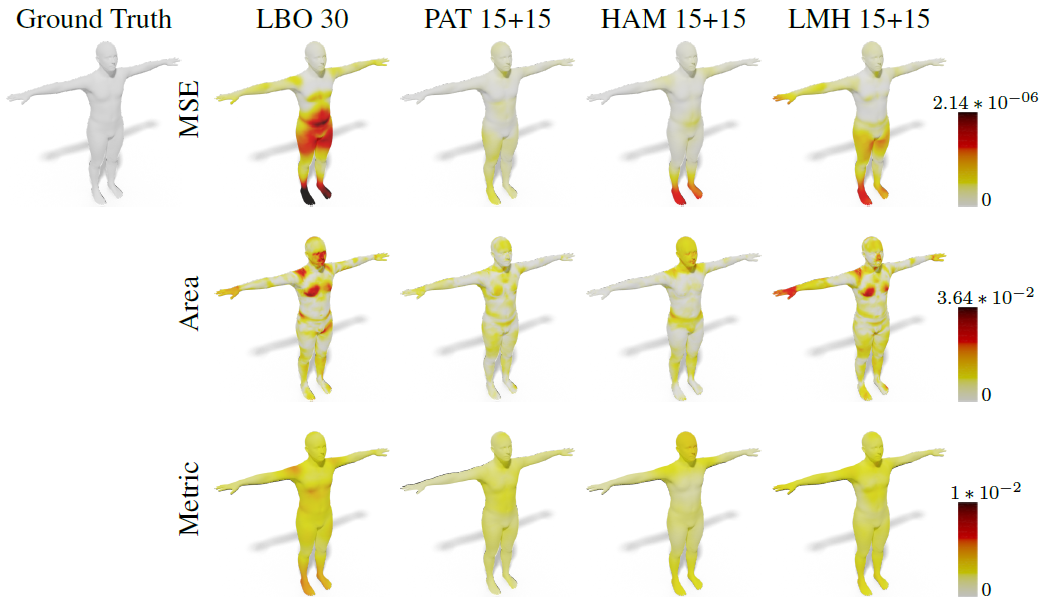}
\end{overpic}
    
    \caption{Qualitative result on the SURREAL dataset. The errors are shown on the reconstructed surfaces of a male with encoded color growing from white to dark red.}
    \label{fig:surrealError2}
\end{figure}

\begin{figure*}[t!]
\centering
            \begin{overpic}
            [trim=0cm 0cm 0cm 0cm,clip,width=0.79\linewidth, grid=false]{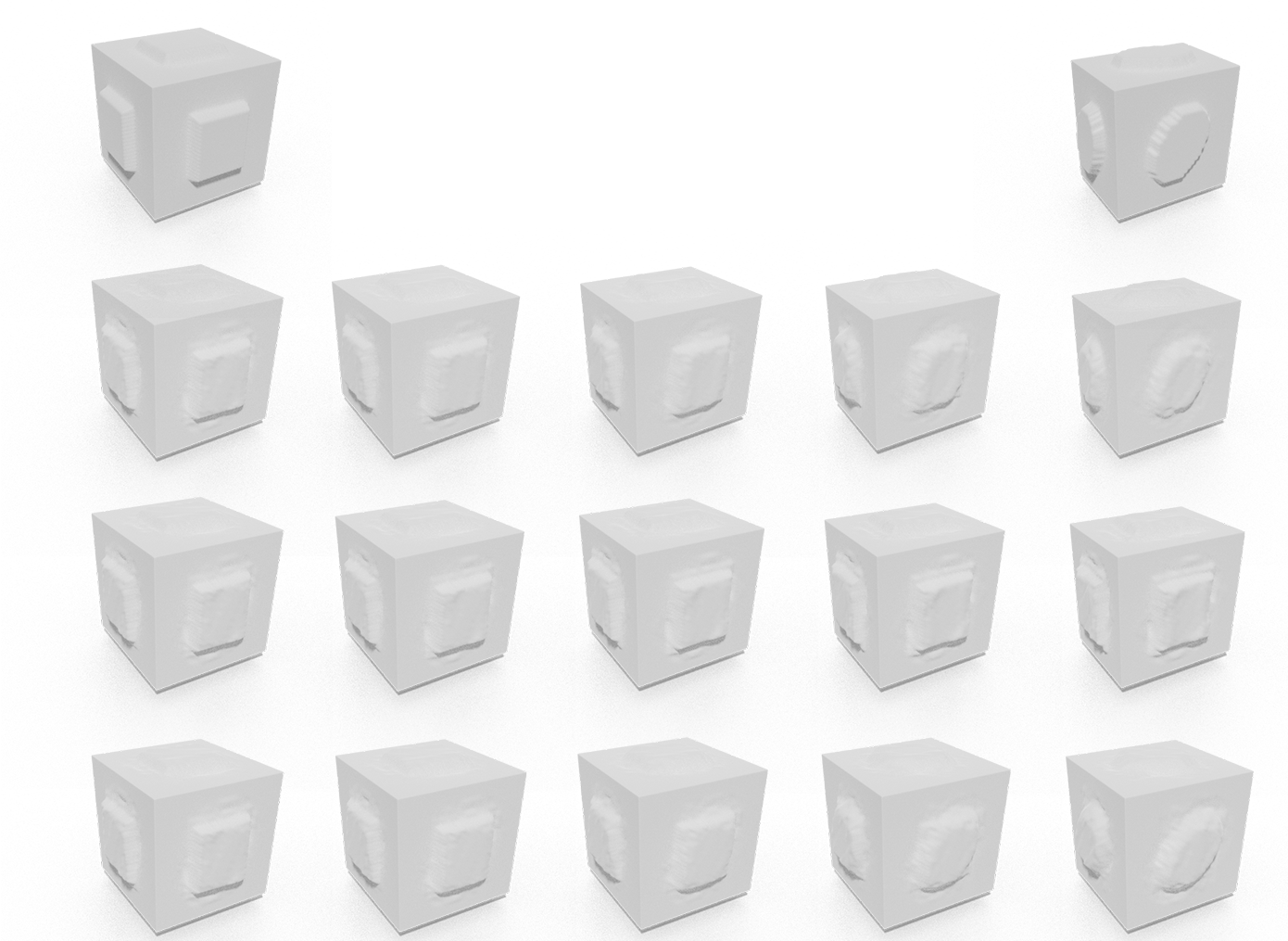}
            \put(4,53){$d\Lambda$}
            \put(4,35){$d\Lambda_{\X}$}
            \put(4,17){ $d\Lambda_{\subX}$}
            \put(11,72){Start}
            \put(87,72){End}
            \end{overpic}
        \caption{\revision{Interpolation results for a pair of cubes with same patterns applied to the all the faces.
        {\em First row}: Two input cubes.
        {\em Second row}: Interpolation of the entire encoding. 
        {\em Third row}: Interpolation of the global part only; observe how the patterns on the faces do not change, while the volume of the entire cube is correctly interpolated.
        {\em Last row}: Interpolation of the local part of the encoding (red in the bar plots), inducing a change in the patterns only.}}
        \label{fig:interpolationCUBEr_full}
\end{figure*}

\revision{In Figure \ref{fig:interpolationCUBEr_full}, we show the complete experiments of Figure 7 in the main paper. We trained our model with the same parameters on a second version of the CUBE dataset, where all the cube's faces manifest the same pattern.
With this modification, a correlation between the selected region and the rest of the cube exists. Our method (second row) can effectively learn this correlation by modifying at the same time the shape of the local pattern on all the faces and the depth of the cube. The global interpolation (third row) only changes the cube's length and leaves the pattern on the faces unchanged. Viceversa, the local interpolation (fourth row) changes as aspected the pattern in the selected region and the ones in the other faces but preserves the same cube's depth.
These results confirm that our encoding can control the factors of variation both when they are correlated and when they are not, generating a shape that maintain a global stylistic coherence. 
}

\subsection{SURREAL}
The SURREAL dataset is our first case of realistic data. 
We perform an analysis similar to before. Firstly, we test our model with a global spectrum only, first row in Table \ref{tab:errorsSURREAL}. Then we consider different combinations of global and local spectra computed from different localized operators. In this analysis, we concentrate more on the performance of different localized operators.

As in the CUBE dataset, the local addition improves both the {MSE} and the intrinsic metrics. In particular, the head region reaches an error lower than the rest of the body.
PAT 15+15 is the best combination, just followed by HAM 20+10. 
These results confirm once again the Laplace-Beltrami operator computed on a patch and the hamiltonian operator as the best-localized operators. 
Moreover, a combination of global and local representations in which they have similar proportions seems to hold the best results. 

In this case, the {ENN} errors don't change drastically between global and localized operators, but it is still significantly higher than the {MSE}. The {EM < ENN} accuracy is $100\%$ in all tests, except in LBO30. 
The similar values of {ENN} in this dataset allows us to make a clearer interpretation with respect to the one done with the CUBE dataset. 
In fact, while in the CUBE dataset the shapes are simple cuboids with details only on one face, the shapes in this dataset are more complex and have details that can vary all over the surface. Then, since in SURREAL the LBO can encode more variations at a global level than in the CUBE dataset, the ENN error is already higher in the global case and slightly increase in the local ones.
This observation highlights the quantity of information encoded in a spectrum when the shape has greater or lesser details spread across the surface.

Qualitative examples of the results in Table \ref{tab:errorsSURREAL} can be found in \figref{fig:surrealError} and in \figref{fig:surrealError2}.
The former is a more robust woman, while the latter is a thin male.
All the combinations are able to create shapes qualitatively similar to the ground truth.
The addition of a local spectrum computed on the head greatly improves the reconstruction not only of the head, but also on the torso. 
We believe that our model learns to associate the information encoded on the head spectrum with other important features of a subject such as his robustness. Similar to the example in \figref{fig:cubeError}, the Area errors accumulate in different parts than the MSE, allowing us to separate reconstruction errors due to the position in the space from the ones due to the "topology" of the shape. For instance, the errors in the hands are high in the MSE metric but low in Area one. This suggests that locally the hands are well reconstructed but globally, they are not in the right position because of an accumulation error on the vertices of the arm.

In the last three rows of Table \ref{tab:errorsSURREAL} we report the errors obtained both with the best spectra combinations (PAT 15+15) and the global spectrum only.
Results show not only that our decoder approach is better than the full architecture even in the LBO setting, but that~\cite{instant2020} is not equally capable of combining local information with its latent space.

\begin{table*}[t!]
\centering
\resizebox{\textwidth}{!}{
\begin{tabular}{ k C C C C C C C C C C C }
\toprule
  {  Method} &
  {  \begin{tabular}[c]{@{} c@{}} MSE \\($*10^{-6}$)\end{tabular}} &
  { \footnotesize \begin{tabular}[c]{@{} c@{}} MSE-$\mathcal{H}$ \\($*10^{-6}$)\end{tabular} } &
  { \footnotesize \begin{tabular}[c]{@{} c@{}} MSE-$\mathcal{T}$ \\($*10^{-6}$)\end{tabular}} &
  {  \begin{tabular}[c]{@{} c@{}} ENN \\($*10^{-6}$)\end{tabular}} &
  { \footnotesize \begin{tabular}[c]{c} \hspace{-0.4cm} EM \textless ENN \end{tabular}} &
  { \begin{tabular}[c]{@{} c@{}} Area  \\($*10^{-2}$)\end{tabular}} &
  { \footnotesize \begin{tabular}[c]{@{} c@{}} Area-$\mathcal{H}$ \\($*10^{-2}$)\end{tabular}} &
  { \footnotesize \begin{tabular}[c]{@{} c@{}} Area-$\mathcal{T}$ \\($*10^{-2}$)\end{tabular}} &
  { \begin{tabular}[c]{@{} c@{}} Metric \\($*10^{-3}$)\end{tabular}} &
  { \footnotesize \begin{tabular}[c]{@{} c@{}} Metric-$\mathcal{H}$ \\($*10^{-3}$)\end{tabular}} &
  { \footnotesize \begin{tabular}[c]{@{} c@{}} Metric-$\mathcal{T}$ \\($*10^{-3}$)\end{tabular}} \\ 
  \midrule[0.09em]
  
  { LBO 30} & 
  \third{ 1.39} &  \third{1.48 } & \third{4.30} & 
  {3.84} & { 80.2\%} &
  \third{1.90} & {2.58} & \third{4.25} & 
  \third{3.58} & \third{7.40} & \third{23.6} \\ 
  \midrule[0.07em] \midrule[0.07em]
  
  {PAT$_H$ 15+15} &
  {\first{1.08}} &  {\first{1.07} } & {\first{3.07}} & 
  {5.47} & { 86.6\%} &
  {\second{1.63}} & {\second{2.12}} & {\second{3.83}} & 
  {\second{3.05}} & {\second{5.96}} & {\second{20.3}} \\ 
  \hline

  {PAT$_T$ 15+15} &
  {3.93} &  {4.20 } & {11.5} & 
  {11} & { 64.6\%} &
  {3.05} & {4.03} & {6.26} & 
  {6.13} & {14.1} & {36.9} \\ 
  
  
  \midrule[0.07em] \midrule[0.07em]
  
    {HAM$_H$ 15+15} &
  \second{1.11} &  \second{1.13} & \second{3.50} & 
  {5.22} & {86.6\%} &
  {\first{1.54}} & {\first{2.01}} & \first{3.59} & 
  \first{3.01} & \first{5.77} & \first{19.6} \\ 
  \hline
      {HAM$_T$ 15+15} &
  {3.13} &  {3.35} & {8.81} & 
  {8.17} & { 68.2\%} &
  {2.77} & {3.68} & {5.75} & 
  {7.5} & {13} & {32.9} \\ 
  \midrule[0.07em] \midrule[0.07em]
  
    {LMH$_H$ 15+15} &
  {1.78} &  {1.9} & {5.11} & 
  {4.47} & { 82\%} &
  {1.93} & \third{2.52} & {4.47} & 
  {3.62} & {7.6} & {24.5} \\ 
  \hline
      {LMH$_T$ 15+15} &
  {3.57} &  {3.8} & {9.76} & 
  {13} & { 67.4\%} &
  {2.94} & {3.93} & {5.9} & 
  {5.89} & {12.6} & {35.8} \\ 
 
\bottomrule
\end{tabular}}
\caption{Reconstruction error on the SMAL test set.
For each column, we highlighted the top three results in red with decreasing intensity.}
\label{tab:errorsSMAL}
\end{table*}

\def\lenC{0.17\textwidth}
\def\lenColorBar{0.02\textwidth}
\def\lenGTspace{0.02\textwidth}

\begin{figure}[t]
    \begin{overpic}
    [trim=0cm 0cm 0cm 0cm,clip,width=1\linewidth]{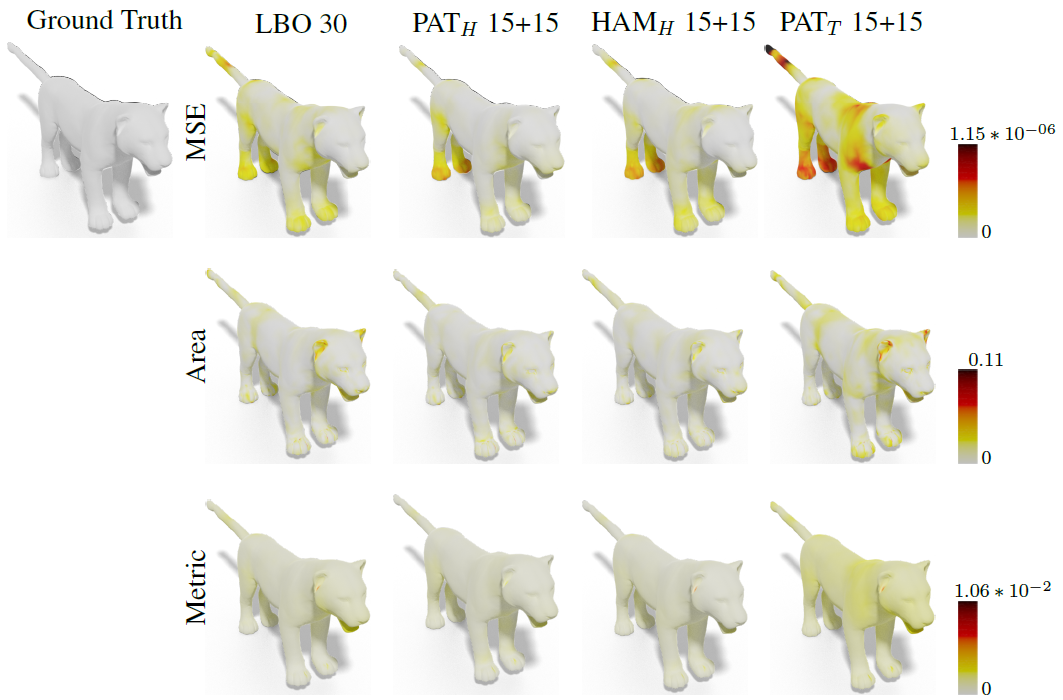}
\end{overpic}
    \caption{Qualitative results on the SMAL dataset with extrinsic and intrinsic measures plotted on the reconstructed surface of a tiger with a white to dark red colormap.}
    \label{fig:cubeSMAL}
\end{figure}

\def\lenMinipage{0.48\textwidth}
\def\lenC{0.19\textwidth}
\def\lenColorBar{0.02\textwidth}
\def\lenGTspace{0.06\textwidth}
\begin{figure}
    \centering
    \begin{minipage}{\lenMinipage}
    \centering
\begin{overpic}
    [trim=0cm 0cm 0cm 0cm,clip,width=1\linewidth]{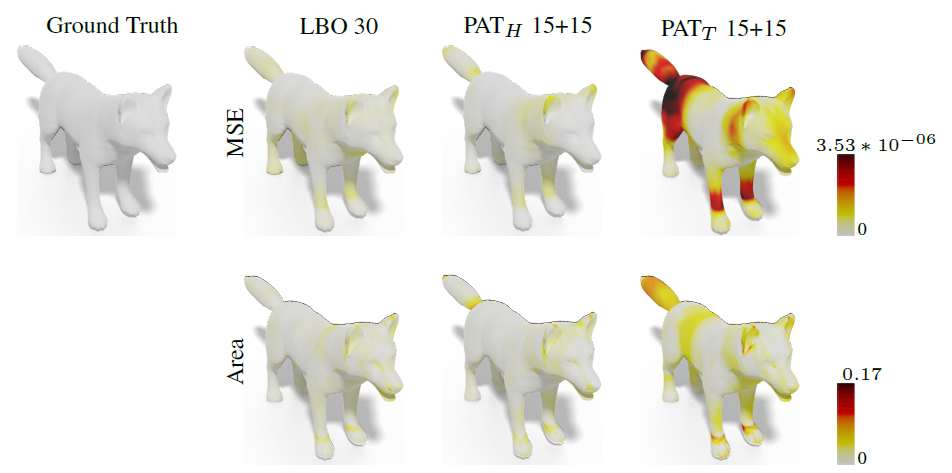}
\end{overpic}
    \subcaption{Wolf}
    \end{minipage}

    \begin{minipage}{\lenMinipage}
    \centering
       \begin{overpic}
    [trim=0cm 0cm 0cm 0cm,clip,width=1\linewidth]{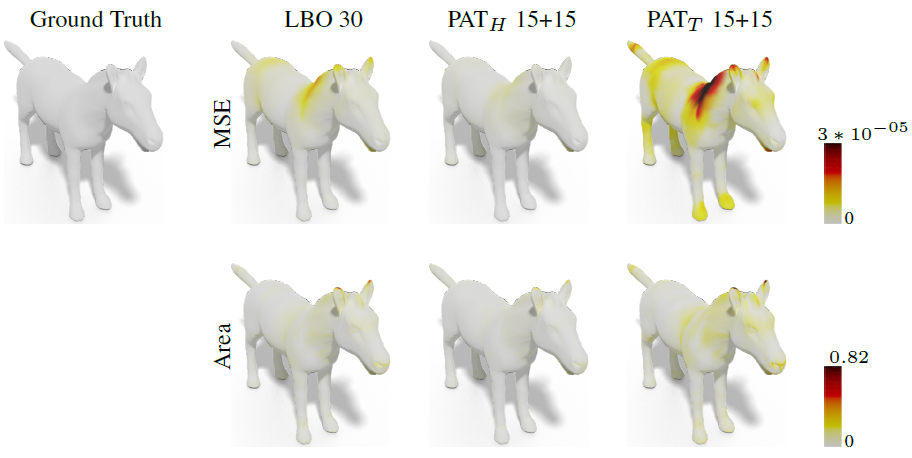}
\end{overpic}
    \subcaption{Zebra}
    \end{minipage}
    
    \begin{minipage}{\lenMinipage}
    \centering
       \begin{overpic}
    [trim=0cm 0cm 0cm 0cm,clip,width=1\linewidth]{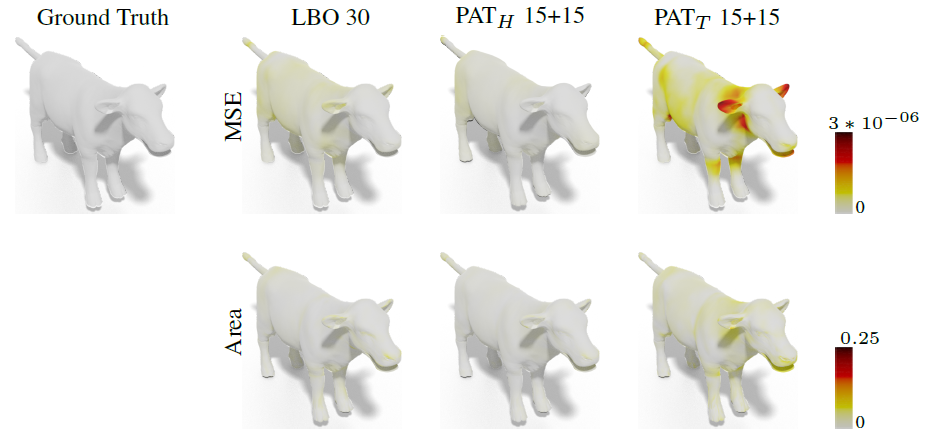}
\end{overpic}
    \subcaption{Cow}
    \end{minipage}

    \begin{minipage}{\lenMinipage}
    \centering
        \begin{overpic}
    [trim=0cm 0cm 0cm 0cm,clip,width=1\linewidth]{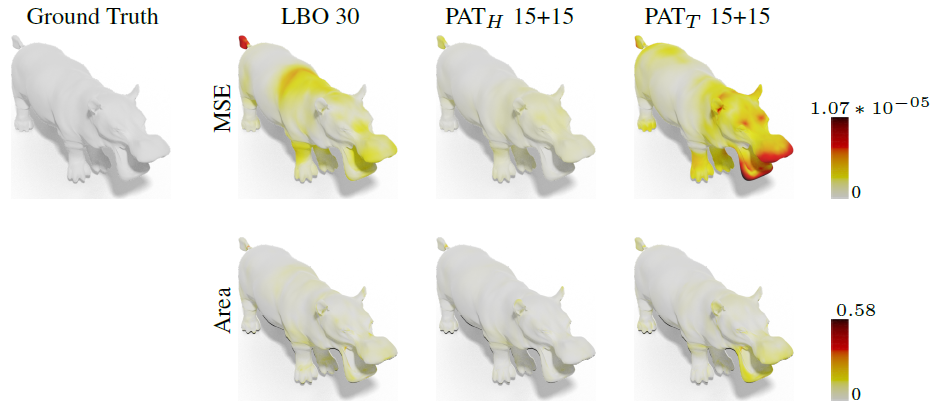}
\end{overpic}
    \subcaption{Hippo}
    \end{minipage}

    \caption{Qualitative results on the SMAL dataset with extrinsic and intrinsic measures plotted on the reconstructed surface of different classes with a white to dark red colormap.}
    \label{fig:cubeSMAL2}
\end{figure}
\subsection{SMAL}
Since animals have several regions which may encode some shape semantic (e.g., tail, head, paws, ...), we used SMAL to focus our analysis on the contribution of different local regions. In particular, we localize the operators on both the head and tail to see their impact on the generation capacity.
For this reason, we modify the taxonomy of the Table \ref{tab:errorsSMAL}:
we consider two distinct regions error $\mathcal{H}$ and $\mathcal{T}$ that correspond respectively to the head and tail of the shape; we differentiate our model trained on a different region with the subscript $_H$ for the head region and $_T$ for the tail.
We also change the columns of the errors computed on a portion of the surface. The {MSE-$\subX$} column is replaced with {MSE-$\mathcal{H}$} that represents the error computed on the head region; while the {MSE-$\subX^C$} column is replaced with {MSE-$\mathcal{T}$} that represents the error computed on the tail region. The same applies to the {Area} and {Metric} columns.

A difference with respect to the previous results is the {ENN}. Its values are higher than MSE, but with a lower gap. Moreover, the {EM$<$ENN} accuracy does not reach the $100\%$ in any test.
The lower gap could suggest that the spectra combination produces less distinctive encoding in this dataset. Therefore our model has more difficulty generating a more accurate shape.
We also remember that SMAL is composed of different animal species and therefore has a higher variation. 
This characteristic allows the decoder to give more importance to the data encoded in the global spectrum since there are shapes that differ also in their global structure, i.e. an hippo compared to a tiger is shorter and bigger.
This deduction is also enforced by the short gap between LBO 30 and PAT$_H$ $15+15$ where the addition of the local spectrum brings less information with respect to the humans of SURREAL. 

\figref{fig:cubeSMAL} shows the different errors plotted on a tiger. Overall, all the methods generate shapes that are qualitative very similar to the ground truth.  
PAT $_H$ $15+15$ has lower results not only on the head, but also on the tail. This suggests a correlation between the two regions.
On the contrary PAT$_T$ $15+15$ has worse results. In particular, even though the local spectrum is computed on the tail, the error on that region is higher. 
This may be due to the inability of the tail spectrum to encode enough information relevant to the whole shape like the head does. As a consequence, the decoder has fewer informative features to generate the whole shape causing a sparse error that in the MSE highly accumulates on the tails. 

Other qualitative examples from the SMAL dataset are reported in \figref{fig:cubeSMAL2}.  
We show the MSE and Area measures of LBO 30, PAT$_H$ $15+15$ and PAT$_T$ $15+15$ on all the remaining classes. 
The error accumulates mainly on the characteristic region of the different animals: the tail and ears in the wolf, the ears and crest in the zebra, the ears in the cow and the snout in the hippo.

\subsection{Main Insights of our analysis}
We summarize the main insights of our analysis:
 
\vspace{1ex}\noindent\textbf{\em Global versus Local.} The standard eigenvalue representation of~\cite{instant2020} is outperformed by mixed PAT encoding, regardless of different $k+h$ values. Nevertheless, even in the presence of orthogonal transformations, introducing a local representation helps the global reconstruction as well. We believe this is possible only if the network can relate each operator spectrum with a shape variation.

\vspace{1ex}\noindent\textbf{\em Different localized operators.} We see that maximizing the locality of the considered representation is beneficial. PAT emerges as the best representation, tightly followed by HAM which is almost equivalent. 
LMH performs the worst. The discussion on this point is further detailed in Section 5.4.

\vspace{1ex}\noindent\textbf{\em Locality proportion.} Our experiments suggest that a balanced mix of global and local information provides the best representation for the inverse spectral problem. Also, increasing the number of eigenvalues is more beneficial in a mixed setup than in the standard LBO setup. We think the information rapidly faints in subsequent eigenvalues, while new operators provide a clearer pattern for the network to harvest.

\vspace{1ex}\noindent\textbf{\em Autoencoder vs decoder-only.} Results show not only that our decoder approach is better than the full architecture even in the LBO setting, but that~\cite{instant2020} is not equally capable of combining local information with its latent space.

\vspace{1ex}\noindent\textbf{\em Evaluation metrics.}We emphasize that extrinsic metrics are not always reflected in the intrinsic ones. While the extrinsic measures directly test the network on the purpose of its training, our measures also reflect the model's intrinsic properties. We find them complementary, and encourage follow-up works to rely on similar measurements both for training and test.
\subsection{Region selection results}
\label{sec:region}


        

\def\lenC{0.18\textwidth}
\def\lenFILL{0.04\textwidth}
\def\lenColorBar{0.02\textwidth}
\def\lenGTspace{0.09\textwidth}

\begin{figure}[t!]
\begin{overpic}
    [trim=0cm 0cm 0cm 0cm,clip,width=0.8\linewidth]{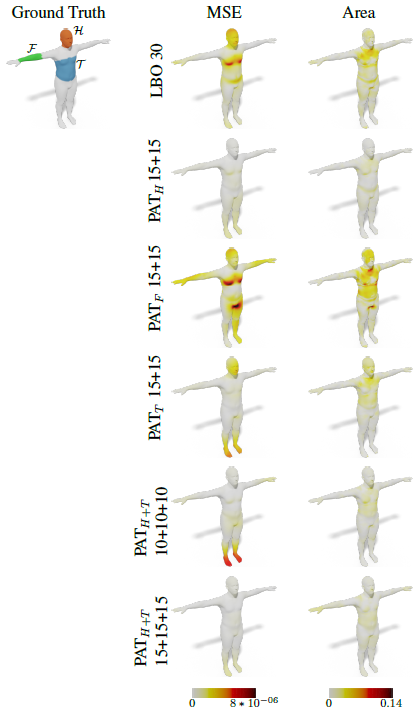}
\end{overpic}
     
    \caption{Qualitative comparisons between different regions on the SURREAL dataset.
    On the top left we display the ground truth with the different regions highlighted: in red the head $\mathcal{H}$;
     in green the forearm $\mathcal{F}$; in blue the torso $\mathcal{T}$. 
     The rows are different methods, while the columns different measures. 
     Errors are color-coded, growing from white to dark red.}
    \label{fig:RegionErrors}
\end{figure}
In this section, we want to investigate more deeply the importance of $\subX$

In Fig.~\ref{fig:RegionErrors}, we show the mean squared and the area error for an example of shape reconstruction comparing PAT$_{F}$ $15+15$ and PAT$_{T}$ $15+15$ with LBO $30$ and PAT $15+15$. 
Concerning LBO $30$, we see that PAT$_{F}$ $15+15$ and PAT$_{T}$ $15+15$ presents a similar intrinsic error on the head.
PAT$_{F}$ $15+15$ has a low Area error on the arms but high on the torso which produces a higher MSE also on the arm.
On the contrary, PAT$_{T}$ $15+15$ is able to improve the torso eliminating the LBO error on the chest.
In all cases, $PAT$ $15+15$ performs significantly better. 

\def\lenC{0.17\textwidth}
\def\lenColorBar{0.02\textwidth}
\def\lenGTspace{0.02\textwidth}
\begin{figure}[t]
   \begin{overpic}
    [trim=0cm 0cm 0cm 0cm,clip,width=1\linewidth]{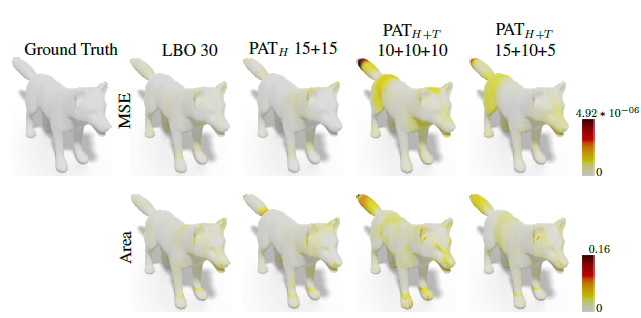}
\end{overpic}
            
    \caption{Qualitative results of different regions on the SMAL dataset.
    We plot on the reconstructed shape extrinsic (first row) and instrinsic (second row) measures with a colormap from white to dark red.}
    \label{fig:SMALMultiregion}
\end{figure}
\begin{figure}[]
\begin{center}
        \begin{overpic}
        [trim=0cm 0cm 0cm 0cm,clip, width=0.9\linewidth]{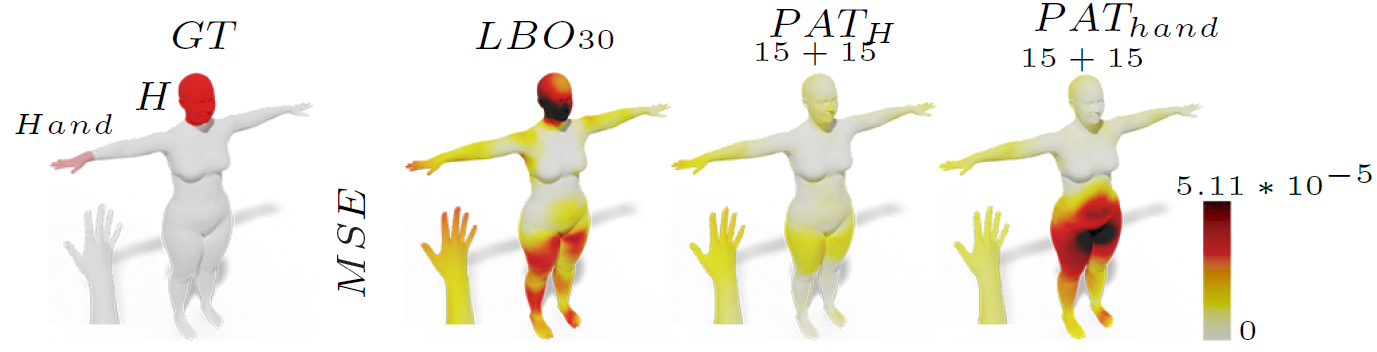}
        \end{overpic}

\end{center}
\caption{An example of reconstruction comparing head or hand selection.}
    \label{fig:RegionHand1}
\end{figure}
\begin{figure}[]
\begin{center}
        \begin{overpic}
        [trim=0cm 0cm 0cm 0cm,clip, width=0.9\linewidth]{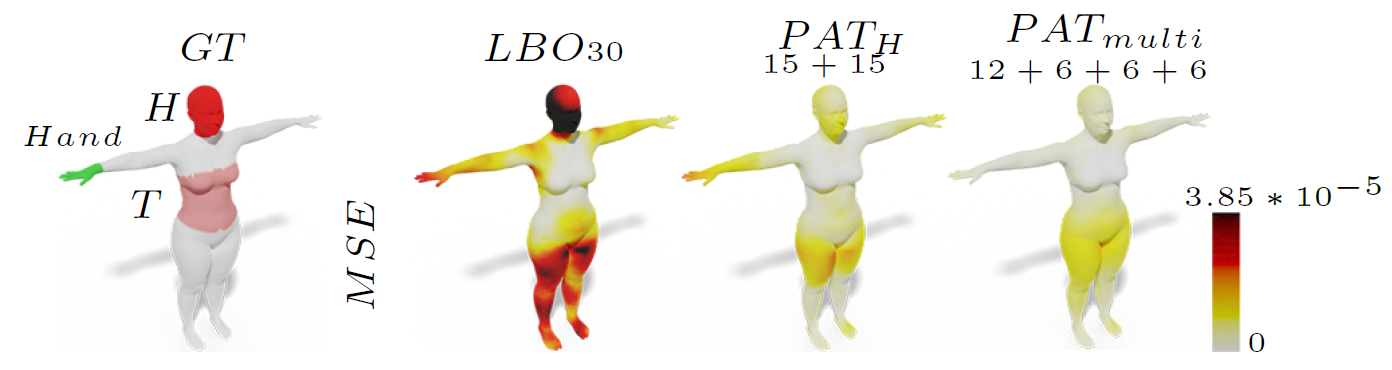}
\end{overpic}
\end{center}
\caption{An experiment of different representation for the input: on the right, it considers $12$ eigenvalue from the global spectrum, $6$ from the head, $6$ from the torso, and $6$ from the hand.}
    \label{fig:RegionHand2}
\end{figure}
\subsection{Multi-region}
\figref{fig:SMALMultiregion} shows a qualitative comparison of multi-region selection over SMAL. PAT$_{H+T}$ $15+10+5$ has a lower error than  PAT$_{H+T}$ $10+10+10$ especially on the back of the shape.
Since PAT$_{H+T}$ $15+10+5$ has a slightly lower MSE than PAT$_{H+T}$ $10+10+10$, we think that the performance may be correlated to the number of eigenvalues assign to each spectrum. In fact, the combination $15+10+5$ has fewer eigenvalues from the tail which is the region with the higher errors.
As consequence, the decoder has a more informative encoding from which generates the correct shape.

The PAT$_{H+T}$ $15+15+15$ combination seems to add enough information from all the spectra to improve the performance. In the last two columns of \figref{fig:RegionErrors} can be seen a significant improvement on the lower part of the body. 
 
This last series of experiments highlight the importance of $\subX$. Not all portions of $\X$ are good candidates as local regions since they don't have enough high-frequency information.
This suggests the necessity of research of meaningful area on which to compute the local spectra.
Moreover, the proportion of eigenvalues assigned to the different spectra affect just as much the overall performance.

\noindent \textit{Multiple-Local areas.} Splitting between regions and global representation requires a careful design. From a general perspective, our experiments show that substituting part of the global information with some local one also provides better reconstruction in the global areas for many different scenarios. This is strengthened by the correlation between the local and global parts.
However, pushing further the number of the local regions maybe not be trivial. As a complement of already seen results, we consider here (Figures \ref{fig:RegionHand1} and \ref{fig:RegionHand2}) the hand region: in the first row as the only local region, in the second one in conjunction of head and torso. These experiments convey all the same message: each part should be represented with enough information. A study on the perfect balance between regions would be domain-dependent and exciting for future works.

\clearpage 
\newpage
\section{Interactive manipulation}
\label{sec:sliders}
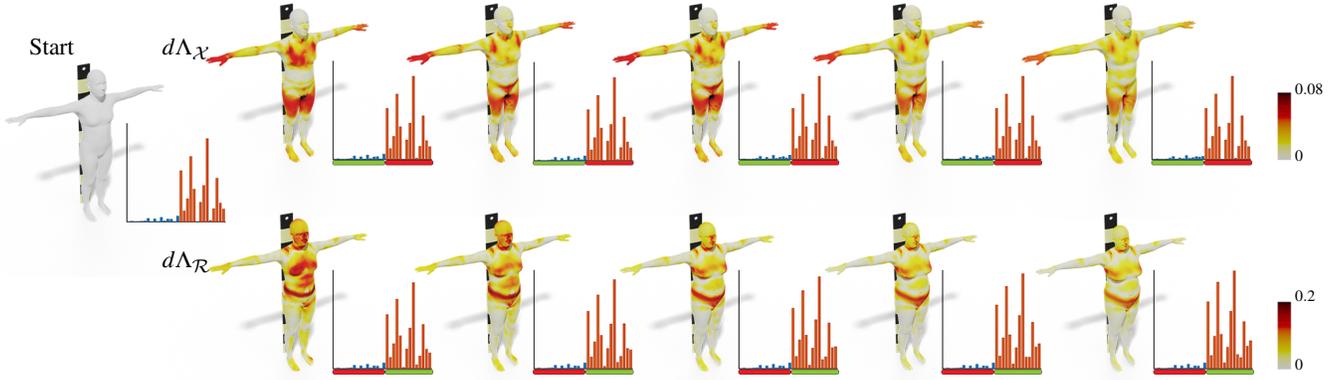
\begin{figure*}[t!]
\begin{center}
        \begin{overpic}
        [trim=0cm 0cm 0cm 0cm,clip,width=1\linewidth]{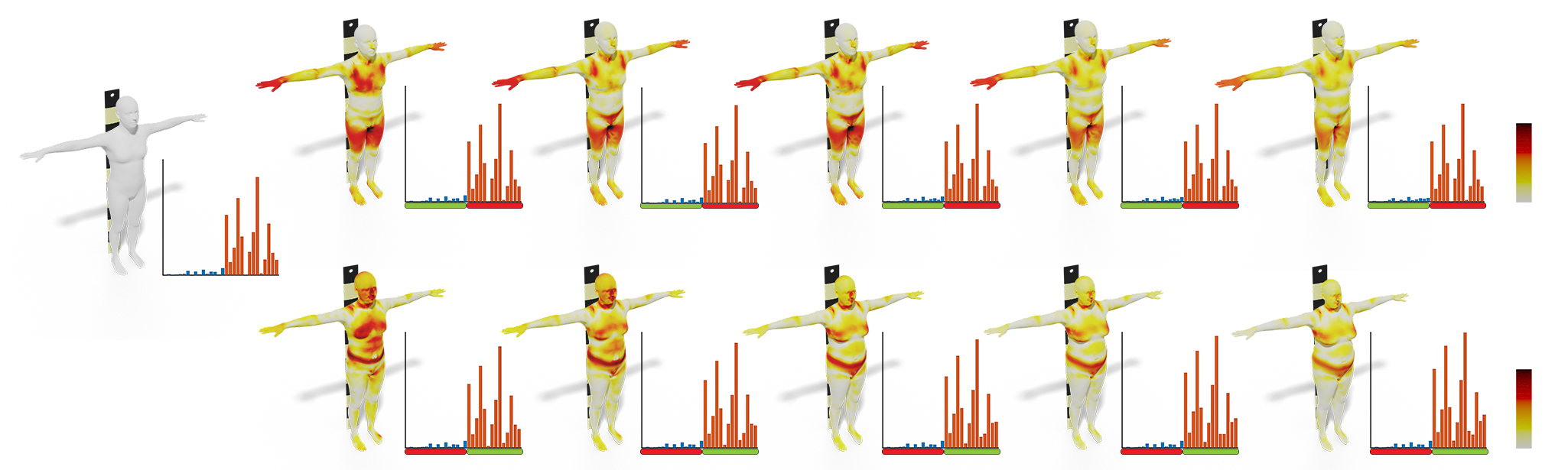}
        \put(98,22){\tiny $0.08$}
        \put(98,17){\tiny $0$}
        \put(98,6.5){\tiny $0.2$}
        \put(98,1.3){\tiny $0$}
        
        \put(13,25){$d\Lambda_{\X}$}
        \put(13,9){$d\Lambda_{\subX}$}
        \put(3,25){Start}

        \end{overpic}
\end{center}
    \caption{Free manipulation of the input. Given a shape (on the left), we decrease the global values ($\Lambda_{\X}$-first row) and increase the local values ($\Lambda_{\subX}$-second row) separately. For each shape, we plot the area variations for each vertex and show the correspondent encoding as barplot (blue:global, red:local). We highlight in green the values that changes and in red the values that we keep constant.}
    \label{fig:sliders2}
\end{figure*}
Our method can be efficiently used in real-time to synthetize shapes and control their deformations by acting on the different parts of our representation. An example can be seen in the \emph{video} attached to this document. In that short sequence, we have two sliders that modify the different components of the proposed spectral encoding. On the left side of the video, we show the spectrum that we input to the network, highlighting with brighter color the part subject to the current modification. In gray, we kept the original spectrum as a reference. On the right side of the video, we visualize the shape produced by our model. The color depicted on the surface encodes the difference between two subsequent modification frames; this visualization helps to identify where the modification represented by the sliders is acting on the $3$D geometry. Moreover, in Fig.~\ref{fig:sliders2} we report an illustrative image of free manipulation provided by our model. We chose a reference shape and modified its global and local encoding separately. Similarly to the video, for each shape, we highlight on the surface the area variations encoded by the colors. 
In the first row, we decrease the global part of the encoding generating alterations scattered on the body,
but with minimal interaction with the head. 
In the second row, we increase the values of the local part of our encoding obtaining a more feminine physiognomy and variations localized on the head and thorax, while the legs are almost left unchanged.




\section{Different representations from training time}
\label{sec:semantic}
\begin{figure}[!t]
\begin{center}
        \begin{overpic}
        [trim=0cm 0cm 0cm 0cm,clip,width=1\linewidth]{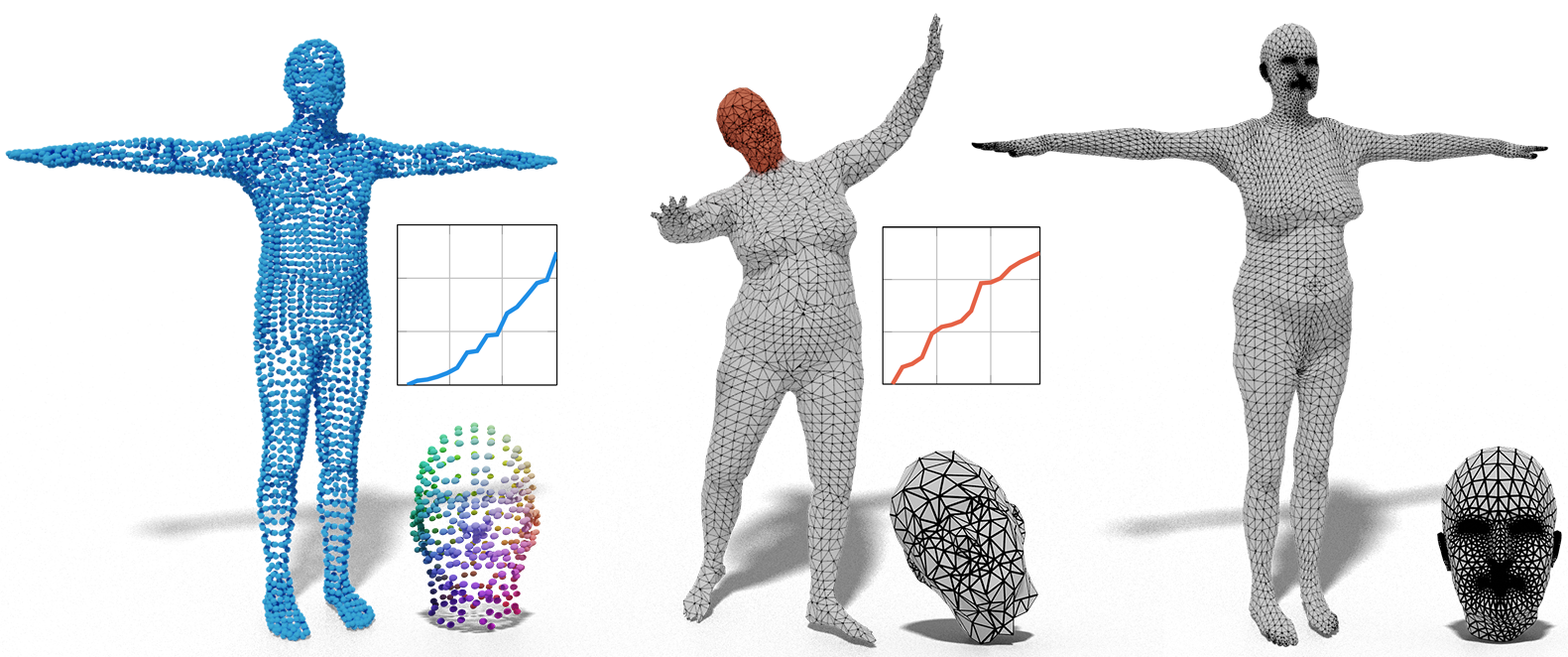}
        \put(37,20){\large $\bigoplus$}
         \put(69,20){\large$\mathbf{=}$}
        \end{overpic}
\end{center}
    \caption{Combination of global spectrum (in blue) of a point cloud (left) with a local spectrum (in red) from a mesh (center) with different discretization and pose.}
    \label{fig:teaserPCR2}
\end{figure}

In Fig.~\ref{fig:teaserPCR2} we report an additional result on semantic control with different representations. We start from a sparse point cloud ($3445$ vertices), depicted on the left, from which we compute the global spectrum with the robust Laplacian~\cite{Sharp2020} and combine it with the local spectrum from a mesh representing a different subject in a different pose, visualized in the middle. On the right we show the resulting shape, which maintains the identity of the second one, but with a thinner body like the first shape. We remark that the network is trained only on meshes; thus we appreciate the robustness of our model also to unseen and noisy data.

\section{Unorganized pointclouds}
\label{sec:airplanes}
Here we present other examples of our airplane experiments (Fig.~8 in the main paper). 

\begin{figure}
\footnotesize 
         \begin{overpic}
         [trim=0cm 0cm 0cm 0cm,clip,width=1\linewidth]{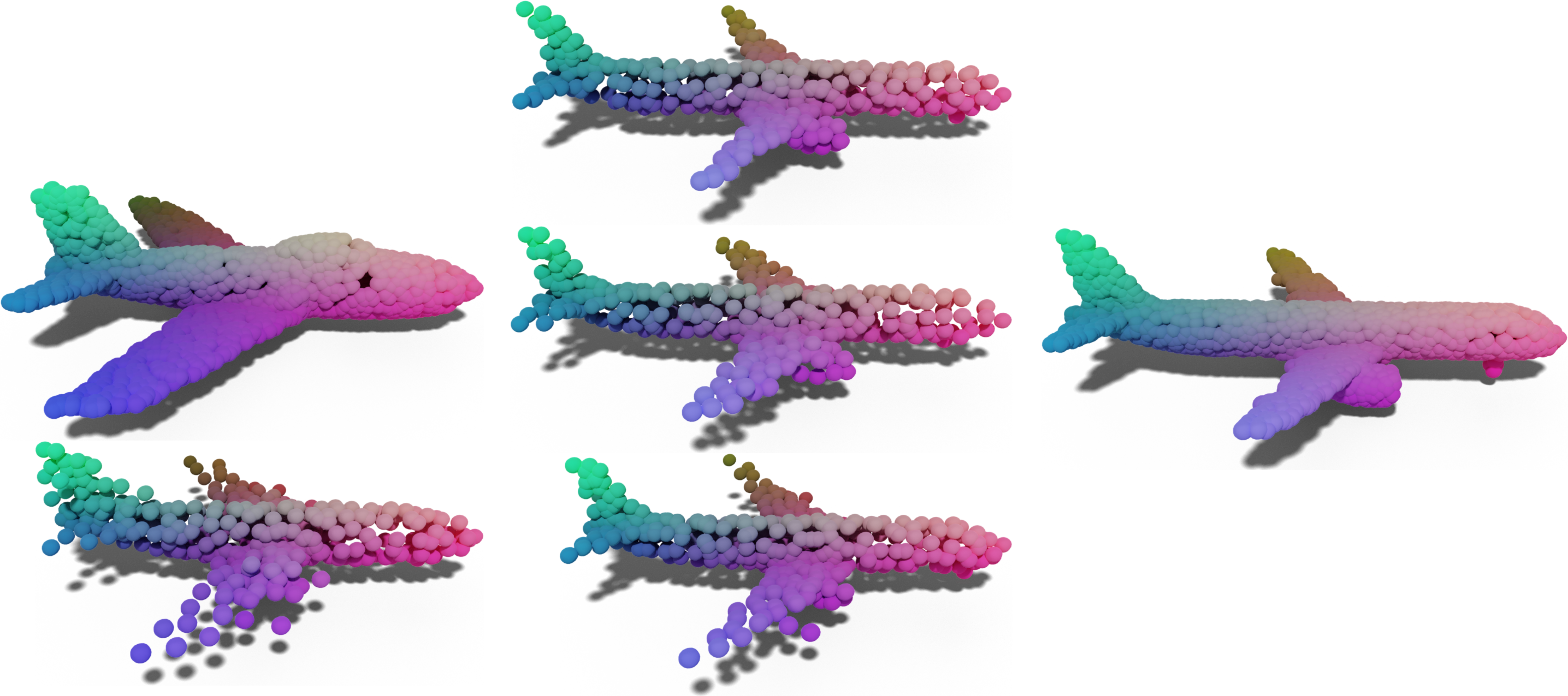}
         \put(20,32){Start}
         \put(20,13){Our}
         \put(52,12){$d\Lambda_{\subX}$}
         \put(52,27){$d\Lambda_{\X}$}
         \put(52,41){$d\Lambda$}
         \put(90,26){Final}
         \end{overpic}
     \caption{A spectrum switch, similar to the interpolation shown in the main manuscript. On the left: the starting airplane and the corresponding output generated by our network. On the right, the second airplane. In the middle, from the top there are the reconstruction generated: using the whole second spectral enconding ($d\lambda$); concatenating the global spectral encoding of the second with the local one of the first($\Lambda_{\X}$); concatenating the global spectral encoding of the first with the local spectral encoding of the second($\Lambda_{\subX}$). Notice how the first two impact the whole plane, while the third changes the tail and preserves the global structure (e.g., wings and turbines). }
     \label{fig:singleairplane}
\end{figure}

In Fig.~\ref{fig:singleairplane} we perform a spectrum switch. The two input planes have a similar tail but a different structure. 
Even in this subtle case, when we change only the local encoding, our method interpolates the two tails without modifying the airplane length, the presence of the turbines, and keeping the wings loyal to the starting plane. On the contrary, the global switch affects the whole plane like in the others interpolations experiments.

\begin{figure}
\begin{center}
        \begin{overpic}
        [trim=0cm 0cm 0cm 0cm,clip,width=1\linewidth]{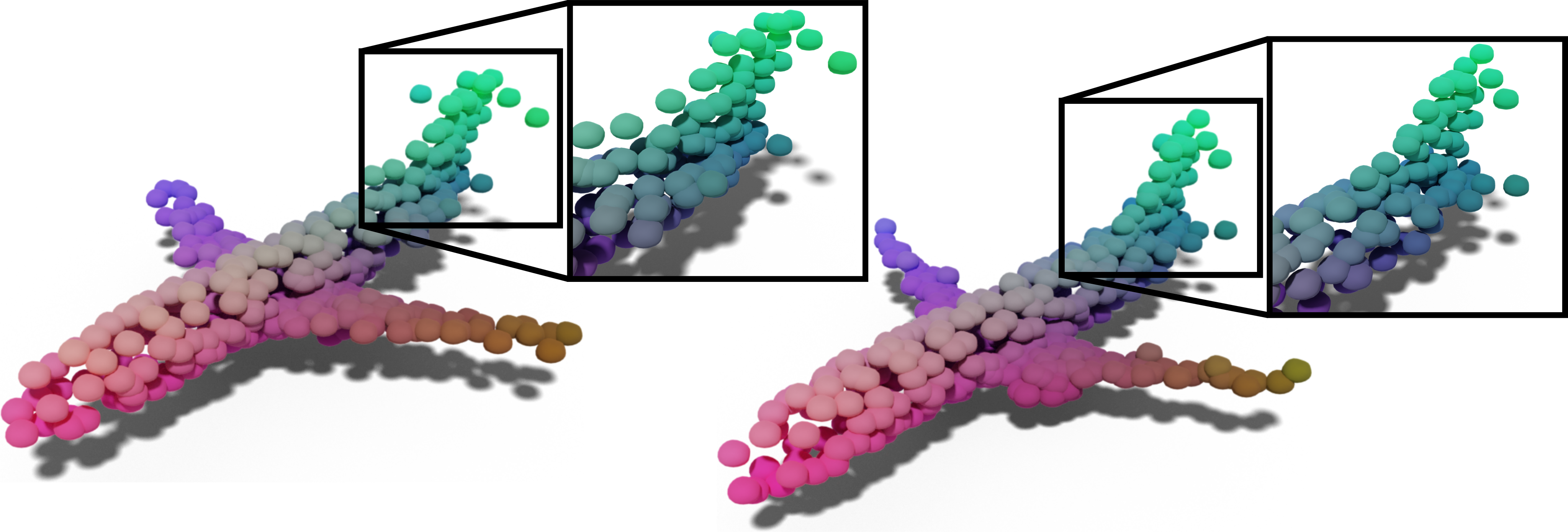}
        \put(25,25){}
        \end{overpic}  
\end{center}
    \caption{The first (left) and the last (right) steps for the $d\Lambda_{\subX}$ interpolation depicted in Figure \ref{fig:airplanes}, with a close-up on the tails.}
    \label{fig:airplanescloseup}
\end{figure}

\begin{figure*}[t]
\footnotesize
\hspace{3cm}
\centering
\begin{overpic}
    [trim=0cm 0cm 0cm 0cm,clip,width=1\linewidth]{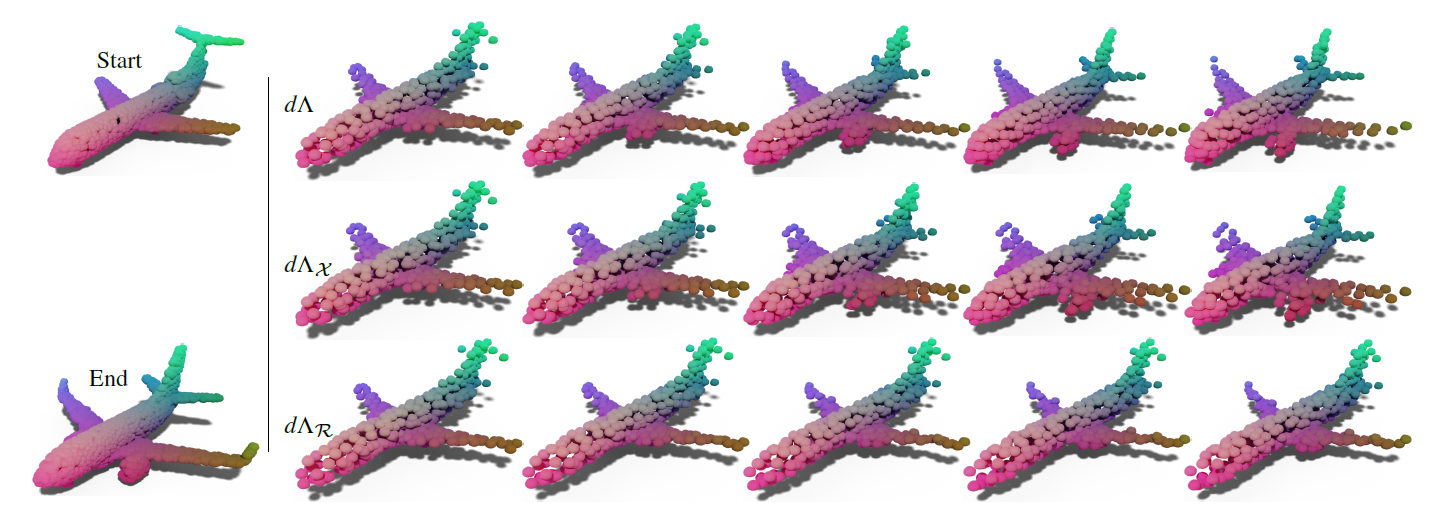}
\end{overpic}
    \caption{Casting different kinds of spectra interpolation into our network gives us different degrees of control. On the left, the models used as initial and final steps; on the right, we interpolated the entire spectral encoding ($d\Lambda$-first row), only the global frequencies ($d\Lambda_{\X}$-second row), and only the local ones ($d\Lambda_{\subX}$-third row).}
    \label{fig:airplanes1515}
\end{figure*}

\begin{figure*}[b]
 \footnotesize
 \centering
 \begin{overpic}
    [trim=0cm 0cm 0cm 0cm,clip,width=1\linewidth]{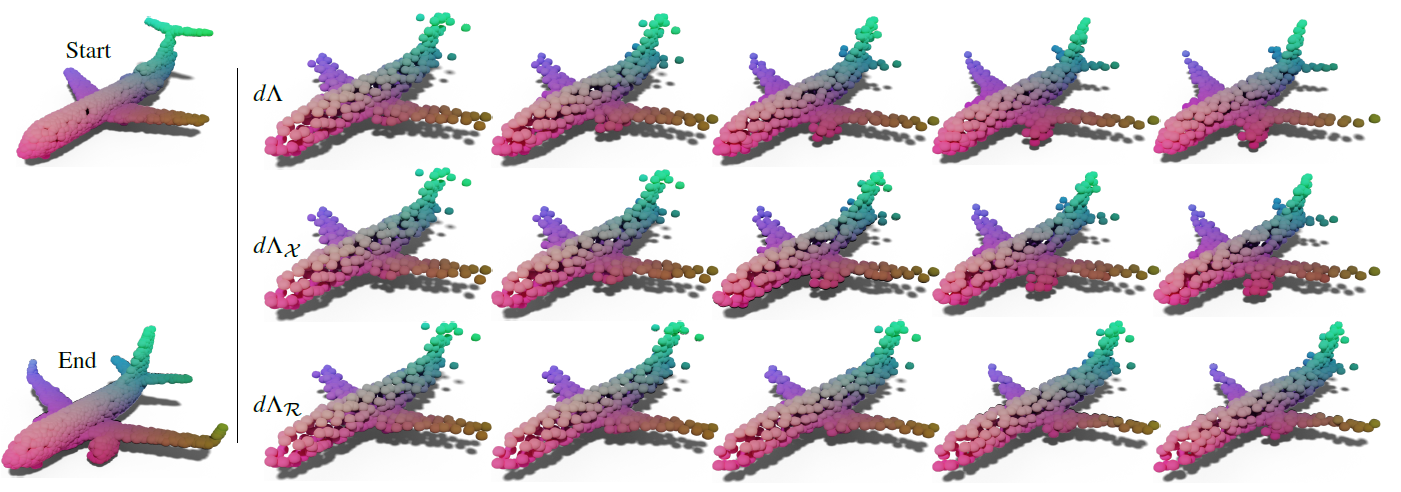}
\end{overpic}
     \caption{Casting different kinds of spectra interpolation into our 20+10 network. On the left, the models used as initial and final steps; on the right, we interpolated the entire spectral encoding ($d\Lambda$-first row), only the global frequencies ($d\Lambda_{\X}$-second row), and only the local ones ($d\Lambda_{\subX}$-third row).}
     \label{fig:airplanes}
 \end{figure*}

{In  Fig.~\ref{fig:airplanes1515} we test our model by looking at the shape generated from the spectra obtained by interpolating the input {spectral encoding} of two shapes (depicted on the left). In the first row, we report the results from the interpolation of the whole {spectral encoding}. We can see that the deformation is smooth both in size (i.e., length of the structure) and features (i.e., turbines appearing, tail morphing). In the second row, we fix the local part of the encoding, and interpolate the global. Coherently, changing the whole structure also requires changing the tail structure (different kinds of airplanes have different tails). Finally, in the third row, we only manipulate the local part maintaining the global one. The local interpolation mainly impacts the tail region (a close-up is depicted in Fig.~\ref{fig:airplanescloseup}), which follows the interpolation pattern of other rows. Remarkably, other global aspects of the airplanes are only slightly modified (i.e., the turbines and the shapes of the wings are almost left unchanged). We consider this result significant, since the spectrum of the tail seems representative enough to relate with different  airplanes. Moreover, our global plus local spectral encoding provides nice interpolation results.}
In Fig. \ref{fig:airplanes} we report the same example of Fig.~\ref{fig:airplanes1515}, but using a 20+10 network instead of a 15+15 one. The results are consistent, showing a certain resilience to different settings.

\end{document}